\documentclass{article}

\usepackage{arxiv}

\usepackage[utf8]{inputenc} 
\usepackage[T1]{fontenc}    
\usepackage{hyperref}       
\usepackage{url}            
\usepackage{booktabs}       
\usepackage{amsfonts}       
\usepackage{amssymb}
\usepackage{nicefrac}       
\usepackage{microtype}      
\usepackage{lipsum}         
\usepackage{graphicx}
\usepackage{natbib}
\usepackage{doi}
\usepackage{xcolor}
\usepackage{caption}
\usepackage{subcaption}
\usepackage{pgf}
\usepackage{tikz}
\usepackage{fontawesome}
\usepackage{algorithm}
\usepackage{algpseudocode}
\usepackage{amsmath}
\usepackage{cleveref}
\usepackage{multirow}
\usepackage{textcomp}

\title{A pipeline for multiple orange detection and tracking with 3-D fruit relocalization and neural-net based yield 
regression in commercial citrus orchards}

\author{ \href{https://orcid.org/0000-0002-9272-3403}{\includegraphics[scale=0.06]{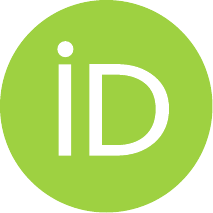}\hspace{1mm}Thiago T. Santos}\\
	Embrapa Digital Agriculture\\
	\texttt{thiago.santos@embrapa.br} \\
	\And
	\href{https://orcid.org/0000-0002-1907-5390}{\includegraphics[scale=0.06]{orcid.pdf}\hspace{1mm}Kleber X. S. de~Souza} \\
	Embrapa Digital Agriculture\\
	\texttt{kleber.sampaio@embrapa.br} \\
	\And
	\href{https://orcid.org/0009-0004-9393-8205}{\includegraphics[scale=0.06]{orcid.pdf}\hspace{1mm}João Camargo~Neto} \\
	Embrapa Digital Agriculture\\
    \texttt{joao.camargo@embrapa.br} \\
	\And
	\href{https://orcid.org/0000-0001-7673-296X}{\includegraphics[scale=0.06]{orcid.pdf}\hspace{1mm}Luciano V. Koenigkan} \\
	Embrapa Digital Agriculture\\
    \texttt{luciano.vieira@embrapa.br} \\
   	\And
	\href{https://orcid.org/0000-0002-0414-9852}{\includegraphics[scale=0.06]{orcid.pdf}\hspace{1mm}Alécio S. Moreira} \\
	Embrapa Digital Agriculture\\
    \texttt{alecio.moreira@embrapa.br} \\
	\And
	\href{https://orcid.org/0000-0002-4276-3263}{\includegraphics[scale=0.06]{orcid.pdf}\hspace{1mm}Sônia Ternes} \\
	Embrapa Digital Agriculture\\
    \texttt{sonia.ternes@embrapa.br} \\
}

\renewcommand{\shorttitle}{A pipeline for MOT with 3-D relocalization and NN-based yield 
regression in citrus}

\hypersetup{
pdftitle={A pipeline for multiple orange detection and tracking with 3-D fruit relocalization and neural-net based yield 
regression in commercial citrus orchards},
pdfauthor={T.~T.~Santos},
pdfkeywords={Orange detection, Orange counting, Multiple-object tracking, Yield estimation, Deep learning, Digital agriculture},
}

\begin{document}

\newcommand{\cmt}[2]{{\color{red} \small {\bf #1} \it #2}}

\newcommand{\Xt}{$\mathbf{X}$\space}
\newcommand{\Xti}{$\mathbf{X}_i$\space}
\newcommand{\Xtj}{$\mathbf{X}_j$\space}
\newcommand{\xbi}{$\mathbf{x}_i$\space}
\newcommand{\xbj}{$\mathbf{x}_j$\space}
\newcommand{\cbi}{$\mathbf{c}_i$\space}
\newcommand{\cbj}{$\mathbf{c}_j$\space}
\newcommand{\bbi}{$\mathbf{b}_i$\space}
\newcommand{\bbj}{$\mathbf{b}_j$\space}
\newcommand{\Pji}{$\mathtt{P}_i$\space}
\newcommand{\Pjj}{$\mathtt{P}_j$\space}
\newcommand{\Ss}{$\mathcal{S}$\space}
\newcommand{\Lost}{\textsc{Lost}\space}
\newcommand{\Active}{\textsc{Active}\space}

\maketitle

\begin{abstract}
Traditionally, sweet orange crop forecasting has involved manually counting fruits from numerous trees, which is a labor-intensive process. Automatic systems for fruit counting based on proximal imaging, computer vision, and machine learning, have been considered a promising alternative or complement to manual counting. These systems require data association components that prevent multiple counting of the same fruit observed in different images. However, there is a lack of work evaluating the accuracy of multiple fruit counting, especially considering (i) occluded and re-entering green fruits on leafy trees, and (ii) counting ground-truth data measured in the crop field. Here, we propose a non-invasive alternative that utilizes fruit counting from videos, implemented as a pipeline. Firstly, we employ convolutional neural networks for the detection of visible fruits. Inter-frame association techniques are then applied to track the fruits across frames. To handle occluded and re-appeared fruit, we introduce a relocalization component that employs 3-D estimation of fruit locations. Finally, a neural network regressor is utilized to estimate the total number of fruit, integrating image-based fruit counting with other tree data such as crop variety and tree size. The results demonstrate that the performance of our approach is closely tied to the quality of the field-collected videos. By ensuring that at least 30\% of the fruit is accurately detected, tracked, and counted, our yield regressor achieves an impressive coefficient of determination of 0.85. To the best of our knowledge, this study represents one of the few endeavors in fruit estimation that incorporates manual fruit counting as a reference point for evaluation. We also introduce annotated datasets for multiple orange tracking (\textsc{MOrangeT}) and orange detection (\textsc{OranDet}), which are publicly available and aim to foster the development of novel methods for image-based fruit counting.
\end{abstract}

\keywords{Orange fruit detection \and Orange fruit counting \and Multiple-object tracking \and Crop yield estimation \and Deep learning \and Digital agriculture}

%
%
\section{Introduction}

Accurate fruit yield estimation is important for making informed decisions about harvesting, 
storage, and marketing, but such estimation can be a challenging task. Consider the major citrus belt 
in Brazil, which covers areas in two states (São Paulo and Minas Gerais) and sums 461,921 hectares,
86\% reserved to sweet orange production, according to the Fund for Citrus Protection, 
\citet{fundecitrus2022}. The most used annual orange crop forecast process, performed by Fundecitrus, is a 
laborious operation that involves manual fruit stripping (the advanced harvest of all fruit in the tree), 
in samples of around 1,500 trees spread across this citrus belt. Such sampling and counting procedures are
a usual way to obtain pre-harvest fruit yield data \citep{wulfsohn2012mlevel,he2022fruit}.

Computer vision-based methods have been considered as a prominent alternative for automatic, non-invasive 
fruit counting, becoming parts of larger yield estimation systems.  Earlier works employed \emph{feature engineering} 
\citep{gongal2015sensors} to create fruit detectors, exploring low-level features in images derived from color and 
texture. An example for citrus was presented by \citet{maldonado2016automatic}, that employed image processing 
techniques and support vector machines to detect green oranges in field images. With the popularization of deep 
learning, convolutional neural networks (CNNs) were adopted for automatic feature learning (\emph{representation 
learning}) \citep{lecun2015deep}. The work by \citet{sa2016deepfruits} is a turning point in the adoption of CNNs 
in fruits detection, and this passage from feature engineering to representation learning is reviewed by 
\citet{koirala2019deep}. 

Despite significant progress in automated fruit detection using deep learning techniques, the 
\emph{counting} stage of the processing pipeline remains a challenging area that requires further 
development. Single images cannot provide reliable estimates of fruit counting because of
occluded fruit: a single standpoint keeps a significant part of the fruit out of sight \citep{wang2019mango}.
A natural alternative is to employ multiple images (or, equivalently, video sequences), to get different
views able to reach most of the targets and perform counting of the detected fruits.  
However, while single-view fruit counting presents underestimation issues, a major limitation of 
multiple-view fruit counting is overestimation of yield due to overlapping images, which must be 
properly managed to ensure accurate counting \citep{villacres2023apple}. An approach being explored 
recently is to scan the orchards using cameras and to assign a unique identifier to each 
fruit along the image sequence \citep{jong2022apple,villacres2023apple}. Such approach, 
\emph{Multi Object Tracking (MOT)}, involves detecting multiple objects in a sequence of images 
and linking them over time to identify individual objects \citep{luiten2021hota}. This is a 
challenging task that requires precise detection, localization, and association of a same object in 
different images. Essentially, the goal is to identify each object's location, movement, and 
identity throughout the entire sequence, avoiding multiple counting of the same object (a fruit).
Fruit detection and tracking can also be considered a component of \emph{semantic scene understanding} 
systems\footnote{Or \emph{SpatialAI} systems, as proposed by \citet{davison2018spatialai}.} for 
agricultural settings, allowing artificial agents to extract semantic information about 
the objects in an orchard and interact usefully with its environment. Examples of such interactions
are crop monitoring, spraying and harvesting \citep{duckett2018agrobot}.

Our objective is to contribute to crop forecast by providing estimates of the actual number of fruits
on a tree. We start with the challenging task of \emph{counting visible oranges} on individual trees under 
real citrus farm conditions using video inputs. Yield prediction is performed in earlier maturation stages of sweet orange, 
when the fruit is green (non-mature), lacking the color contrast against green leaves observed in 
mature fruit \citep{gong2013citrus}. The imaging is performed on ground level, between trees rows, employing 
smartphone  cameras. In such a setting, there is an issue: tall trees (up to 5 meters height) cannot be fully 
imaged in a single frame. In other words, considering the available inter-row space, the trees' height and the 
absence of wide-angle optics, it is not possible to keep the entire tree in the camera's field of view (FOV). Even in 
the case of wide-angle optics availability, a single view of a tree cannot provide a full assessment of the 
entire plant fruit set because of occlusions. Therefore, we need of a multiple-view counting system to solve this issue. After fruit 
counting, we proceed to the \emph{yield estimation} for each tree. This value differs 
from the counted fruit by any computer vision system, since there are oranges located in the inner part 
of the plant canopy, not visible in the images collected in the field even considering multiple views. 
We implement a neural net-based regressor to estimate the actual number of 
fruit from the fruit counting, integrating other data as plant height, variety, and age. Although we use a 
different set of methods, our work is similar to the one by \citet{hani_etal2020} in that they also propose 
a pipeline comprising fruit detection, 3-D projection, fruit counting and yield estimation in apple orchards. 
However, our study faces a more challenging scenario considering occlusions and fruit tracking.

The main contributions of this work are:
\begin{itemize}
  \item a simple but effective annotation methodology for multiple object tracking 
  in citrus, exploiting camera pose data by structure from motion to estimate the 
  fruits' 3-D location and automate the generation of 2-D tracks ground-truth;
  \item the introduction of \textsc{MOrangeT} \citep{dataMOrangeT}, a public MOT dataset for citrus composed by 
  images sequences (frames) recorded on citrus farms plus ground-truth tracking data in the 
  MOT16 format \citep{milan2016mot16}, whose annotation effort was performed by our team;
  \item a multiple fruit detection and tracking method based on convolutional 
  networks for detection, on the Hungarian algorithm \citep{kuhn1955hungarian} for inter-frame 
  association, and a relocalization algorithm to treat occlusions and fruits 
  exiting/entering the camera's field of view;
  \item an evaluation of the proposed method employing the \emph{multiple object tracking accuracy}
(MOTA) \citep{bernardin2008evaluating} and the \emph{higher order tracking accuracy} (HOTA) \citep{luiten2021hota}, and 
  \item proposition of a regressor based on artificial neural networks to estimate the number
  of fruit in a tree given the number of fruit counted using MOT, validated using a set of 1,139 trees 
  with real yield manually collected.
\end{itemize}
A useful property of the proposed tracking method is the fruit's relative position in 3-D is also estimated, 
allowing assessment of the spatial distribution of fruits in the tree.

This work is organized as follows. In Section~\ref{sec:relatedwork} we provide a comprehensive overview of related works in the 
field of fruit counting and yield estimation. We discuss the various methods employed for fruit detection and inter-frame fruit 
association. Section~\ref{sec:methods} presents our proposed methodology, which encompasses the \textsc{MOrangeT} dataset, a 
novel annotation technique for MOT ground-truth, the methods employed for orange tracking, and tree yield estimation. 
The results of our experiments are presented in Section~\ref{sec:results}, followed by a detailed discussion of the findings in 
Section~\ref{sec:discussion}. Finally, in Section~\ref{sec:conclusion}, we present our concluding remarks, summarizing the main 
contributions of this work and outlining potential paths for future research. For a visual overview, a video showing our 
fruit counting results is available\footnote{\url{https://youtu.be/hOq42KMskLQ}}.

%
%
\section{Related work}
\label{sec:relatedwork}

After the raise of CNN-based methods for fruit detection, initiated by \citet{sa2016deepfruits}, researchers have 
explored the problem of fruit counting in multiple images, considering important challenges as fruit recounting and
occlusions \citep{wang2019mango, jong2022apple, villacres2023apple}. Most of these works add a 
\emph{data association} procedure to link detected oranges in different images, avoiding multiple counting, and 
commonly employing a tracking algorithm. Some researchers extended their fruit counting systems to 
yield estimation \citep{liu2019monocular, hani_etal2020, koirala2021unseen}, generally using (linear) regression.

\citet{liu2019monocular} proposed a fruit counting system tested on data from mangoes orchards. A dataset 
of $1,500$ images, presenting a resolution of $500 \times 500$ pixels, was employed to train a Faster~R-CNN detector 
\citep{ren2017faster}. The mangoes' tracking was performed by a method employing Kanade-Lucas-Tomasi algorithm 
(\emph{optical flow} or KLT), Kalman filters, and the Hungarian algorithm \citep{kuhn1955hungarian}. To deal with 
long-time occlusions and reappearing fruits, they employed a relocalization procedure based on the fruit's 
estimated 3-D position and reprojection on the current frame. They used \emph{structure from motion} (SfM) 
\citep{dellaert2021factor}, using the mangoes as \emph{3-D landmarks} and their tracked positions in images as 
\emph{measurements}. In SfM, the three-dimensional positions of landmarks and the camera are estimated from the 
measurements (2-D positions in each frame) using non-linear least squares \citep{hz2004mvg, dellaert2021factor}. 
Liu {\it et al.} employed COLMAP \citep{schonberger2016colmap} as SfM implementation, getting estimations for the 
mangoes locations in 3-D and the camera position in space at each frame. To avoid multiple counting, 
mangoes/landmarks and Faster~R-CNN detected bounding boxes are reassociated using the Hungarian algorithm and a specially 
designed association cost function. Liu {\it et al.} achieved a coefficient of determination ($R^2$) of $0.78$ 
in a linear regression model relating fruit count to yield after considering views from two different vantage points 
of two rows facing opposite sides of the trees.

Concurrently, \citet{wang2019mango} proposed another image-based counting system for mangoes. Using 10 fps video
sequences scaled to $1024 \times 1024$ pixels frames, they employed their MangoYOLO model, a 33-layers adaption 
of YOLOv3 \citep{koirala2019mangoyolo}, to detect fruits. Kalman filter was employed on tracking, estimating the 
fruit positions in the next frame. The Hungarian algorithm was used to assign predicted fruit positions to 
detections in the current frame, using the fruits' centroids distances on the cost function. Wang et al. 
adopted a threshold for lost trackers: after 15 frames without associations to MangoYOLO detections, the Kalman-based 
tracker is ended -- if the fruit re-appear, it is considered a new fruit, incrementing counting. In an experiment 
involving a row containing 21 trees, the authors reported a root-mean-square error (RMSE) of 18~fruits/tree. However, reported results
indicate a canceling occurring in the mean error between overestimation by repeated counting and underestimation by 
missed fruits.  

\citet{gan2020active} proposed a novel approach where a vehicle carrying nozzles applied a water mist to citrus
trees. The mist induced temperature contrast between fruits and leaves and a thermal camera, also embedded in the
vehicle, recorded video streams of thermal frames. Careful experimentation was conducted to find optimal 
parameters for the amount of water and kind of nozzle, and the impact of ambient temperature and humidity considering
the goal of maximizing temperature contrast. A Faster~R-CNN model was trained for orange detection in the thermal
images, reaching 87.2\% average precision (AP). KLT tracking was employed to avoid multiple counting, but the data 
association used by the authors for multiple fruit tracking is unclear. In a row containing 25 trees, their method was able to count 
96\% of the 747 manually counted fruits in the field. However, the thermal camera's FOV was unable to cover 
the entire canopy, so the manually collected ground-truth was restricted to the view area, marked using nylon strings.

\citet{gao2022novel} have explored the modern fruiting-wall growing system for apples, where the 
apple tree canopy is thin, making the fruits and trunks visible. Fruit detection and tree trunk detection 
were performed using the YOLOv4-tiny architecture \citep{bochkovskiy2020yolov4}, a fast neural network 
aimed for real-time applications. They assume a linear movement of the camera 
(embedded in a vehicle), so their tracking is adapted to deal with horizontal movements only. The 
CSR-DCF algorithm \citep{lukezic2017csr}, a single target tracker, was employed on trunk tracking, 
and the estimated horizontal movement guided the apples tracking. Inter-frame fruit assignment is 
guided by minimum Euclidean distance between fruits' centroids, but the employed assignment 
algorithm is not completely clear. The counting system, evaluated using annotated videos (not the true 
yield or the manual counting on the field), reached $R^2 = 0.98$. However, the reported results show 
detection errors in a single frame (mainly false negatives) produce immediate tracking errors.

The APPLE MOTS dataset was introduced by \citet{jong2022apple}. Two UAVs and a wearable device, 
where a camera is embedded in a helmet, were employed to get video sequences in orchards 
presenting three different apple varieties. The CVAT annotation tool was employed 
to produce almost 86,000 manually annotated masks associated to 2,304 unique apple instances 
across 1,673 frames. The authors further evaluated two different MOT algorithms, 
TrackR-CNN \citep{voigtlaender2019mots} and PointTrack \citep{xu2020pointrack}. 
The best results were produced by PointTrack, that 
reached 52.9 in the multi-object tracking and segmentation accuracy (MOTSA) metric, a variation of the MOTA metric \citep{milan2016mot16} 
for segmented objects\footnote{In the segmentation case, \emph{masks} are provided as object
segmentation ground-truth, not just the rectangular bounding boxes.}. The authors call  
attention to the fact that multiple fruit tracking involves \emph{homogeneous objects}, 
i.e., objects very similar to each other. This differs from the people tracking, a popular 
task in computer vision because of research on surveillance and autonomous vehicles. 
De~Jong~et~al. consider if this homogeneity in object appearance could impact methods developed 
originally for pedestrian  tracking in different ways.

Another work employing a MOT formulation was presented by \cite{he2022cascade}, again following a tracking
by detection framework: a YOLOv3 network \citep{redmon2018yolov3} was employed to find camellia fruit and apples, 
and Kalman filters used on fruit tracking. The Mahalanobis distance and the Kalman-based predictions are employed 
to define eligible associations between tracks and new detections in the current frame. Then a similarity 
function based on appearance is employed on association by a nearest-neighbor algorithm. Such a combination of 
spatial filtering and appearance similarity was named \emph{Cascade-SORT} by the authors. \citet{he2022cascade}, have 
reached MOTA values beyond 0.70 for most of their tests. However, their results are composed of just four short
video sequences (one for camellia and three for apples).

\citet{zhang2022citrus} proposed a multiple object tracking method for citrus in-field, composed of a detection component, 
OrangeYOLO, and a tracking component, OrangeSORT, which are modified versions  of the YOLOv3 network \citep{ 
redmon2018yolov3} and the SORT algorithm \citep{bewley2016sort}, respectively. A wide-angle action camera\footnote{A DJI Osmo 
Action camera, presenting {145\textdegree} FOV.} was adapted to a field rover, that performed a linear movement across 
orchards' rows at 2~m/s. Zhang~{\it et al.} carefully selected three scales from YOLOv3's backbone (Darknet53) after 
analyzing the receptive fields and the oranges' sizes, getting significant improvements in detection. OrangeYOLO also 
presents a channel-spatial attention mechanism, but the ablation experiment results showed a marginal gain in detection by 
this extension to the architecture. OrangeSORT computes the average motion displacement observed in the tracked fruits 
to update the state of lost trackers (Kalman filters), i.e., trackers not associated to any OrangeYOLO's detection at the
current frame. In the six videos collected in-field for testing, counting error varied from 0.61\% (the best case) to 
24.29\% (worst case).

\citet{villacres2023apple} evaluated five different MOT algorithms for tracking apples in-field: (i) multiple Kalman 
filters combined with the Hungarian algorithm; (ii) kernelized correlation filter; (iii) multiple hypothesis tracking 
(MHT) \citep{kim2015mht}; (iv) SORT \citep{bewley2016sort} and (v) DeepSORT \citep{wojke2018deepsort}. They trained 
the employed apple detectors, Faster R-CNN and YOLOv5, using 
public datasets as MinneApple \citep{hani2020minneapple} and Fuji-SfM  \citep{gene2020fuji}. Nine videos recorded 
in-field using smartphones\footnote{Smartphones iPhone 6S Plus and iPhone 13, Apple Inc.} were employed on 
the apples tracking evaluation. Annotations in the MOT format \citep{milan2016mot16} for the videos were produced 
using the open-source tool CVAT\footnote{CVAT -- \url{http://cvat.ai}.}. The authors conducted an interesting 
sensitivity analysis, employing the ground-truth bounding boxes as input for the trackers, but degrading the 
detection probability from 100\% (all boxes in the ground-truth) to 20\%. MHT and DeepSORT showed the 
best tracking performances, even at 60\% detection rate. When considering detections from the neural nets, 
YOLOv5 produced the best detections for tracking, while DeepSORT produced the best results: 20.1\% error on 
average for fruit counting.  

What is the proper way to evaluate MOT trackers? According to \citet{luiten2021hota}, evaluation metrics serve two 
primary purposes. Firstly, they allow for straightforward comparison between various methods to determine which ones 
perform better than others. To achieve this objective, it is recommended to have a single metric that can be used to 
rank and compare these methods. Secondly, evaluation metrics are essential for analyzing and comprehending the 
various types of errors that algorithms make, identifying where they are likely to fall short when applied. The HOTA 
metric \citep{luiten2021hota} was proposed with these two objectives in mind for the complex task of multiple object 
tracking. As stated by its proponents:
\begin{quote}
HOTA  measures  how  well  the  trajectories  of  matching detections align, and averages this over all matching 
detections, while also penalizing detections that don't match.
\end{quote}
Few works have employed proper MOT metrics for multiple fruit tracking \citep{jong2022apple, villacres2023apple}, 
most of them employing the \emph{multiple object tracking accuracy} (MOTA) metric proposed by 
\citet{bernardin2008evaluating}. However, MOTA has been criticized because it tends to prioritize detection over 
association \citep{luiten2021hota}, while HOTA offers a balanced combination of detection and association scores. 

Lost detections, caused by occlusions and the harsh light conditions in orchards, lead to fruit track breaks and 
yield overestimation, as noted by \citet{he2022cascade} and observed in results by \citet{gao2022novel}. We argue that
a \emph{long-term process}, as relocalization, is needed for fruit counting, providing robustness to 
\emph{short-term} detection association in neighbor video frames. The same was
speculated by \citet{villacres2023apple} while \citet{liu2019monocular} developed their landmark-based 
re-association method to tackle this issue. The present work is closely related to \citep{liu2019monocular}:
the 3-D information is employed to, using geometrical restrictions, properly re-assigned lost fruits, and 
both works use SfM to get 3-D information. However, \citet{liu2019monocular} use the fruits as landmarks, 
an interesting approach to reduce SfM complexity, but it raises questions about SfM performance when only 
a few fruits (or none) are seen by the camera. In such case, the SfM framework could not have landmarks to 
proper camera motion estimation. In our method, SfM is performed to get the \emph{camera relative motion} (ego-motion), and 
fruit localization in 3-D is performed as an independent process.
The tracking method presented in our work, however, is more similar to SORT-based methods, in the sense the assignments 
are performed online if ego-motion data is available, differently from the offline SfM processing in \citet{liu2019monocular}. 
Our approach is complementary to the method presented by \citet{he2022cascade}, which assumed unknown camera position (no ego-motion) 
and employed Kalman filter-based tracking to predict the trajectories of occluded or misdetected fruits. Conversely, we explore 
relocalization for long-term tracking, based not on appearance, but on 3-D localization estimated using ego-motion data. 

The methods for yield estimation vary depending on the type of data collected or the crop.  \citet{wulfsohn2012mlevel} 
sample trees from an orchard and select branches and segments of branches using a multi-stage systematic sampling 
approach. They then apply statistical methods to propose a yield estimator. On the other hand, \citet{hani_etal2020} used 
machine learning and tracking to provide a direct estimation of fruits. In their case, the structure of the plants makes 
the fruit visible on the surface of the canopy, allowing direct counting. Direct estimation of fruits based on images 
was also performed by \citet{maldonado2016automatic}. In our problem, direct estimation is not viable because we 
identified a large gap between the visible fruits and the total number of fruits obtained through manual harvesting. 
This gap may result from the fact that some fruit is located inside the canopy. To address this issue, we propose a 
neural network regressor for the yield that considers the number of identified fruits and other relevant variables, as 
detailed in Section~\ref{subsec:regressor}. A similar approach was used in estimating mango fruit load, where 
\citet{koirala2019mangoyolo} applied a correction factor to estimate the fruit load per orchard. 
However, unlike the regressor we are proposing, this correction factor is calculated per orchard based on the average 
ratio of the number of fruits identified in images to the manual harvest count per tree in the field.

%
%
\section{Materials and methods}
\label{sec:methods}

\begin{figure}
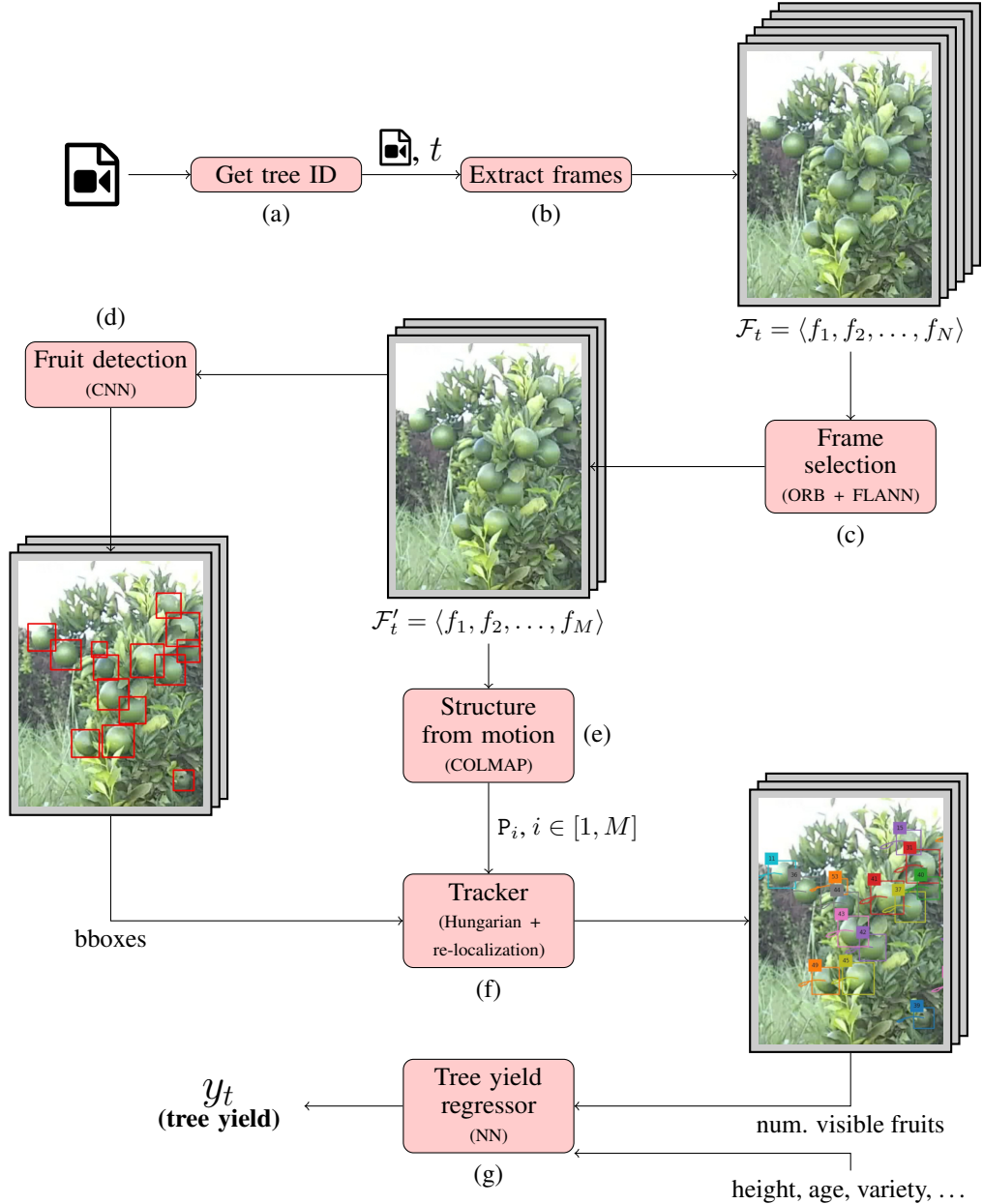

\centering

\input pipeline.tikz

\caption{Pipeline overview. In the input video (single face of a single tree), a marker identifying the tree $t$ is visible 
in the first frames. The tree is identified (a), and its frame sequence $\mathcal{F}_t$ is extracted (b). A frame selection 
procedure removes redundant frames (few or no motion), producing a reduced sequence $\mathcal{F}_t'$ (c). Fruit detection 
by a CNN (d) and Structure from motion by COLMAP (e) are performed for the frames in $\mathcal{F}_t'$. The ego-motion 
(set of projective matrices $\mathtt{P}_i$) and the found bounding boxes feed the tracker (f). Finally, the number of 
found visible fruits and other tree data (height, age, \dots) are employed by a regressor to estimate the total number 
of oranges in tree $t$ (g).} 
\label{fig:overview}
\end{figure}

Our pipeline takes video files that record both faces of a tree, one for each side of the row, in the field. 
It then generates an estimation of the total yield for that tree, measured in the number of fruits. 
Figure~\ref{fig:overview} shows a pipeline overview. Videos are recorded by field 
staff (Section~\ref{sec:vidrecording}), and ID numeric plates are displayed in the first frames to identify each tree. 
The first step of the pipeline is to identify the plate and extract the tree ID $t$. The entire sequence of video
frames is extracted from the video file, and a frame sampling procedure (Section~\ref{sec:framesampling}) selects a subset. 
As the recording is performed by handheld cameras (smartphones), and the human camera operator slowly films the tree, several 
frames can be very similar because of too slow camera motion: the frame selection employs the well-known framework of feature 
detection and matching \citep{muja2014flann} to select frames presenting distinguished changes, but avoiding large movement 
that could prejudice ego-motion estimation and fruit tracking. The selected frames feed a structure from motion procedure 
that estimates the camera parameters, including position and orientation, for each frame (Section~\ref{sec:egomotion}). The 
selected frames are also inputted to a fruit detection neural network, producing a set of rectangular bounding boxes that 
mark the observed oranges in each frame (Section~\ref{sec:det}). Our fruit tracking procedure (Section~\ref{sec:mot}) 
takes the camera parameters and the detected boxes in each frame to track the oranges in the frame sequence. A 3-D 
relocalization component deals with occlusions and fruits exiting and entering the camera FOV. The bounding boxes are 
grouped in \emph{tracks}, defining the path that every single orange performed in the frame sequence. These tracks are not 
necessarily \emph{contiguous}, properly dealing with long-term occlusions and fruit reappearance. The number of tracks 
corresponds to the number of observed fruits in the video. Finally, this number is combined with the value found for the 
other face of the tree and with additional data, such as tree variety, height, and plant age. Such data is inputted to a yield 
regressor (Section~\ref{sec:yieldreg}), that produces a final estimation of the number of fruits $y_t$ in the tree $t$.

\subsection{Dataset}

The pipeline shown in Figure~\ref{fig:overview} was executed for a set with more than 1,500 sweet orange trees. However,
the computer vision-based components need annotated datasets for training and evaluation. Such annotation effort, although
favored by the new annotation tool presented in Section~\ref{sec:annot}, could not be performed for the entire video set. 
A subset of 12~videos was annotated for multiple object tracking evaluation, forming the \textsc{MOrangeT} dataset. Part
of these annotations were combined to previously annotated images to form another dataset, \textsc{OranDet}, built for
orange detection. Such data allowed the training and evaluation of the fruit detection and tracker
modules in the pipeline. 

Although it was not feasible to evaluate multiple object tracking with all 1,543 trees, the whole set was processed 
through the pipeline and forwarded to the yield regressor. This last step, however, processed only 1,139 trees, which 
satisfied the requirement of having at least a fruit counting for one of its sides, as well as other information 
considered relevant, as one can see in Section~\ref{subsec:dataregress}. The following sections provide a more in-depth description of such data.

\subsubsection{Orange Crop Forecast data}

The Orange Crop Forecast, conducted by \citet{pes2022forecast}, utilized a stratified random sampling technique to select 
a representative sample of 1,560 orange trees. The initial drawing involved 1,200 trees that were proportionally 
distributed across the citrus belt, stratified based on their region, variety, and age. An additional drawing included 360 
resets of younger ages to replace trees lost due to diseases, and other causes. To 
obtain the sample, a fruit stripping method was employed, involving the advanced harvest of all fruits from each tree. 
During the stripping process in the field, fruits from different flowering periods and 
consequently, in different phenological stages were sampled: green fruit close to the final size (classified as fruit from 
the 1st flowering period, \emph{F1}); green fruit with a table tennis ball size ($\sim$~4.5 cm diameter, classified as 
fruit from the 2nd flowering period, \emph{F2}); and green fruit with a marble ball size (up to $\sim$~3.0 cm diameter, 
classified as from the 3rd period, \emph{F3}) \citep{barbasso2005feno}. The fruit-stripping operation took place between 
March 28 and May 11, 2022, and the harvested fruits were transported to a laboratory in Araraquara, SP, Brazil. At the 
laboratory, the fruits were sorted into different bloom categories and quantified using automatic counting equipment 
before being weighed. Fundecitrus provided the ground-truth for fruit counting, tree height, age, and fruit variety for 
1,543 trees. 

\subsubsection{Video recording}\label{sec:vidrecording}

\begin{figure}
     \centering
     \begin{subfigure}[b]{0.45\textwidth}
         \centering
         \includegraphics[width=\textwidth]{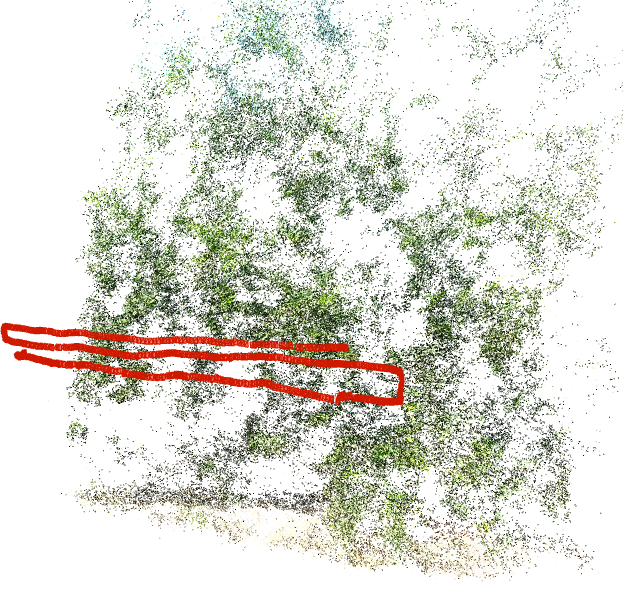}
         \caption{Recording a 4.9~m height {\it Valencia} orange tree, camera path in red.}
         \label{fig:sfm}
     \end{subfigure}
     \hfill
     \begin{subfigure}[b]{0.45\textwidth}
         \centering
         \includegraphics[width=\textwidth]{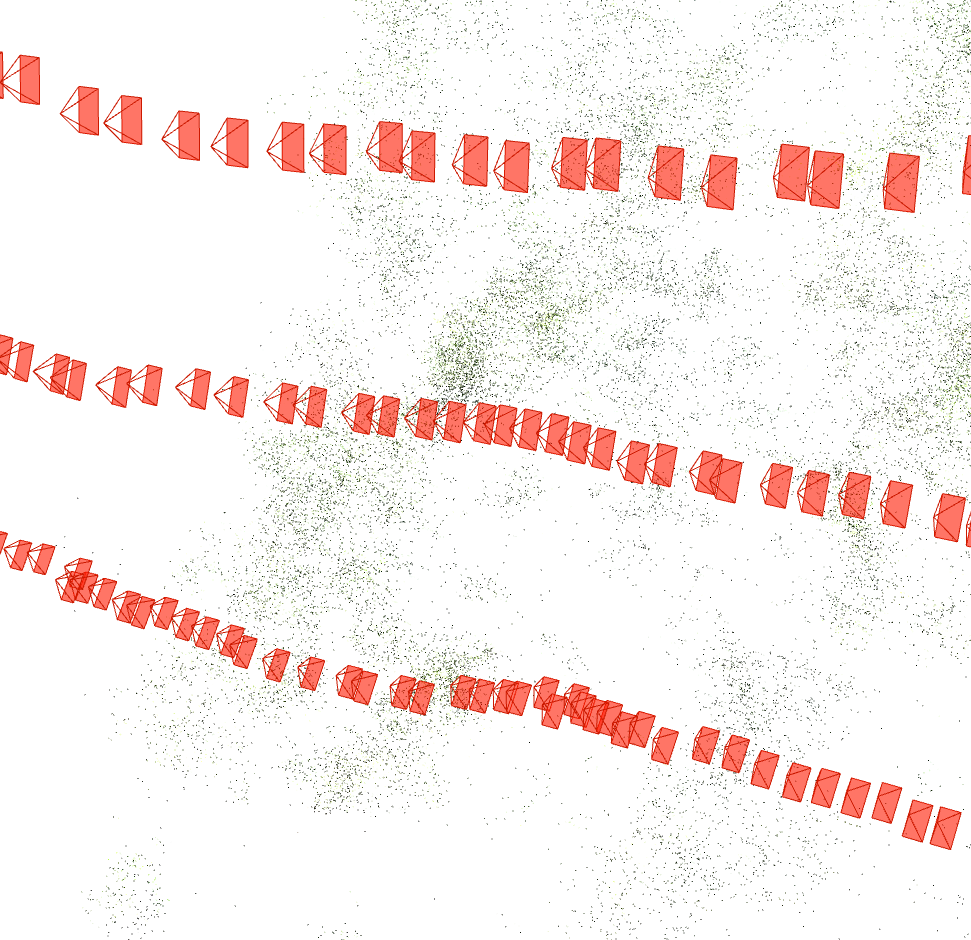}
         \caption{Detail: each red pyramid shows the camera's projection center position and orientation.}
         \label{fig:sfm-detail}
     \end{subfigure}\\
     \begin{subfigure}[b]{0.18\textwidth}
         \centering
         \includegraphics[width=\textwidth]{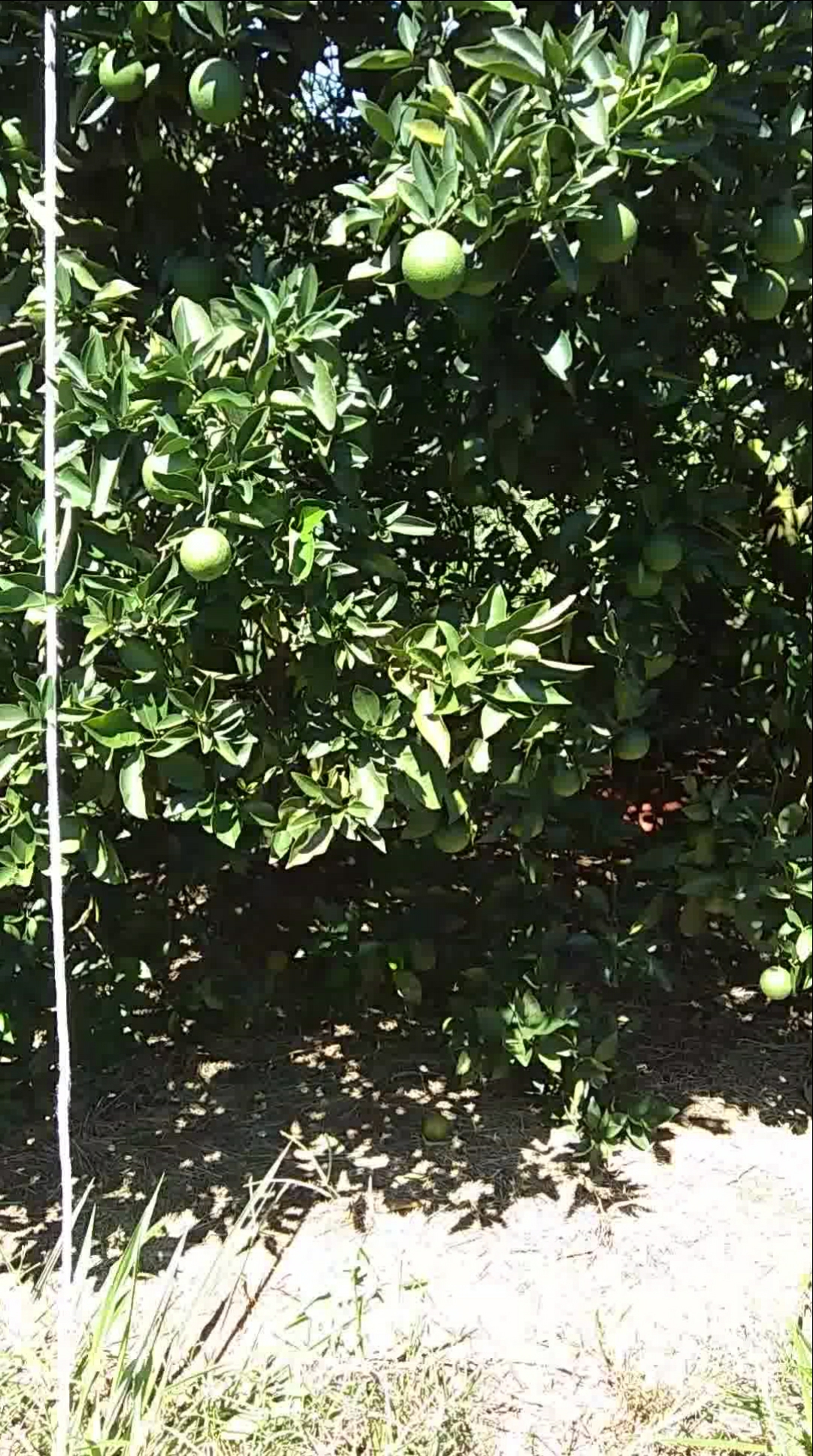}
         \caption{Frame 100}
         \label{fig:frame100}
     \end{subfigure}  
     \hfill
     \begin{subfigure}[b]{0.18\textwidth}
         \centering
         \includegraphics[width=\textwidth]{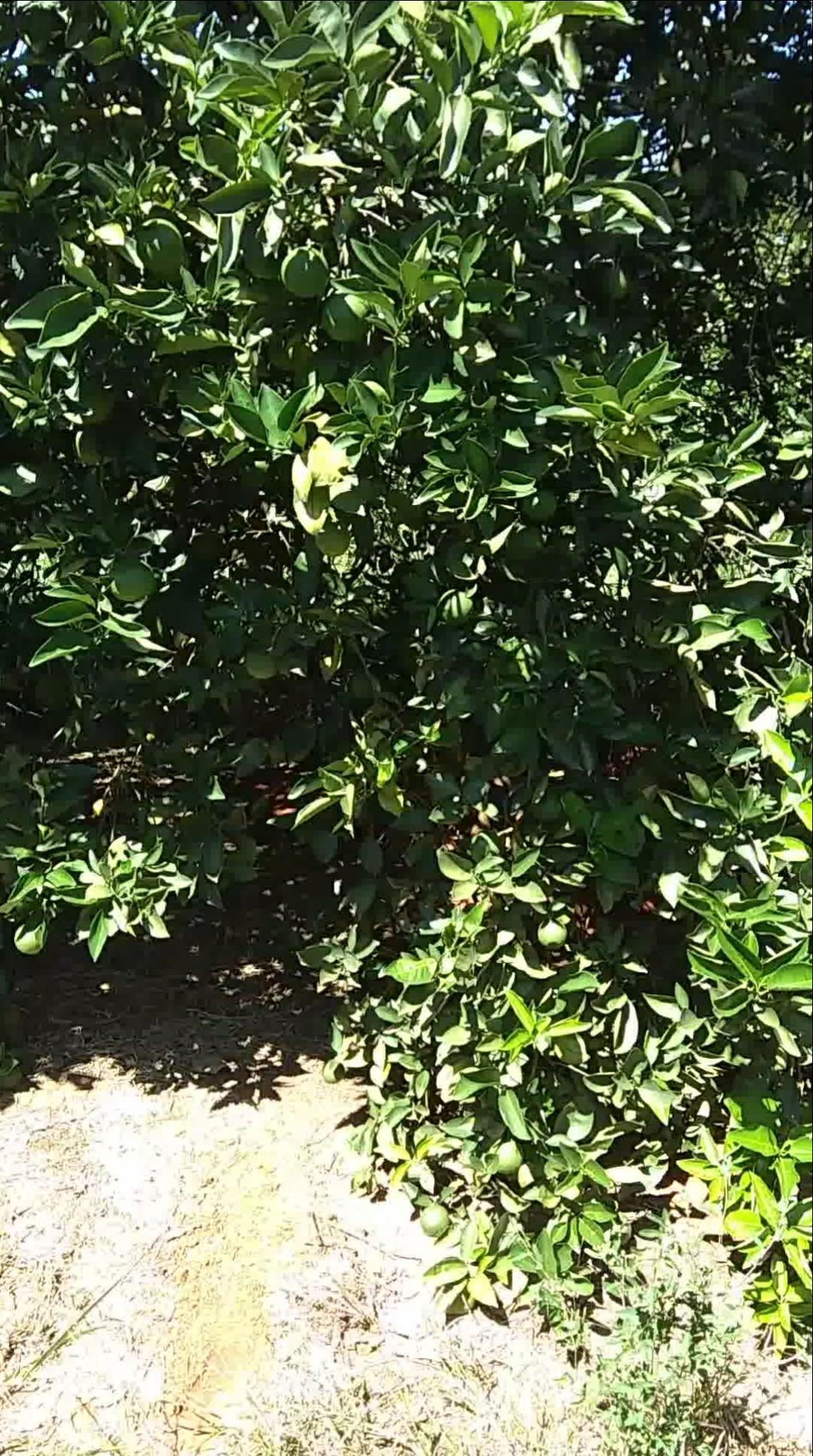}
         \caption{Frame 200}
         \label{fig:frame200}
     \end{subfigure}  
     \hfill
     \begin{subfigure}[b]{0.18\textwidth}
         \centering
         \includegraphics[width=\textwidth]{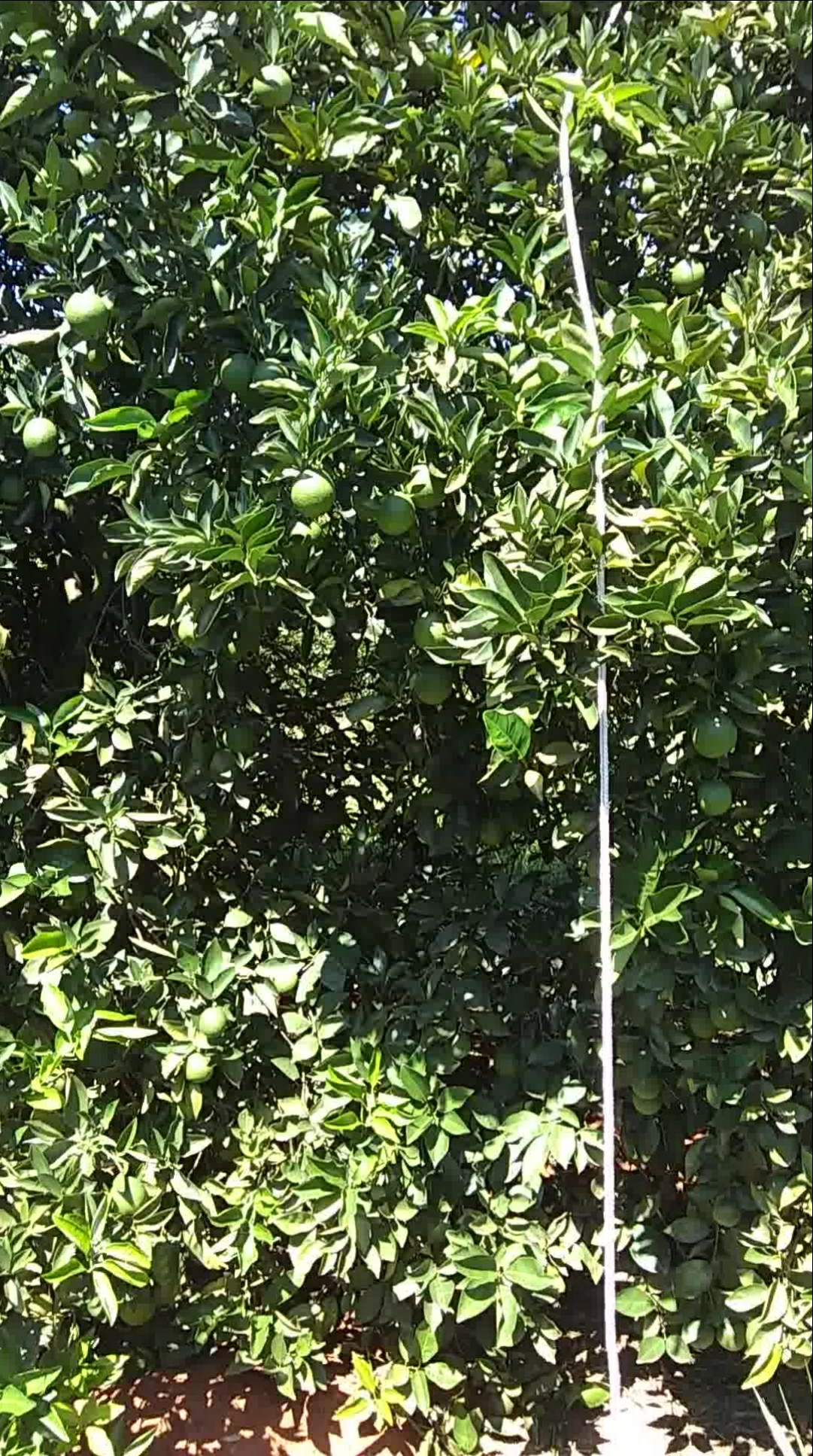}
         \caption{Frame 300}
         \label{fig:frame300}
     \end{subfigure}  
     \hfill
     \begin{subfigure}[b]{0.18\textwidth}
         \centering
         \includegraphics[width=\textwidth]{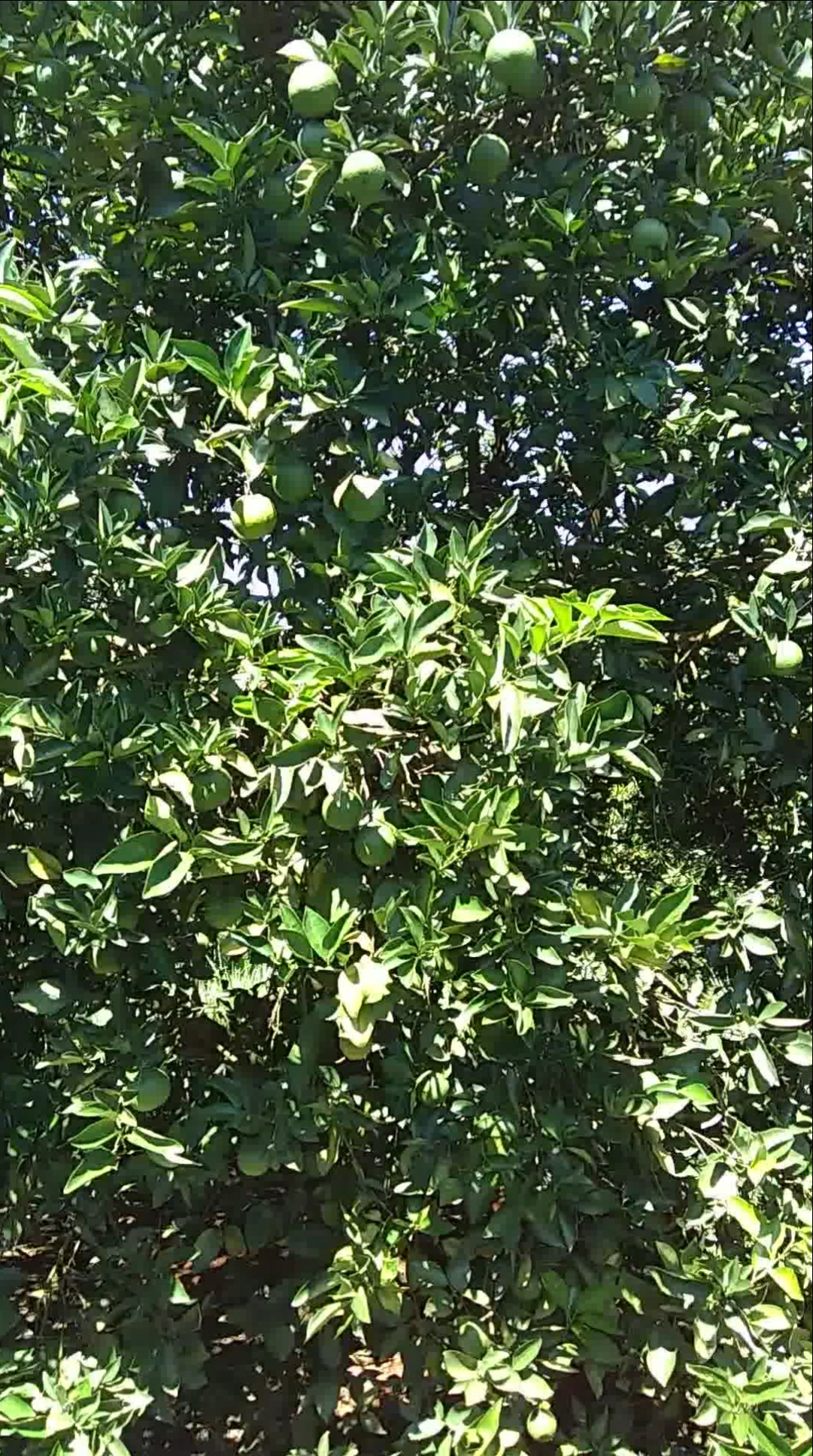}
         \caption{Frame 400}
         \label{fig:frame400}
     \end{subfigure}  
     \hfill
     \begin{subfigure}[b]{0.18\textwidth}
         \centering
         \includegraphics[width=\textwidth]{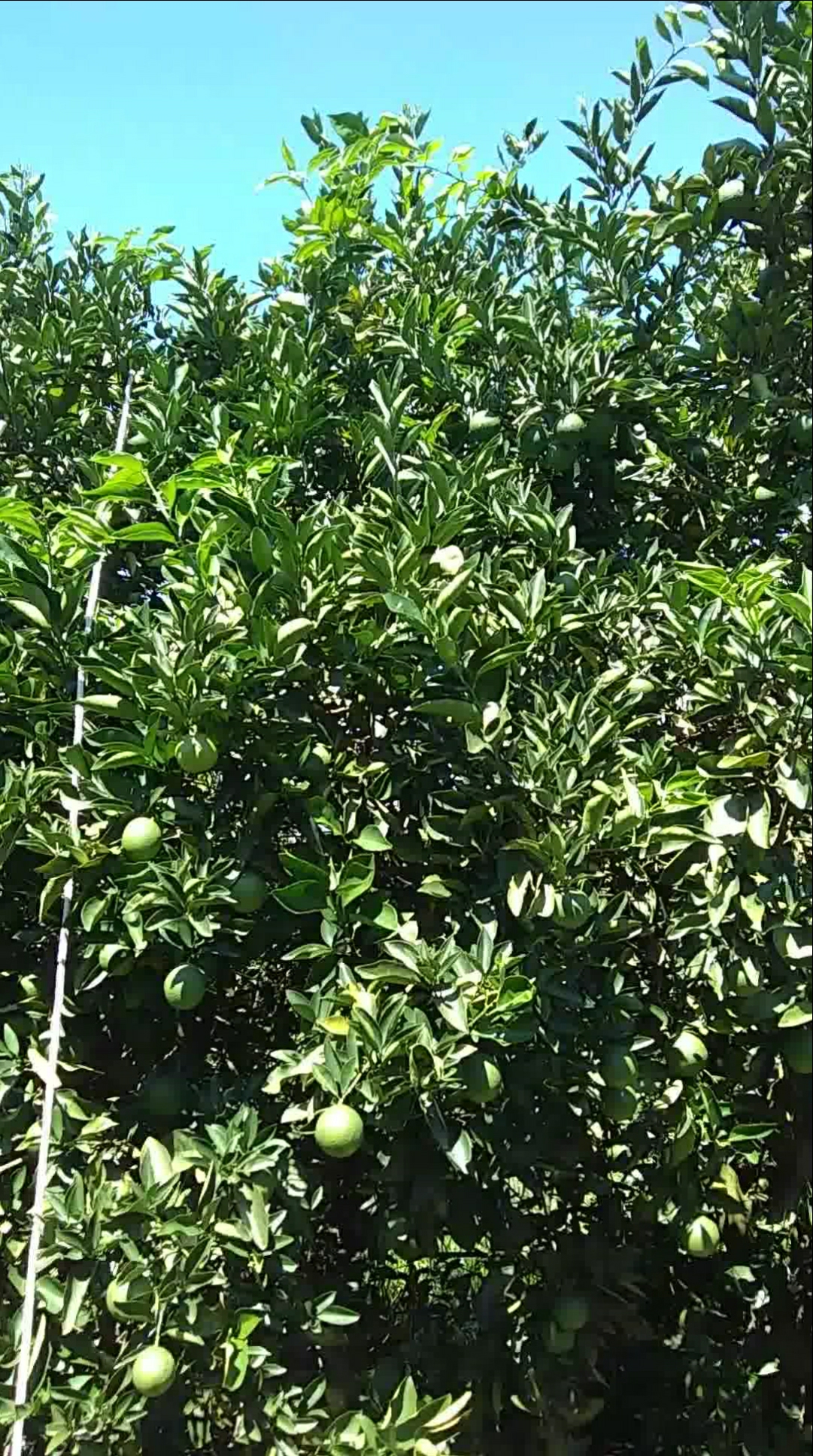}
         \caption{Frame 500}
         \label{fig:frame500}
     \end{subfigure}  
     \caption{The tree recording process. A smartphone is used to record a video, following the path 
     showed in red (a). The operator starts imaging the tree bottom, then following to the middle 
     and top of the canopy, changing the camera orientation as needed (b). A sample of five frames is 
     also shown (c--g).} 
     \label{fig:camera-path}
\end{figure}

The crop forecast staff was instructed to record video sequences for each tree before the manual fruit stripping. 
Smartphones were employed to record videos presenting a resolution of 1940 $\times$ 1080 
pixels. The staff was oriented to record the videos in portrait orientation, maximizing 
the field of view in the vertical direction. Considering the trees' density in commercial farms, it is not feasible to 
cover an entire tree canopy in a single pass, considering the employed smartphones do not present wide-angle lenses. So, for each 
tree in the sample, the forecast staff recorded a video for each side of that tree, i.e., the facade facing a row of the 
orchard. For each side, the staff member performed a trajectory, recording first the bottom of the tree, then moving to 
the central part and, finally, recording the top of the canopy, all contained in a single video take, as shown in 
Figure~\ref{fig:camera-path}. 

The staff works on a rigid schedule determined by the forecasting deadline, so the video sequences must be 
recorded under the available light conditions at that moment. If the light conditions are not ideal, for example, when the sun is 
behind the tree of interest, the staff member has to record the video anyway. This commonly produces frames presenting direct 
sunlight over the camera for the higher parts of the canopy. For the same reason, the video set presents diverse exposures, 
caused by different sunlight and shadows patterns in the trees.

\subsubsection{Frame sampling}\label{sec:framesampling}

For each video (single tree side), all frames are initially extracted using the 
FFmpeg library\footnote{FFmpeg -- \url{https://ffmpeg.org}.}, producing a sequence of frames 
$\mathcal{F} = \langle f_1, f_2, \ldots , f_{N} \rangle$ (see Figure~\ref{fig:overview}). However, not all the $N$ frames 
are needed for the fruit detection and tracking: similar neighboring frames can be removed to reduce processing time. The 
method employed for frame sampling is divided into two steps. Firstly, the ORB keypoint detector and descriptor 
\citep{rublee2011orb} is employed to extract features for each frame. After the first frame $f_1$ is added to the sample, 
the following procedure is iteratively repeated. Let $f_i$ be the last frame in $\mathcal{F}$ included in the sample. 
Descriptors of the frame $f_i$ are compared to those of a frame $f_j$ , $j \in [i + 1, N]$ using the FLANN algorithm 
\citep{muja2014flann}, that produces a set of matching features between the two frames. If the number of corresponding 
descriptors is below a threshold, that indicates significant differences between frames, and $f_j$ is included in the 
sample. Otherwise, the spatial coordinates are used to calculate the average distance between matched keypoints. 
If this average distance exceeds a defined threshold (10~pixels), $f_j$ is added to the sample. This process is repeated 
between the last selected frame and the subsequent frames until all the original $N$ frames are analyzed. After sampling, 
we have a sequence $\mathcal{F}' = \langle f_1, f_2, \ldots , f_{M} \rangle$, $M < N$ (the remaining frames are sequentially 
re-indexed). On average, this sampling procedure reduces the number of frames by 50 to 70 percent.

\subsubsection{Camera matrix estimation}\label{sec:egomotion}

Both our annotation tool and our fruit relocalization procedure rely on knowledge about the \emph{camera projective
matrix}. This matrix encodes all geometrical information needed to model the projective process, including the camera 
intrinsic parameters (focal distance, for example) and the extrinsic parameters: the camera position and orientation in 
the tridimensional space (ego-motion) \citep{hz2004mvg}. For each frame $f_i \in \mathcal{F}'$, we need a $3 \times 4$ 
projective matrix $\mathtt{P}_i$ such that:
\begin{equation}
  \mathbf{x}_i = \mathtt{P}_i \mathbf{X},
\label{eq:proj}   
\end{equation}
where $\mathbf{X}$ is a tridimensional point in the scene and $\mathbf{x}_i$ its projection on frame $f_i$. The 
$\mathtt{P}_i$ matrices can be estimated from image sequences by pose graph optimization (PGO), simultaneous 
localization and mapping (SLAM), 
structure from motion (SfM) \citep{dellaert2021factor} or visual-inertial navigation (VIN) \citep{huang2019vins}. In 
this work, we have employed structure from motion, using the COLMAP tool proposed by \citet{schonberger2016colmap}. 
Figure~\ref{fig:sfm} and Figure~\ref{fig:sfm-detail} show an example of the camera position and orientation recovered by 
COLMAP for a frame sequence in the dataset, and includes some tree \emph{structure}: several points $\mathbf{X}$ in 
the scene that acted as landmarks  in the SfM process \citep{dellaert2021factor}. The SfM results from COLMAP are also 
included in the public dataset.

\subsubsection{Annotation and the \textsc{MOrangeT} dataset for fruit tracking}\label{sec:annot}

As pointed by \citet{jong2022apple}, multiple fruit tracking involves \emph{homogeneous objects}. 
Identifying fruits that have been hidden from view for extended periods, either due to occlusions 
or being out of sight, is a challenging error-prone task, even for humans. Even when the fruit is
visible, registering a bounding box for an orange in each frame is a slow and tedious work. Considering 
these issues, we have developed a new annotation tool that exploits the spherical shape of oranges and
employs camera position information estimated by structure from motion.      

The developed tool lets the users in charge of annotations to draw square bounding boxes for an orange 
in a few frames. Using
the camera projective matrices $\mathtt{P}_i$ for each frame, the orange's center and ray are estimated
\emph{in the 3-D space}. The fruit is reprojected on \emph{all} frames (using Equation~\ref{eq:proj}), 
being the tool able to automatically check if the fruit is in the FOV of each frame and, in the positive 
cases, draw a proper bounding box. The user can adjust every bounding box in every frame and, 
if needed, re-estimate the spherical orange in 3-D and reproject it again, refining the fruit localization. 
Nonetheless, the user is required to search for instances of occlusion where the orange is obstructed by 
leaves, branches, or other 
fruits, even though it is within the camera's view. This method has made the process of annotating orange 
tracks faster and more reliable, especially when faced with challenging relocalization instances. The 
capability to modify the orange's ray in 3-D and then project it onto frames results in well-adjusted 
bounding boxes, which would have otherwise required laborious efforts from annotators. 

\begin{figure}
	\centering
	\includegraphics[width=\textwidth]{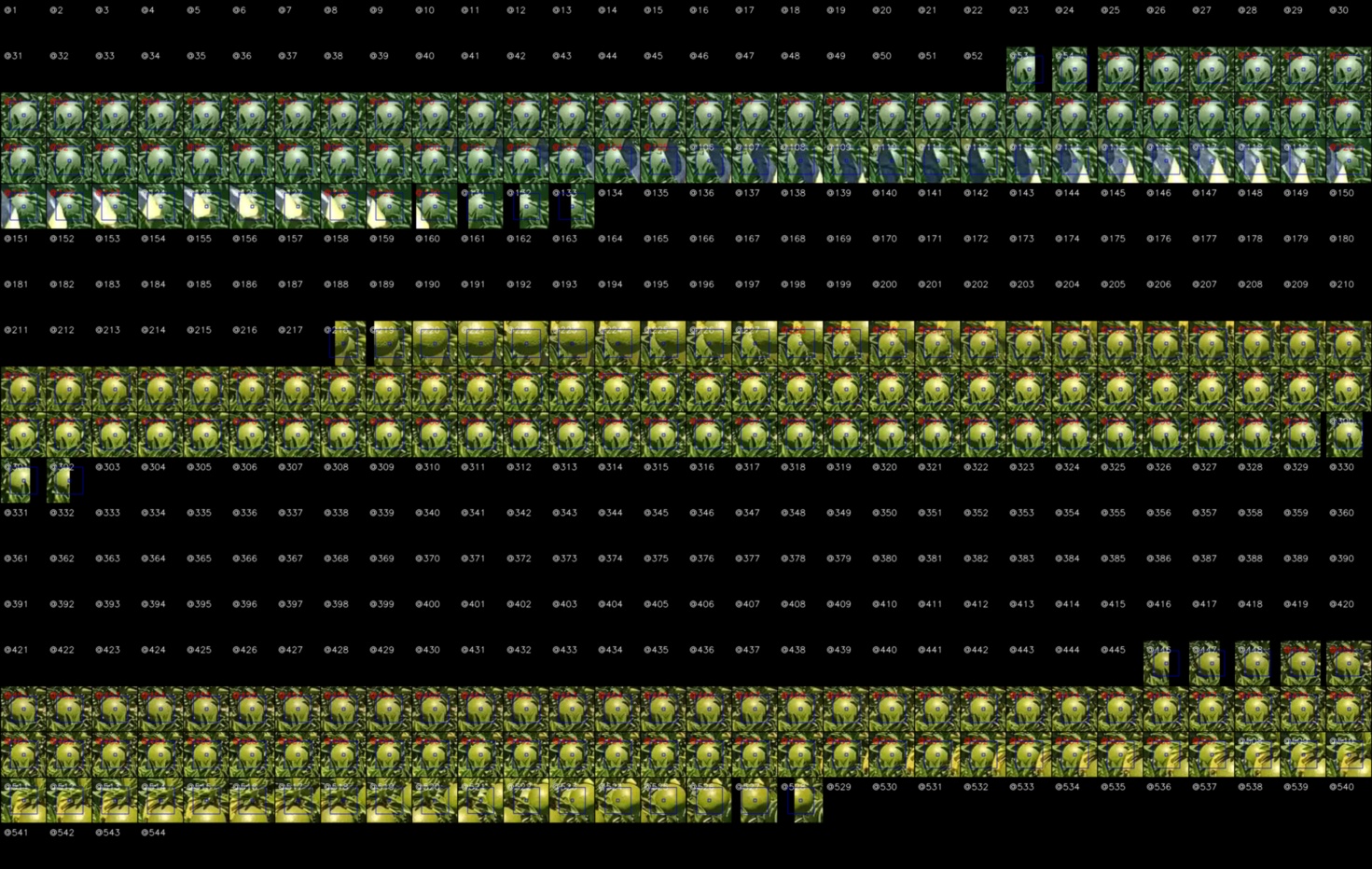}
	\caption{The mosaic window from the custom annotation tool for orange tracks annotation. The sequence 
	is composed by 544 images (frames), represented by 544 squares in the mosaic. When the orange is 
	in the FOV, a view extracted from the corresponding frame is displayed in the mosaic. White 
	marks mean occlusion, while the red ones means above 50\% visibility (non-occluded). Best seen in 
        digital format.}
	\label{fig:oreviewer}
\end{figure}

Figure~\ref{fig:oreviewer} shows the \emph{mosaic window} from the custom annotation tool. In 
this example, the tool is displaying the views of a single orange in a 544 frame long sequence. 
Each square in the mosaic corresponds to the view of that orange in the current frame, including 
part of its neighborhood. The orange’s center and bounding box are displayed in blue. Note the 
same orange enters and exits the field of view \emph{three times}, entering at frames 53, 218 
and 446. Furthermore, three occluded segments are observed: between frames 106--119, 218--227 
(occlusion by other orange) and 508--528. We have adopted two conventions: if more than 50\% 
of the fruit is visible, it is considered non-occluded (marked in red in 
Figure~\ref{fig:oreviewer}). Otherwise, the orange is considered occluded (marked in white). 
Moreover, if any part of the bounding box is out FOV, the orange is considered non-visible, 
to avoid unfairly penalizing trackers in borderline cases.

\begin{table}
	\caption{\textsc{MOrangeT} frame sequences. Each sequence corresponds to one side of one tree from five
	different regions in São Paulo state: Limeira (LIM), Duartina (DUA), Porto Ferreira (PFE), 
	Brotas (BRO) and Avaré (AVA).}
	\centering
	\begin{tabular}{llllcrrr}
\toprule
Sequence & Gimble & Region & Variety  & Planting year & Height (cm) & Num. frames & Num. oranges \\
\midrule
V01      & No     & LIM    & {\it Pera}     & 2007          & 345         & 408              & 110               \\
V02      & Yes    & LIM    & {\it Pera}     & 2012          & 250         & 409              & 46                \\
V03      & Yes    & DUA    & {\it Valencia} & 2005          & 490         & 634              & 90                \\
V04      & No     & DUA    & {\it Valencia} & 2012          & 410         & 384              & 105               \\
V05      & No     & PFE    & {\it Natal}    & 2017          & 200         & 293              & 148               \\
V06      & Yes    & BRO    & {\it Valencia} & 2009          & 400         & 447              & 122               \\
V07      & Yes    & BRO    & {\it Valencia} & 2006          & 414         & 553              & 154               \\
V08      & Yes    & DUA    & {\it Valencia} & 2013          & 320         & 544              & 192               \\
V09      & Yes    & AVA    & {\it Hamlin}   & 2019          & 183         & 222              & 16                \\
V10      & Yes    & AVA    & {\it Hamlin}   & 2019          & 183         & 244              & 13                \\
V11      & Yes    & AVA    & {\it Hamlin}   & 2008          & 380         & 268              & 87                \\
V12      & Yes    & AVA    & {\it Hamlin}   & 2008          & 380         & 307              & 115               \\
\bottomrule
\end{tabular}

	\label{tab:datainfo}
\end{table}

The \textsc{MOrangeT} \citep{dataMOrangeT} dataset is composed of 12 sequences, as shown in Table~\ref{tab:datainfo}. Nine of them were 
recorded by smartphones connected to a gimbal for stabilization (Zhiyun Smooth 4, Guilin Zhishen Information 
Technology Co.). The set included trees of four orange varieties, {\it Valencia}, {\it Natal}, {\it Pera}, and 
{\it Hamlin}, presenting different ages and heights, from four different regions in the São Paulo citrus belt. 
Although some mature fruits, most of the oranges are green: the crop forecast program needs to perform the 
assessments at an early stage of the fruits' development. Note that the \textsc{MOrangeT} dataset is employed to evaluate
multiple fruit tracking and the counting of visible fruits in videos of single trees (single side), while the Orange Crop 
Forecast data is used for the evaluation of yield prediction for the 1,543 trees set.

\subsubsection{\textsc{OranDet} dataset for fruit detection}

Utilizing the camera's pose information enables the annotation of an extensive array of bounding boxes within the frames of 
videos in the \textsc{MOrangeT} dataset. We have selected four sequences, V04, V05, V07 and V11 (Table~\ref{tab:datainfo}) 
to compose a training set for orange detection. Each frame underwent a process of extracting a set of tiles, each measuring 
$416 \times 416$ pixels, which were then incorporated into the training set. Given the relatively diminutive size of fruits 
compared to frame dimensions, the application of tiling emerged as a straightforward yet efficacious strategy to mitigate the 
risk of overlooking small objects. This process resulted in a compilation of 21,031 tiles extracted from these frame sequences, 
with an additional 3,065 tiles incorporated from a previous study on orange detection \citep{camargo2019uso}. Subsequently, the 
dataset was partitioned into training and validation subsets. The eight remaining frame sequences were exclusively designated 
for a test set, comprising 33,716 tiles. Table~\ref{tab:datainfo:orandet} outlines the definitive composition of the orange detection 
dataset, denominated as \textsc{OranDet}, with illustrative examples depicted in Figure 4.

\begin{table}
	\caption{\textsc{OranDet}, a dataset for orange detection. The set includes more than 220,000 bounding boxes annotated
 for 57,812 images (tiles).}
	\centering
	\begin{tabular}{lrr}
\toprule
 & Num. images & Num. bounding boxes\\
\midrule
Training & 20,662 & 89,318 \\
Validation & 3,434 & 11,198\\
Test & 33,716 & 121,685 \\ \hline
Total & 57,812 & 222,201\\
\bottomrule
\end{tabular}
	\label{tab:datainfo:orandet}
\end{table}

Within the context of the pipeline illustrated in Figure~\ref{fig:overview}, the \textsc{OranDet} dataset \citep{dataOranDet} was instrumental 
in both training and evaluating CNN-based fruit detection. Our orange tracker, depicted in Figure~\ref{fig:overview}~(f) 
and expounded upon in Section 3.2.2, does not use any learning-based component. Consequently, the \textsc{MOrangeT} dataset was 
exclusively employed to \emph{evaluate} the tracking performance. Finally, the yield regression employs the Orange Crop Forecast 
data and the fruit counting from tracking to train and evaluate the yield predictions, as seen in Figure~\ref{fig:overview}~(g).

\begin{figure}[!htb]
     \centering
     \includegraphics[width=\textwidth]{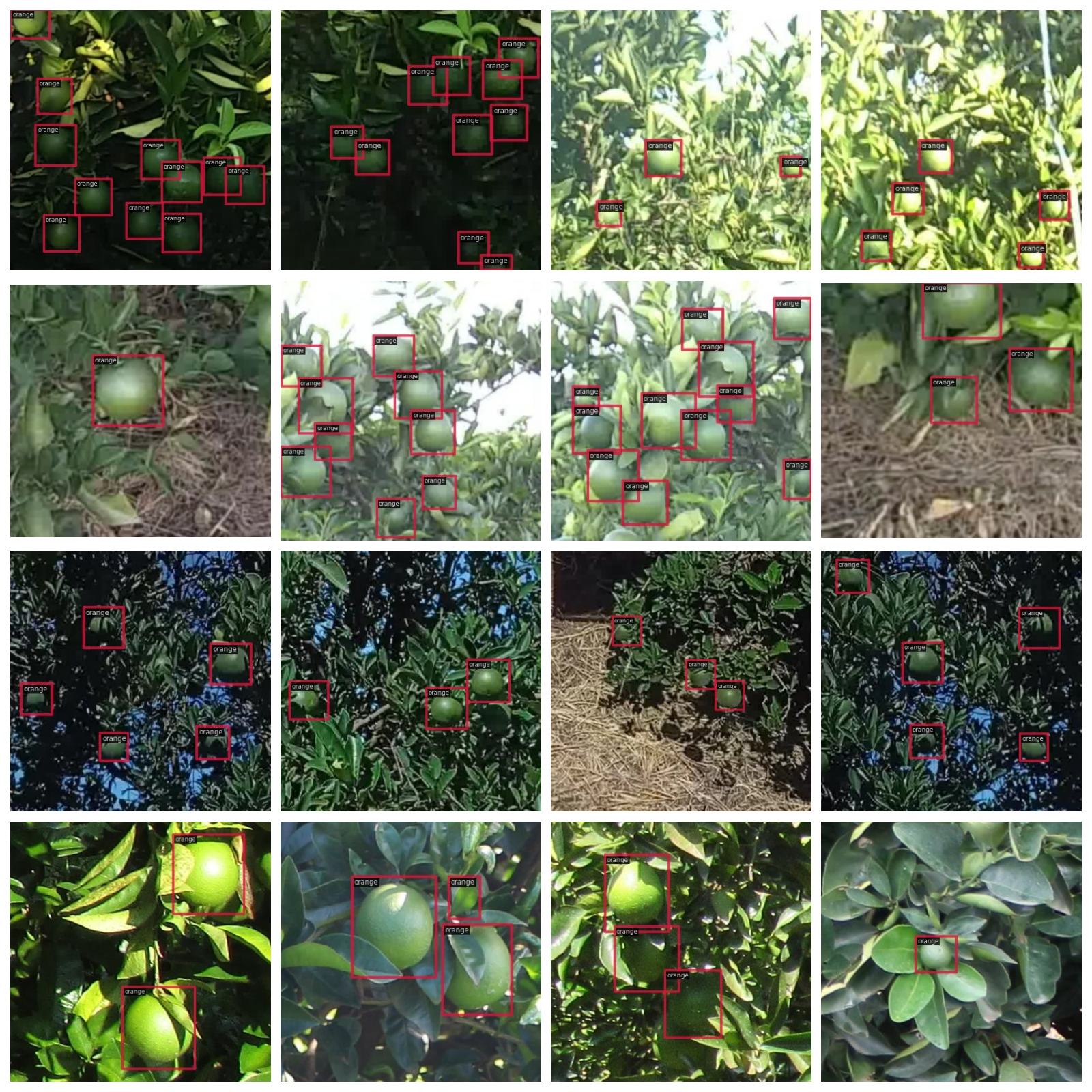}
     \caption{Examples in the \textsc{OranDet} training set. Images present immature green fruits. Note the
     variety in light, shadows, size, and blur.} 
     \label{fig:orandet-examples}
\end{figure}

\subsection{Orange detection} \label{sec:det}

We trained and evaluated six different architectures for orange detection: YOLOv3 \citep{redmon2018yolov3}, 
YOLOv6 \citep{li2022yolov6}, YOLOv7 \citep{wang2022yolov7}, EfficientDet \citep{tan2020effdet}, and the YOLOv5 and 
YOLOv8 models proposed by Ultralytics\footnote{See \url{https://docs.ultralytics.com/models}.}. Nowadays, there are 
several neural network architectures for object detection, based on CNNs \citep{arkin2023survey} and transformers \citep{shehzadi2023object}, 
and the selection of architectures is not straightforward. Considering the extensive use of YOLO-based networks in 
fruit detection \citep{tian2019apple, wang2019mango}, specially in 
citrus detection \citep{camargo2019uso, mirhaji2021fruit, xu2023real}, we have selected YOLOv3, the last architecture
proposed by \citet{redmon2018yolov3} (the authors that introduced YOLO), and a few of recent YOLO-based approaches. 
To provide a comparative analysis, we have opted for EfficientDet as an alternative approach. 

We used the implementations available in MMYOLO\footnote{\url{https://github.com/open-mmlab/mmyolo}} and MMDetection \citep{mmdetection}.
All models were trained using stochastic gradient descent (SGD) optimizer with momentum. Random flip augmentation was applied in the training 
pipeline for all models, and the Mosaic augmentation, introduced after YOLOv3, was used during the training of YOLOv5, YOLOv6, YOLOv7, 
and YOLOv8.

\subsection{Multiple orange tracking}\label{sec:mot}

Our multiple fruit tracking method gets as input the bounding boxes set $\mathcal{B}_i$ from the orange detector and the 
projective matrices $\mathtt{P}_i$ for each frame $f_i$ in the frame sequence $\mathcal{F}_t'$ (see 
Figure~\ref{fig:overview}). The method is based on a few components, described as follows.

\subsubsection{Bounding box assignments}

An essential component of the method is data association: \emph{bounding box assignment} is performed between two box sets, matching boxes in a set $\mathcal{B}_i$ to boxes in another set $\mathcal{B}_{j}$. Such association is performed by a variant of the Hungarian algorithm \cite{kuhn1955hungarian}, using a cost function based on intersection over union (IoU):
\begin{equation}
    \mathrm{cost}(\mathbf{b}_{i}^{(p)}, \mathbf{b}_{j}^{(q)}) = 1 - \mathrm{IoU}(\mathbf{b}_{i}^{(p)}, \mathbf{b}_{j}^{(q)}),
\end{equation}
where $\mathbf{b}_{i}^{(p)} \in \mathcal{B}_i$, $\mathbf{b}_{j}^{(q)} \in \mathcal{B}_j$, and IoU is 
the \emph{Jaccard index} defined over the intersection area and union area between $\mathbf{b}_{i}^{(p)}$ and $\mathbf{b}_{j}^{(q)}$:
\begin{equation}
\mathrm{IoU}(\mathbf{b}_{i}^{(p)}, \mathbf{b}_{j}^{(q)}) = J(\mathbf{b}_{i}^{(p)}, \mathbf{b}_{j}^{(q)}) = \frac{|\mathbf{b}_{i}^{(p)} \cap \mathbf{b}_{j}^{(q)}|}{|\mathbf{b}_{i}^{(p)} \cup \mathbf{b}_{j}^{(q)}|}.
\end{equation}
Considering $P$ bounding boxes in $\mathcal{B}_i$ and $Q$ boxes in $\mathcal{B}_j$, the assignment matrix 
$\mathtt{A}$ is a $P \times Q$ matrix where $\mathtt{A}[p,q] = 1$ iff $\mathbf{b}_{i}^{(p)}$ and $\mathbf{b}_{j}^{(q)}$ 
are associated, otherwise $\mathtt{A}[p,q] = 0$. Assume, without loss of generality, that $P \leq Q$. The association problem is a \emph{2-D rectangular assignment} minimization problem \citep{crouse2016assign}:
\begin{equation}\label{eq:hungarian}
  \mathtt{A}^* = \arg \min_{\mathtt{A}} \sum_{p = 1}^{P} \sum_{q = 1}^{Q} \mathrm{cost}(\mathbf{b}_{i}^{(p)}, \mathbf{b}_{j}^{(q)}) \mathtt{A}[p, q],
\end{equation}
subject to:
\begin{equation}\label{eq:hungcond1}
    \sum_{q=1}^{Q} \mathtt{A}[p,q] = 1, \forall p
\end{equation}
\begin{equation}\label{eq:hungcond2}
    \sum_{p=1}^{P} \mathtt{A}[p,q] \leq 1, \forall q.
\end{equation}
Note that $\mathbf{b}_i^{(p)}$ refers to the $p$-th bounding box in $\mathcal{B}_i$ and $\mathbf{b}_{j}^{(q)}$ refers 
to the $q$-th bounding  box in $\mathcal{B}_{j}$. For this minimization, we have employed a modified 
Jonker-Volgenant algorithm with no initialization \citep{crouse2016assign}, a $O(n^3)$ variant of the Hungarian algorithm. 
This implementation supports \emph{unbalanced assignments}, so working in cases where $P \neq Q$: boxes in $i$ not assigned 
(lost boxes) or boxes in $j$ not assigned (novel boxes). As a post-processing step, we remove assignments where 
$\mathrm{IoU}(\mathbf{b}_{i}^{(p)}, \mathbf{b}_{j}^{(q)}) = 0$ eventually produced by the minimization algorithm. The assignment
component is repeatedly employed in this work:
\begin{itemize}
\item to perform data association between box sets $\mathcal{B}_i$ and $\mathcal{B}_{i+1}$, corresponding to detected fruits in two neighboring frames $f_i$ and $f_{i+1};$
\item to perform relocalization, associating reprojected bounding boxes of previously seen oranges to boxes in $\mathcal{B}_i$, detected in the current frame $f_i$, and
\item to evaluate results, matching detected boxes to the ground-truth annotation set $\hat{\mathcal{B}}_i$ for frame $f_i$.
\end{itemize}

\subsubsection{Orange 3-D localization}

The association procedure, when applied to bounding boxes sets from neighboring 
frames $f_i$ and $f_{i+1}$, produces \emph{contiguous tracks} in the form 
$\mathcal{T} = \langle \mathbf{b}_i, \mathbf{b}_{i+1}, ... , \mathbf{b}_{i+n}\rangle$ 
(here the upper script indexes are omitted to simplify the notation). When a contiguous
track is sufficiently long (we have adopted $n \geq 5$), we employ the camera matrices 
$\mathtt{P}_i, \mathtt{P}_{i+1},..., \mathtt{P}_{i+n}$ to estimate the three-dimensional position of the 
corresponding orange. Algorithm~\ref{alg:ransac} takes as input a track $\mathcal{T}$ and the
corresponding sequence of camera matrices $\mathcal{P}$ to estimate the center of the orange in the 3-D space, 
$\mathbf{X}$. The algorithm is based on two concepts: \emph{reprojection error} and \emph{random sample consensus}. 
The reprojection error is the distance between the bounding box center \cbi (line~4) and $\mathbf{x}_i$, the reprojection of \Xt  
on the $i$-th frame (Equation~\ref{eq:proj}). A bounding box is considered an \emph{inlier} iff this distance is within a 
threshold, the maximum reprojection error allowed (lines 15--16). The RANdom SAmple Consensus (RANSAC) is a general 
framework for robust estimation algorithms proposed by \citet{fischler1981ransac}: (i) a sample \Ss is randomly selected
to estimate the target; (ii) the points in the full set within a distance threshold are defined as inliers (consensus); 
(iii) the target is re-estimated using all inliers, and (iv) the process is iterated if the number of inliers is 
insufficient. In Algorithm~\ref{alg:ransac}, our sample \Ss is a combination of three bounding boxes in the track (line~6)
and a target $\mathbf{X}_\mathcal{S}$ is estimated using the \emph{direct linear triangulation} procedure 
(DLT) \citep{hz2004mvg} (line~7). The reprojection is employed to define the set of inliers (lines 12--19) and, 
if the number of inliers is sufficient (line~20), \Xt is estimated from the consensus set, again using DLT. However, 
considering consensus could not be achieved and the number of combinations of three 
boxes (line~6) can be massive, a maximum number of iterations is defined (lines~24--27). If the maximum number of 
iterations is reached or if all samples were considered with no consensus, the algorithm returns \Xt as \textsc{Nil} and 
an empty inliers set (no success). In Algorithm~\ref{alg:ransac}, \textsc{Indexes}($\cdot$) returns the frame 
indexes $i$ for the bounding boxes \bbi in the set (in a valid track, there is only one bounding box per frame). 

\begin{algorithm}
\caption{RANSAC-based triangulation}\label{alg:ransac}
\begin{algorithmic}[1]
\Procedure{RansacTriangulation}{$\mathcal{T}, \mathcal{P}$}
\State $\mathrm{n\_iters} \gets 1$
\ForAll{$\mathbf{b}_i \in \mathcal{T}$}
    
    \State $\mathbf{c}_i \gets \left(\frac{\mathbf{b}_i.x + \mathbf{b}_i.\mathrm{width}}{2}, \frac{\mathbf{b}_i.y + \mathbf{b}_i.\mathrm{height}}{2}\right)^\intercal$ \Comment{Bounding box centroid}
\EndFor
\ForAll{$\mathcal{S} \gets \Call{Combinations}{\mathcal{T}, 3}$}    \Comment{RANSAC loop}
    \State $\mathbf{X}_{\mathcal{S}} \gets \Call{DltTriangulation}{(\mathbf{b}_j, \mathtt{P}_j), j \in \mathrm{\textsc{Indexes}}(\mathcal{S})}$ \Comment{Direct Linear Transformation (DLT)}
    \If{$\mathbf{X}_{\mathcal{S}}$ is \textsc{Nil}}
      \State \textbf{continue}
    \EndIf
    \State $\mathcal{I} \gets \emptyset$
    \ForAll{$j \in \Call{Indexes}{\mathcal{T}}$}
        \State $\mathbf{x}_j \gets \mathtt{P}_j \cdot \mathbf{X}_{\mathcal{S}}$ \Comment{Homogeneous form: $\mathbf{x}_j = (x, y, z)^\intercal$}
        \State $\mathbf{x}_j \gets (x/z, y/z)^\intercal$
        \State $\mathrm{geom\_error} \gets \| \mathbf{x}_j - \mathbf{c}_j\|$ \Comment{Reprojection error}
        \If{$\mathrm{geom\_error} \leq \mathrm{max\_geom\_error}$}
            \State $\mathcal{I} \gets \mathcal{I} \cup \{\mathbf{b}_j\}$
        \EndIf
    \EndFor
    \If{$\frac{|\mathcal{I}|}{|\mathcal{T}|} \geq \mathrm{inliers\_ratio}$}
        \State $\mathbf{X} \gets \Call{DltTriangulation}{(\mathbf{b}_j, \mathtt{P}_j), j \in \mathrm{\textsc{Indexes}}(\mathcal{I})}$ \Comment{Re-estimate $\mathbf{X}$ using \emph{all} inliers}
        \State \textbf{return} $\mathbf{X}, \mathcal{I}$
    \EndIf
    \State $\mathrm{n\_iters} \gets \mathrm{n\_iters} + 1$
    \If{$\mathrm{n\_iters} > \mathrm{max\_iters}$} \Comment{Exceeds the maximum number of iterations allowed}
        \State \textbf{return} \textsc{Nil}, $\emptyset$
    \EndIf
\EndFor
\State \textbf{return} \textsc{Nil}, $\emptyset$ \Comment{It could not find an acceptable estimation}
\EndProcedure
\end{algorithmic}
\end{algorithm}

Algorithm~\ref{alg:orange_estimation} models the orange as a sphere and estimates the fruit's center \Xt and ray 
$r_\mathrm{3D}$. The three-dimensional point \Xt and the inliers bounding boxes set $\mathcal{I}$ are computed using 
Algorithm~\ref{alg:ransac}. Then the ray is estimated using the dimensions of the boxes in $\mathcal{I}$ and 
similar triangles' ratio: 
\begin{equation}\label{eq:r3d}
r_\mathrm{3D} \approx \frac{r_i d }{f_\mathrm{focal}}
\end{equation}
where $d$ is the distance between the camera center position $\mathbf{C}_i$ at frame $f_i$ and 
the orange center \Xt (line~10), $f_\mathrm{focal}$ is the camera focal distance, and $r_i$ the 2-D ray 
estimated from the bounding box \bbi dimensions (line~8). Each $r_i$ provides an estimation for $r_\mathrm{3D}$ (line~11), 
and we use the median of the estimations set $\mathcal{R}$ as our final estimation for the orange three-dimensional ray. 
We have noted that the bounding boxes usually exceed the fruit boundaries\footnote{Note also Equation~\ref{eq:r3d} is an approximation: even in the absence of noise, the equality would just hold if the sphere center was perfectly projected 
in the image's principal point \citep{hz2004mvg}.}, overestimating the ray, then we multiply
the median by a constant $c$ for a better fit (line~13, we have adopted $c = 0.9$). The camera center $\mathbf{C}_i$ is
easily computed from \Pji (line~9) using:
\begin{equation}\label{eq:simtrig}
    \mathbf{C}_i = -\mathtt{M}^{-1}\mathbf{p}_4,
\end{equation}
where $\mathtt{P}_i = [\mathtt{M} \,|\, \mathbf{p}_4]$, i.e., the $\mathtt{M}$ matrix is composed by the three first columns of \Pji and $\mathbf{p}_4$ is the last column of $\mathtt{P}_i$.

\begin{algorithm}
\caption{Estimate orange's position and ray in 3-D}\label{alg:orange_estimation}
\begin{algorithmic}[1]
\Procedure{EstimateOrange}{$\mathcal{T}, \mathcal{P}$}
\State $\mathbf{X}, \mathcal{I}\gets$ \Call{RansacTriangulation}{$\mathcal{T}, \mathcal{P}$} \Comment{$\mathcal{I}$ is the set of inliers}
\If{$\mathbf{X}$ is \textsc{Nil}
    \State \textbf{return} \textsc{Nil}, \textsc{Nil}
\EndIf
\State $\mathcal{R} \gets \emptyset$
\ForAll{$\mathbf{b}_i \in \mathcal{I}$} }
    \State $r_i \gets \max(\mathbf{b}_i.\mathrm{height}, \mathbf{b}_i.\mathrm{width})/2$ \Comment{Estimate 2-D ray from bounding box dimensions}
    \State $\mathbf{C}_i \gets$ \Call{GetCameraCenter}{$\mathtt{P}_i$} \Comment{Camera projection center}
    \State $d \gets \| \mathbf{X} - \mathbf{C}_i\|$ \Comment{Orange to camera distance}
    \State $\mathcal{R} \gets \mathcal{R} \cup \left\{\frac{d \cdot r_i}{f_\mathrm{focal}}\right\}$
\EndFor
\State $r_{\mathrm{3D}} \gets c \cdot \mathrm{median}(\mathcal{R})$
\State \textbf{return} $\mathbf{X}, r_{\mathrm{3D}}$ 
\EndProcedure
\end{algorithmic}
\end{algorithm}

In our MOT tracking for oranges, tracks can be \emph{discontinuous}, meaning that the fruit is not visible at some 
frames because it is occluded or out of view. Note that Algorithm~\ref{alg:ransac} and 
Algorithm~\ref{alg:orange_estimation} have no need for continuous tracks: they can operate on discontinuous ones as new
bounding boxes are added to the tracks. Tracks can present one of two possible states in our tracking system: 
\textsc{Lost}, if the track has no bounding box in the current frame $f_{i}$, and \textsc{Active}, if the track presents 
an observed box $\mathbf{b}_{i}$. When processing the current frame $f_i$, our system starts performing \emph{relocalization},
using the Hungarian association to match \textsc{Lost} tracks to bounding boxes in $\mathcal{B}_{i}$ yet not associated to 
any track. 

\subsubsection{Relocalization}

The \Lost tracks have their fruits' position and ray estimated by 
Algorithm~\ref{alg:orange_estimation}: a center \Xt and a ray $r_\mathrm{3D}$ are available for each track. Using Equation~\ref{eq:proj}, 
we find the reprojection of the orange's center $\tilde{\mathbf{x}}_i = \mathtt{P}_i \mathbf{X}$ at frame $f_i$. The ray 
$r_\mathrm{3D}$ and Equation~\ref{eq:simtrig} are employed to find the ray $\tilde{r}_i$ for the orange on $f_i$. Finally, the reprojected 
bounding box is defined as:
\[
\tilde{\mathbf{b}}_i = [\tilde{x}_i - \tilde{r}_i, \tilde{y}_i - \tilde{r}_i, 2\tilde{r}_i, 2\tilde{r}_i].
\]
where $\tilde{\mathbf{x}}_i = (\tilde{x}_i, \tilde{y}_i, 1)^\intercal$ (homogenous coordinates) and $\tilde{\mathbf{b}}_i$
is the \emph{reprojected bounding box} for the track in frame $f_i$ (the $\tilde{\cdot}$ notation is used to indicate the values came
from the orange \emph{estimation}, not from orange \emph{detection}). Such boxes are computed for all \Lost tracks, producing a set of boxes $\tilde{\mathcal{B}}_i$. The Hungarian association (Equation~\ref{eq:hungarian}) is employed to
match boxes from \Lost tracks in $\tilde{\mathcal{B}}_i$ to boxes in $\mathcal{B}_i$ yet not matched to any track. In the 
case of a track is successfully associated to a box, its state is changed to \Active.

\subsubsection{Next frame association} 

After relocalization, the tracking system will employ the Hungarian association to match boxes 
$\mathbf{b}_i^{(p)} \in \mathcal{B}_i$ to boxes $\mathbf{b}_{i+1}^{(q)} \in \mathcal{B}_{i+1}$. If 
the association algorithm matches $\mathbf{b}_i^{(p)} \rightarrow \mathbf{b}_{i+1}^{(q)}$, and $\mathbf{b}_i^{(p)}$
belongs to an \textsc{Active} track $\mathcal{T}$, the box $\mathbf{b}_{i+1}^{(q)}$ is appended to $\mathcal{T}$. 
Otherwise, a novel track $\mathcal{T}' = \langle \mathbf{b}_i^{(p)}, \mathbf{b}_{i+1}^{(q)} \rangle$ is created. 
After association, tracks not matched to any box $\mathbf{b}_{i+1}^{(q)}$ change to the \textsc{Lost} state, 
while all the remaining \textsc{Active} tracks presenting more than 5 bounding boxes have their oranges re-estimated using 
Algorithm~\ref{alg:orange_estimation}. The entire procedure (relocalization followed by next frame association) is repeated 
for each frame until $f_{M-1}$.

In summary, contiguous \Active tracks presenting at least 5 boxes initialize 3-D spherical models for oranges, 
parameterized by central point and ray. After this initialization, the fruit is passive of relocalization: if the bounding boxes'
association fails by any reason, producing \Lost tracks, the reprojection of the spherical model on the current frame can be
employed to relocalize the fruit, associating the reprojected box to a detected one. After relocalization, the track is \Active 
again and can be updated with new detected boxes. Such track is now discontinuous, registering where (bounding box) and when 
(frame) the fruit is visible. Furthermore, the orange model (central point and ray) is continuously refined as new detections are 
added to the track. Only the tracks with successfully estimated 3-D orange models are considered for fruit counting.

\subsubsection{Implementation}

The tracker was implemented in Python~3. The 2-D rectangular assignment implementation employed is
from SciPy's optimization module \citep{crouse2016assign}. The DLT algorithm was implemented using singular value 
decomposition \citep{golub2013matrix} available in SciPy's linear algebra module.

\subsubsection{Tracking evaluation}

\citet{villacres2023apple} presented a \emph{sensitivity analysis} aimed to evaluate the performance of a fruit tracker under
varying probabilities of fruit detection. The analysis begins with perfect detection, from the ground-truth data, where all 
fruits are identified accurately, and then progressively eliminates certain bounding boxes. Each bounding box in the ground 
truth is treated as a random variable with a uniform distribution within the range $[0, 1]$ for each bounding box in 
$f_i$. Based on the value, each bounding box is kept or removed from the detections set $\mathcal{B}_i$. Specifically, 
if the random value of a fruit is below or equal to a pre-defined threshold, it is considered detected; otherwise, it 
is discarded. Similarly to Villacrés et al., in the present work we evaluated different values of the probability of detection, 
using 0.4, 0.6, 0.8, and 1 as thresholds (a threshold of 1 equals to ``perfect detection'', a copy of ground-truth bounding 
box data). The tracker is also evaluated using the detections from the CNN-based orange detection module presented in 
Section~\ref{sec:det}. The evaluation present values for HOTA \citep{luiten2021hota} and MOTA \citep{bernardin2008evaluating}. 
We also include in the evaluation two components of HOTA: DetA, the percentage of aligning detections, and AssA, the average 
alignment between matched trajectories. Appendix~\ref{apx:hota} presents mathematical definitions for HOTA, DetA and AssA, 
adapting the original \citet{luiten2021hota} formulation to the notation adopted in this work.

\subsection{Yield regressor}\label{sec:yieldreg}
\label{subsec:regressor}
After the fruit detection and tracking stages, we have all the fruits in a tree identified and mapped in 3-D. However, some fruits may have not been identified or tracked due to problems in video acquisition or because they are located inside the canopy. To address  this problem, we propose a regressor that takes the number of fruits on each side of the plant as input and estimates the actual number of fruits on that plant. This estimator can be based on statistics, as proposed in \citep{koirala2019mangoyolo}, in which the authors consider statistical differences in each orchard to propose a correction factor. Alternatively, it can be implemented as a machine learning model. In our case, we use a machine learning model that learns an implicit representation accounting for the impact of other variables, such as variety, group of variety, dimensions of the plant, region, and sector, to propose an estimated number of fruits.

\subsubsection{Data used for regression} 
\label{subsec:dataregress}
Although we received raw data for 1,543 trees, only 1,197 trees could be successfully processed in the previous stages of the pipeline. Trees with ill-recorded videos fail in the SfM part of the pipeline, and do not follow through the tracking and the yield detection steps. Table~\ref{tab:dataregressor} shows the first four lines of a regression data table. Tracking generated the last two columns, \emph{CbyT-A} and \emph{CbyT-B} (\emph{counting by tracking}), which correspond respectively to the fruits automatically counted on sides A and B of the tree. These counting columns can change according to the tracker employed, for example, a tracking using YOLOv3-based detections or
counting results from a YOLOv5-based one. Such counting values were aggregated to the original data to enrich the information provided to the regressor. However, not all 1,197 records could be used for training the regressor, as we had to exclude plants presenting missing data and those without fruit counting from tracking for either side.

\begin{table}[!htb]
	\caption{Sample containing the first four lines of the dataset used for training and testing of the regressor. ID 
 corresponds to the plant identifier; Sector and Region are references for geographical location; VarG and Var refer to the variety 
 group and variety of the plant, respectively; AG is a classification for the age of the plant; $F1$, $F2$, $F3$ and $F4$ are the number of 
 fruits manually counted in sizes corresponding to the first to the fourth flowering; H, W and D are the dimensions of the plant; 
 CbyT-A and CbyT-B are the number of visible fruits automatically counted using the vision-based pipeline for both sides of the plant.}
	\centering
	\begin{tabular}{llllcrrrrrrrrrr}
\toprule
ID & Sector & Region & VarG & Var.  & AG & $F1$ & $F2$ & $F3$ & $F4$ & $H$ & $W$ & $D$ & CbyT-A & CbyT-B \\
\midrule
103      & North    & LIM   & VT   & {\it Val.} & 1 & 0 & 5  & 26 & 1 & 280 & 280 & 300 & 15 & \\
1031     & South    & LIM   & NT   & {\it Nat.} & 3 & 0 & 5 & 54 & 10 & 290 & 360 & 295 &   & 19 \\
106      & North    & DUA   & PRME & {\it Pera} & 2 & 0 & 0 & 58 & 36 & 340 & 300 & 335 & 10 & 10 \\
104      & North    & DUA   & PRME & {\it Pera} & 3 & 400 & 0 & 11 & 0 & 310 & 460 & 495 &   & 32 \\

\bottomrule
\end{tabular}
	\label{tab:dataregressor}
\end{table}

\subsubsection{Data preprocessing}
Before submitting data to the model, it is necessary to perform some operations to adequate these data to the 
dynamics of the processing of the artificial neural network. The following operations were executed over the dataset:

\begin{description}
    \item[Transformation of categorical variables:] neural networks only accept numerical values as inputs to their neurons, which means that categorical variables must be transformed into numerical values. For example, the value \textsl{Hamlin}, which is one of the possible values for the variable \textsc{Variety}, has to be converted to a number. One of the possible methods is using One Hot Bit Encoding \citep{geron2019hands}, which creates a new column for each possible value of the variable.
    The column \textsl{Hamlin}, for instance, receives value 1 for whenever a sample is classified with that variety. This procedure is executed for all categorical variables.
    
     \item[Standardization of numerical values:] to facilitate convergence to the optimal training point, numerical values should be standardized. Variables with very discrepant scales, such as one ranging from [0–100,000] and another from [-1.0–1.0], can cause problems because the gradients calculated during back propagation will have very different effects when applied to different values. One solution for this problem is to apply standardization using the formula $\mathbf{u}_i = (\mathbf{v}_i - \overline{\mathbf{v}})/\sigma_\mathbf{v}$, where vector
     $\mathbf{u}_i$ contains the standardized values of $\mathbf{v}_{i}$, $\overline{\mathbf{v}}$ is the mean value of the vectors 
     $\mathbf{v}_i$ and $\sigma_\mathbf{v}$ its standard deviation.
\end{description}

\subsubsection{Machine learning regression model}
We have tried several machine learning algorithms for the regression problem, such as Support Vector Machines 
\citep{cortes1995support}, Bagging \citep{breiman1996bagging} and Gradient Boosting \citep{friedman01boosting}. None of them, 
however, had a better performance than a multilayer feed-forward neural network \citep{goodfellow2017deeplearning}.

Neural networks are highly flexible in their assembly. For instance, one can create a neural network with just a single layer, 
and it still is able to solve a problem. However, tasks such as image classification usually requires dozens or even hundreds of 
layers. This flexibility is also a weakness, as it necessitates searching through a wide range of configurations to find the most 
suitable one for a specific problem \citep{geron2019hands}. A common heuristic approach is to gradually increase the number of 
layers and neurons and evaluate how the results converge towards the desired outcome. Once the parameter range with the best 
results is identified, it can be further refined. In the case of the yield regressor, the neural network should not be 
excessively complex due to the amount of data available. If a neural network has too many parameters, it may suffer from 
overfitting, where it learns details in the training dataset and performs poorly with the test data as compared to its training 
performance.

In our path to find the best network configuration, we relied on \emph{cross validation}, a widely used technique in machine 
learning, to statistically assess whether one computational experiment performs better than another \citep{kohavi1995crossvalid}. 
Cross validation involves running the same algorithm multiple times, allowing for more reliable performance statistics and a 
certain level of confidence.
In our research, we divided the training dataset into ten parts, also known as folds. The process entails training the algorithm 
on nine folds while reserving the tenth fold for evaluation. This process repeats until all ten folds have been used as the test 
set. To determine the level of certainty regarding the superiority of a model, we employed a \emph{statistical hypothesis test}. 
Since we had ten folds, the assumption that the data follow a normal distribution does not hold true. The normal distribution is 
typically reserved for datasets larger than 30 elements with known variance \citep{king2019Statistics}. Instead, we used the 
t-Student distribution for our set of ten performance measurements. We set a p-value of 0.05 to ensure that we select a model 
with 95\% confidence.

We conducted experiments with neural networks comprising one to six layers, and varying the number of neurons per layer from seven 
to 28. The last layer of the regressor consists of a single neuron, which is responsible for the calculation of the yield 
estimation. Among the tested architectures, the one presented in Table~\ref{tab:architecture} demonstrated the best results, as will 
be discussed in the next section. The model was implemented in Python, using the Keras API\footnote{\url{https://keras.io/about}.}, 
which is built on top of the TensorFlow\textsuperscript{\texttrademark} platform. 

\begin{table}
	\caption{Feed-forward neural network architecture employed on yield regression. $B$ corresponds to
 the batch size, i.e., the number of data entries.}
	\centering
	\begin{tabular}{lll}
\toprule
Layer (type) & Output Shape & Num. Params \\
\midrule
dense\_1 (Dense)      & ($B$, 7)    & 294  \\
batch\_normalization\_1     & ($B$, 7)  & 28 \\
dense\_2 (Dense)      & ($B$, 7)    & 56  \\
dense\_3 (Dense)      & ($B$, 1)    & 8  \\

\bottomrule
Total params: 386 \\
Trainable params: 372 \\
Non-trainable params: 14\\
\bottomrule
\end{tabular}
	\label{tab:architecture}
\end{table}

%
%
\section{Results}
\label{sec:results}

\subsection{Tracking sensitivity to detection}\label{sec:sensitivity}

In the sensitivity analysis, tracking is performed by the algorithms presented in Section~\ref{sec:mot}, with no component
based on machine learning: all boxes come from the ground-truth, with random removal of some boxes to emulate detection
components presenting different detection rates. Table~\ref{tab:sensitivity} shows HOTA, DetA\footnote{Note that, 
for 100\% detection rate, to see DetA values below 1.0 sounds counterintuitive: readers should consider that 
the assignments between detections and ground-truth are defined by the association matrices
$\hat{\mathtt{A}}^{(\alpha)}$ that \emph{maximize HOTA} \citep{luiten2021hota}, so association errors could produce some false 
positives and false negatives in Equation~\ref{eq:DetAa}, see Appendix~\ref{apx:hota}.} and AssA results for four different rates. The table 
also displays the MOTA metric for comparison, considering that a few works \citep{jong2022apple,villacres2023apple} employed that metric. 
The \emph{CbyT} (counting by tracking) value is the fruit counting, corresponding exactly to the number of (possibly discontinuous) tracks 
found by our MOT system, while \emph{CbyT-GT} is the ground-truth value. The last column displays the relative error: the ratio of 
absolute error to the ground-truth value.

\begin{table}
\caption{Sensitivity analysis. The table shows how tracking (and fruit counting) degrades with detection performance.}
    \centering
    \footnotesize
        \begin{tabular}{llllllrrrr}
    
        Detection (\%) & Sequence & HOTA & DetA & AssA & MOTA & CbyT & CbyT-GT & Relative error (\%)  & Median \\ \hline
        \multirow{13}{*}{100\%} & V01 & 0.90691 & 0.90390 & 0.91079 & 0.96139 & 103 & 110 & 6.36\% & \multirow{12}{*}{1.23\%} \\ 
        ~ & V02 & 0.90277 & 0.90130 & 0.90683 & 0.97152 & 45 & 46 & 2.17\% & ~ \\ 
        ~ & V03 & 0.94512 & 0.94140 & 0.94907 & 0.97387 & 89 & 90 & 1.11\% & ~ \\ 
        ~ & V04 & 0.94159 & 0.94022 & 0.94306 & 0.98173 & 104 & 105 & 0.95\% & ~ \\ 
        ~ & V05 & 0.92378 & 0.93034 & 0.91765 & 0.95370 & 146 & 148 & 1.35\% & ~ \\ 
        ~ & V06 & 0.92984 & 0.92576 & 0.93423 & 0.97897 & 121 & 122 & 0.82\% & ~ \\ 
        ~ & V07 & 0.93595 & 0.92823 & 0.94428 & 0.97714 & 149 & 154 & 3.25\% & ~ \\ 
        ~ & V08 & 0.95137 & 0.94812 & 0.95515 & 0.99380 & 190 & 192 & 1.04\% & ~ \\ 
        ~ & V09 & 0.78204 & 0.62303 & 0.98163 & 0.63495 & 10 & 16 & 37.50\% & ~ \\ 
        ~ & V10 & 0.96471 & 0.96313 & 0.96640 & 0.99781 & 13 & 13 & 0.00\% & ~ \\ 
        ~ & V11 & 0.92858 & 0.90792 & 0.95016 & 0.95871 & 85 & 87 & 2.30\% & ~ \\ 
        ~ & V12 & 0.94809 & 0.94741 & 0.94898 & 0.99573 & 115 & 115 & 0.00\% & ~ \\ 
        ~ & \textbf{All} & 0.93516 & 0.92761 & 0.94361 & 0.97308 & 1170 & 1198 & \textbf{2.34\%} & ~ \\ \hline
        \multirow{13}{*}{80\%} & V01 & 0.68845 & 0.68009 & 0.69767 & 0.71980 & 101 & 110 & 8.18\% & \multirow{12}{*}{2.20\%} \\ 
        ~ & V02 & 0.69783 & 0.69759 & 0.70022 & 0.74322 & 47 & 46 & 2.17\% & ~ \\ 
        ~ & V03 & 0.72111 & 0.72464 & 0.71773 & 0.73939 & 88 & 90 & 2.22\% & ~ \\ 
        ~ & V04 & 0.72459 & 0.72899 & 0.72030 & 0.75362 & 104 & 105 & 0.95\% & ~ \\ 
        ~ & V05 & 0.67268 & 0.70495 & 0.64206 & 0.70454 & 147 & 148 & 0.68\% & ~ \\ 
        ~ & V06 & 0.72021 & 0.71768 & 0.72294 & 0.75514 & 119 & 122 & 2.46\% & ~ \\ 
        ~ & V07 & 0.71791 & 0.71040 & 0.72589 & 0.74529 & 144 & 154 & 6.49\% & ~ \\ 
        ~ & V08 & 0.72352 & 0.72368 & 0.72372 & 0.74971 & 189 & 192 & 1.56\% & ~ \\ 
        ~ & V09 & 0.63278 & 0.56457 & 0.70923 & 0.57254 & 12 & 16 & 25.00\% & ~ \\ 
        ~ & V10 & 0.76247 & 0.75657 & 0.76851 & 0.78134 & 13 & 13 & 0.00\% & ~ \\ 
        ~ & V11 & 0.69139 & 0.67515 & 0.70832 & 0.71356 & 82 & 87 & 5.75\% & ~ \\ 
        ~ & V12 & 0.72052 & 0.72119 & 0.72001 & 0.75194 & 116 & 115 & 0.87\% & ~ \\ 
        ~ & \textbf{All} & 0.71210 & 0.71085 & 0.71386 & 0.73859 & 1162 & 1198 & \textbf{3.01\%} & ~ \\ \hline
        \multirow{13}{*}{60\%} & V01 & 0.42052 & 0.37871 & 0.46725 & 0.39628 & 77 & 110 & 30.00\% & \multirow{12}{*}{12.35\%} \\ 
        ~ & V02 & 0.43747 & 0.42167 & 0.45582 & 0.44970 & 36 & 46 & 21.74\% & ~ \\ 
        ~ & V03 & 0.45983 & 0.46422 & 0.45555 & 0.46182 & 81 & 90 & 10.00\% & ~ \\ 
        ~ & V04 & 0.48905 & 0.49809 & 0.48025 & 0.49698 & 100 & 105 & 4.76\% & ~ \\ 
        ~ & V05 & 0.41860 & 0.43778 & 0.40030 & 0.41636 & 116 & 148 & 21.62\% & ~ \\ 
        ~ & V06 & 0.46982 & 0.45067 & 0.48991 & 0.46902 & 113 & 122 & 7.38\% & ~ \\ 
        ~ & V07 & 0.48637 & 0.47560 & 0.49763 & 0.49048 & 132 & 154 & 14.29\% & ~ \\ 
        ~ & V08 & 0.48001 & 0.47564 & 0.48464 & 0.48208 & 172 & 192 & 10.42\% & ~ \\ 
        ~ & V09 & 0.42851 & 0.38464 & 0.47741 & 0.38014 & 11 & 16 & 31.25\% & ~ \\ 
        ~ & V10 & 0.46489 & 0.48524 & 0.44543 & 0.47886 & 12 & 13 & 7.69\% & ~ \\ 
        ~ & V11 & 0.44513 & 0.41282 & 0.48018 & 0.42409 & 67 & 87 & 22.99\% & ~ \\ 
        ~ & V12 & 0.47738 & 0.47390 & 0.48098 & 0.48000 & 108 & 115 & 6.09\% & ~ \\ 
        ~ & \textbf{All} & 0.46614 & 0.45801 & 0.47467 & 0.46427 & 1025 & 1198 & \textbf{14.44\%} & ~ \\ \hline
        \multirow{13}{*}{40\%} & V01 & 0.16977 & 0.10001 & 0.28839 & 0.10474 & 30 & 110 & 72.73\% & \multirow{12}{*}{55.77\%} \\ 
        ~ & V02 & 0.16722 & 0.10682 & 0.26188 & 0.11438 & 15 & 46 & 67.39\% & ~ \\ 
        ~ & V03 & 0.19442 & 0.14486 & 0.26096 & 0.14433 & 42 & 90 & 53.33\% & ~ \\ 
        ~ & V04 & 0.20442 & 0.17707 & 0.23600 & 0.17250 & 56 & 105 & 46.67\% & ~ \\ 
        ~ & V05 & 0.16615 & 0.13389 & 0.20618 & 0.12626 & 57 & 148 & 61.49\% & ~ \\ 
        ~ & V06 & 0.19135 & 0.13771 & 0.26587 & 0.14189 & 51 & 122 & 58.20\% & ~ \\ 
        ~ & V07 & 0.21561 & 0.17215 & 0.27009 & 0.17652 & 75 & 154 & 51.30\% & ~ \\ 
        ~ & V08 & 0.21683 & 0.17808 & 0.26408 & 0.17686 & 101 & 192 & 47.40\% & ~ \\
        ~ & V09 & 0.18710 & 0.10897 & 0.32127 & 0.11076 & 4 & 16 & 75.00\% & ~ \\ 
        ~ & V10 & 0.25155 & 0.22578 & 0.28028 & 0.21137 & 8 & 13 & 38.46\% & ~ \\ 
        ~ & V11 & 0.18189 & 0.12892 & 0.25676 & 0.13379 & 32 & 87 & 63.22\% & ~ \\ 
        ~ & V12 & 0.19288 & 0.14647 & 0.25407 & 0.15039 & 57 & 115 & 50.43\% & ~ \\ 
        ~ & \textbf{All} & 0.19832 & 0.15263 & 0.25773 & 0.15328 & 528 & 1198 & \textbf{55.93\%} & ~ \\ \hline
    \end{tabular}

    \label{tab:sensitivity}
\end{table}

Perfect fruit detection does not imply in perfect tracking and counting: association errors, as seen in Figure~\ref{fig:hota},
induced by occlusions, cause tracking errors. However, the tracking performance in high (HOTA and MOTA above 0.9),
resulting in accurate counting: 2.34\% relative error in counting of visible fruits for a 100\% detection rate (considering individual trees, the median for the relative error is 1.23\%). But accurate counting can be reached even under imperfect detections: an 80\% detection rate could reach a 3.01\% relative error (median 2.20\%). As expected, low detection rates will severely damage counting: a 60\% detection rate implied in a 14.44\% error in counting (median 12.35\%), while a  40\% rate reached an error of 55.93\% (median 55.77\%).

\begin{figure}
     \centering
     \begin{subfigure}[b]{0.18\textwidth}
         \centering
         \includegraphics[width=\textwidth]{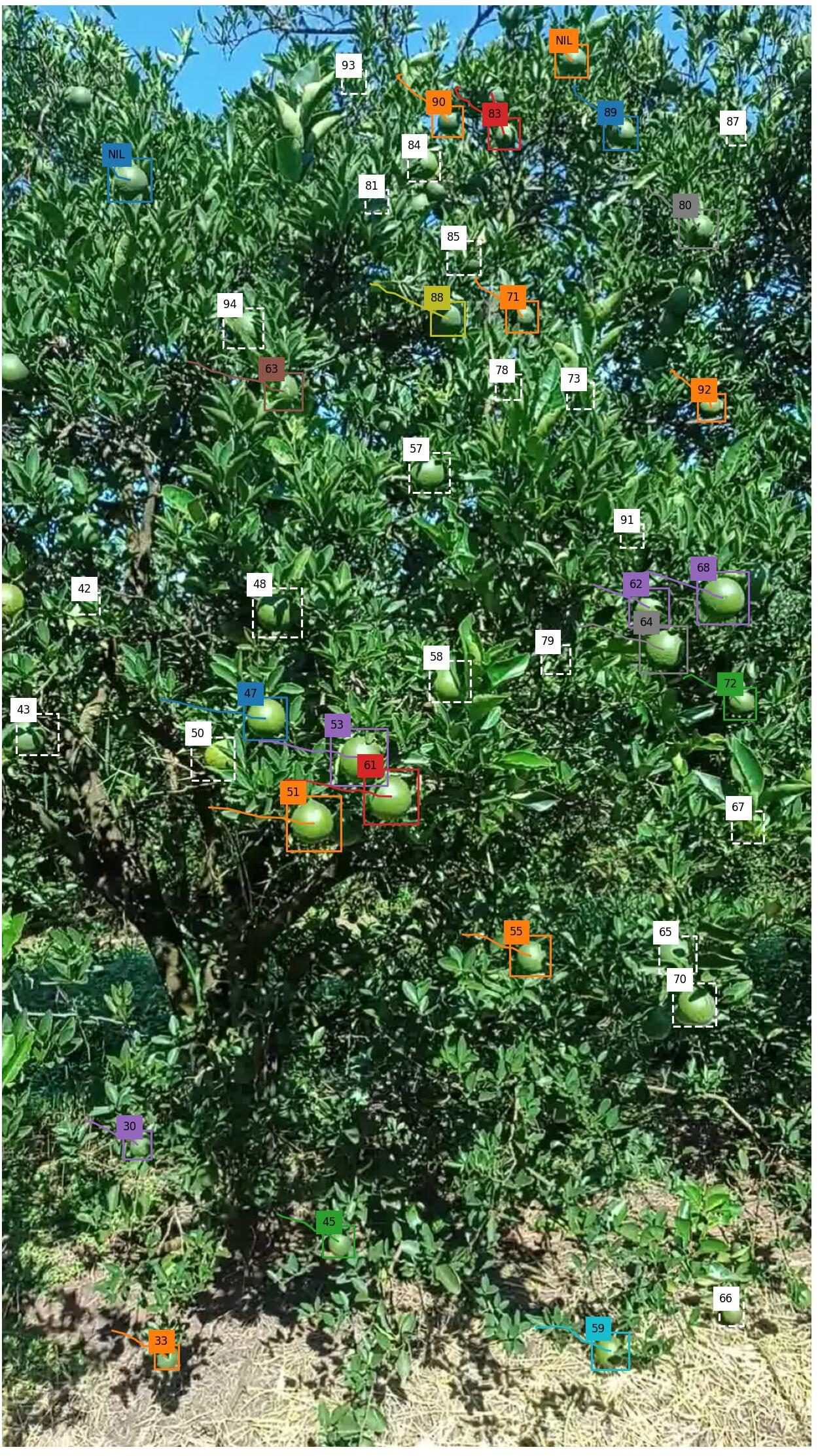}
         \caption{V07, $f_{260}$}
         \label{fig:1253A:frame260}
     \end{subfigure}  
     \hfill
     \begin{subfigure}[b]{0.18\textwidth}
         \centering
         \includegraphics[width=\textwidth]{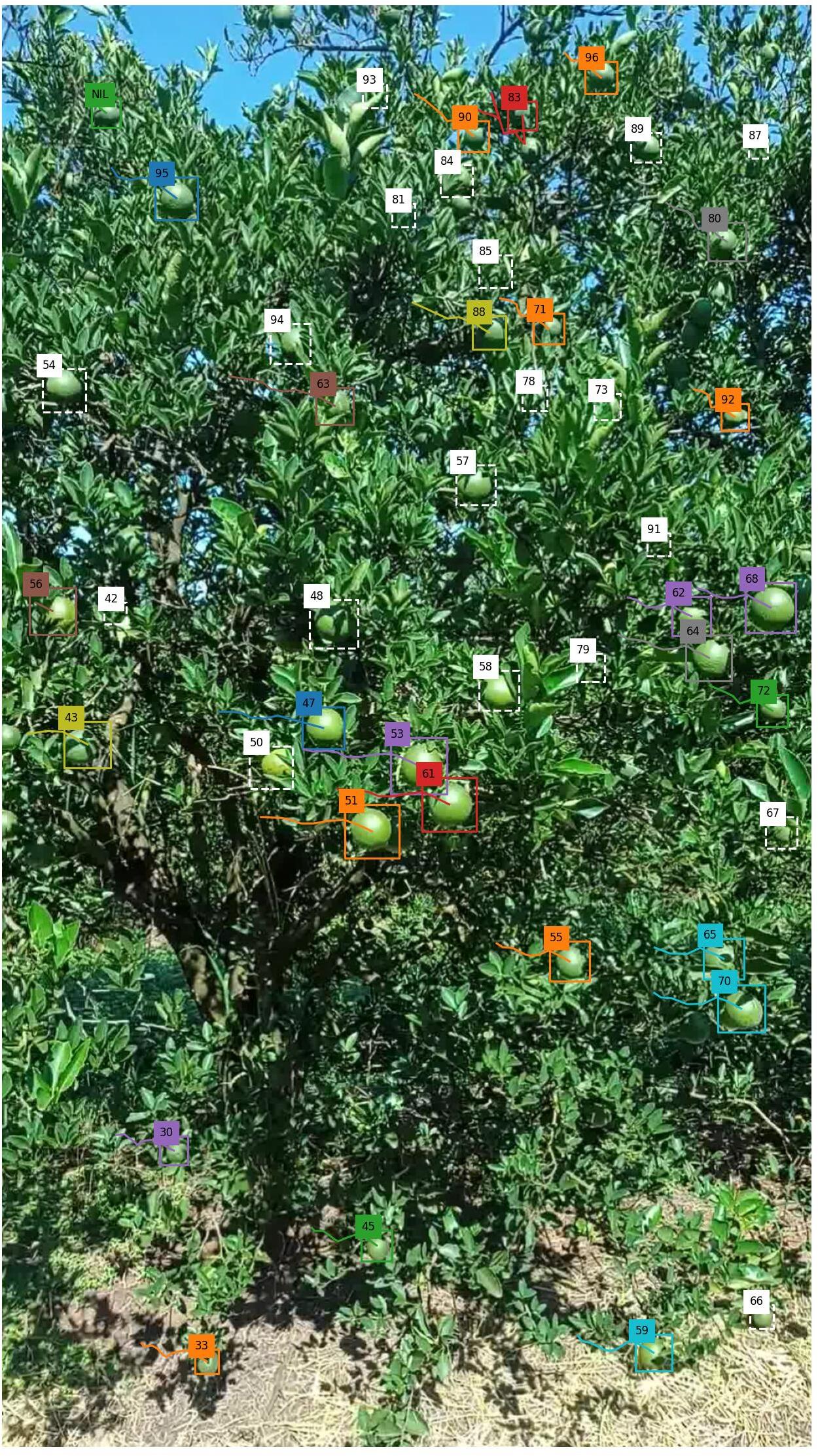}
         \caption{V07, $f_{265}$}
         \label{fig:1253A:frame265}
     \end{subfigure}  
     \hfill
     \begin{subfigure}[b]{0.18\textwidth}
         \centering
         \includegraphics[width=\textwidth]{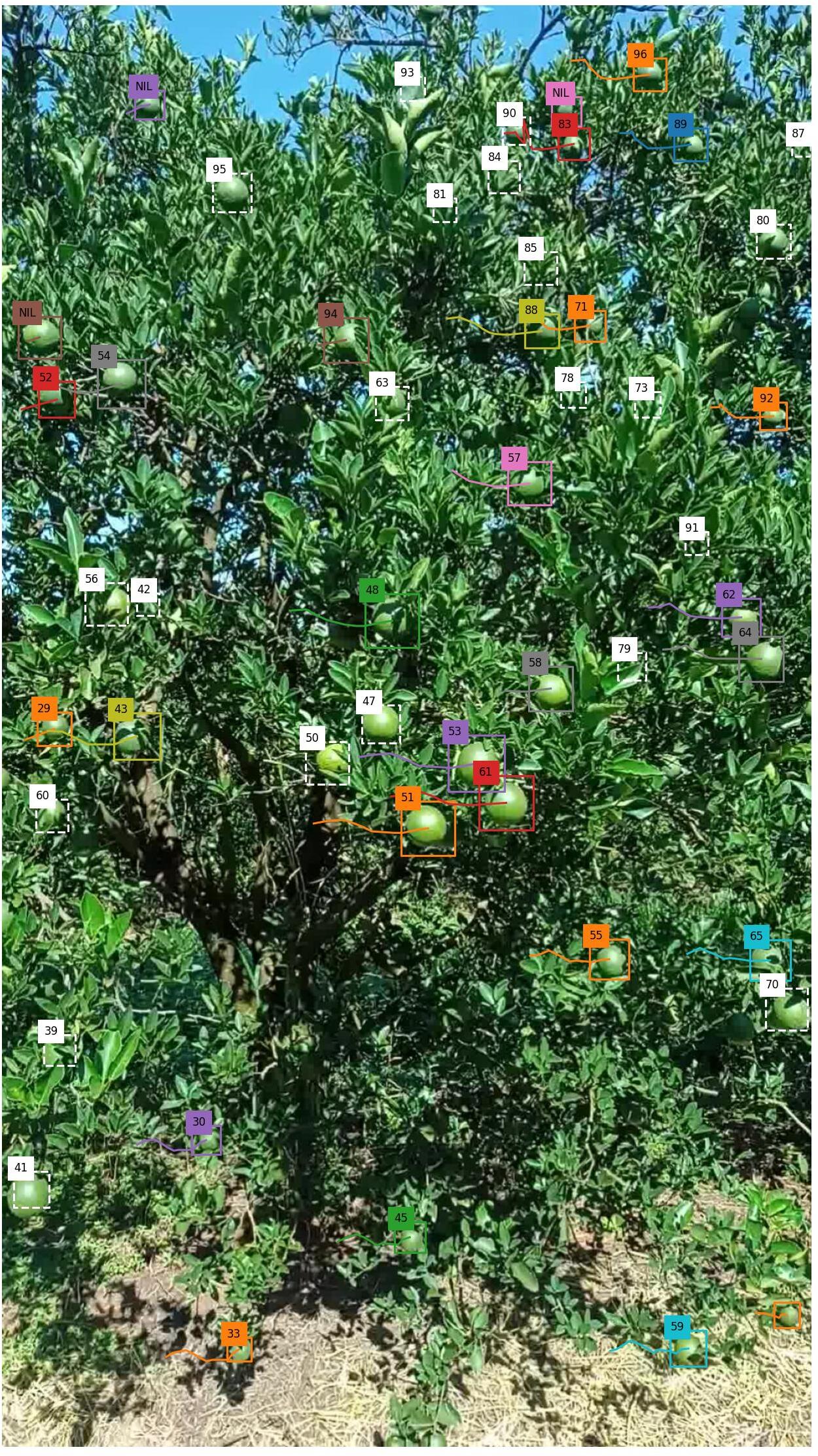}
         \caption{V07, $f_{270}$}
         \label{fig:1253A:frame270}
     \end{subfigure}  
     \hfill
     \begin{subfigure}[b]{0.18\textwidth}
         \centering
         \includegraphics[width=\textwidth]{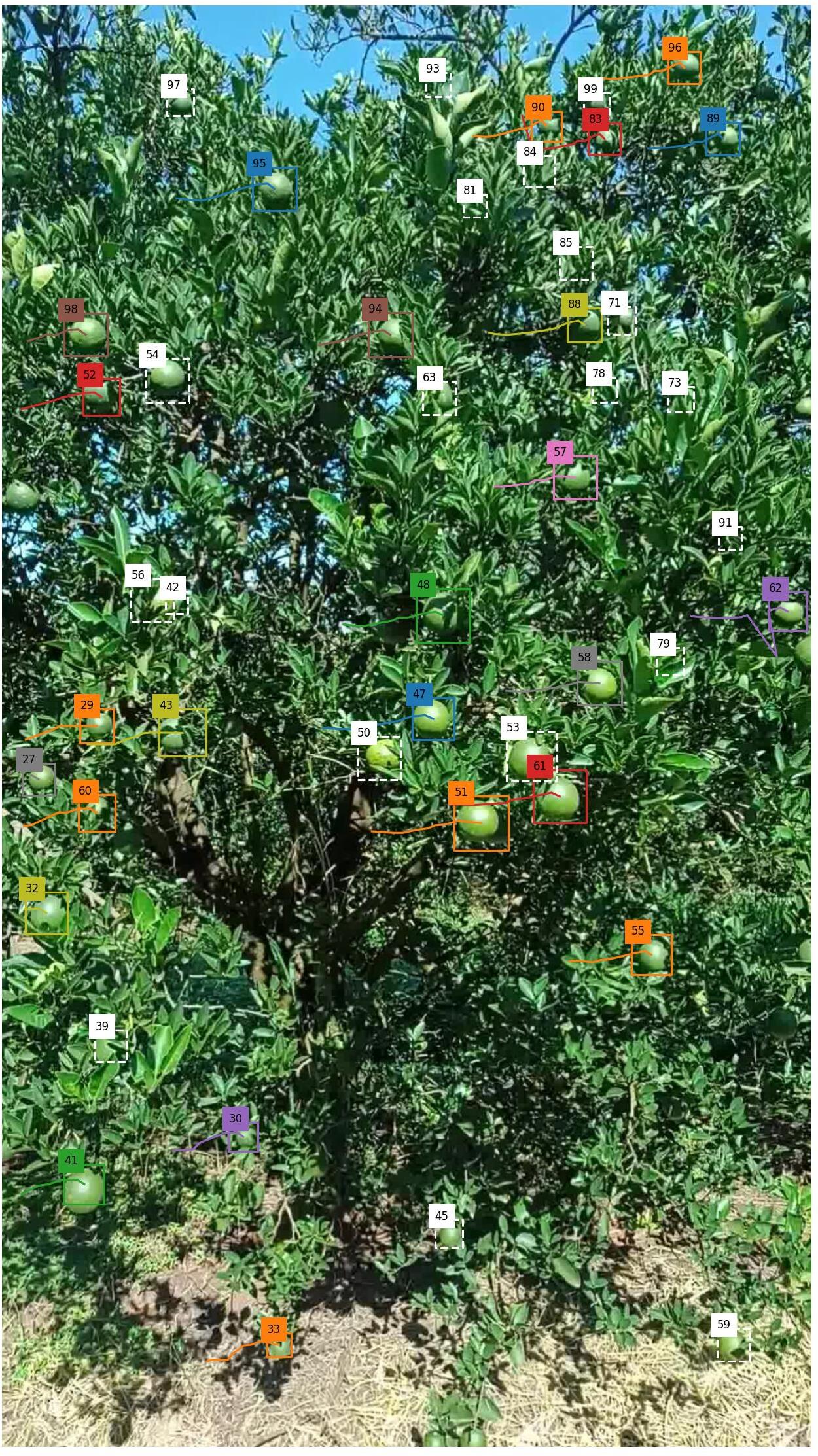}
         \caption{V07, $f_{275}$}
         \label{fig:1253A:frame275}
     \end{subfigure}  
     \hfill
     \begin{subfigure}[b]{0.18\textwidth}
         \centering
         \includegraphics[width=\textwidth]{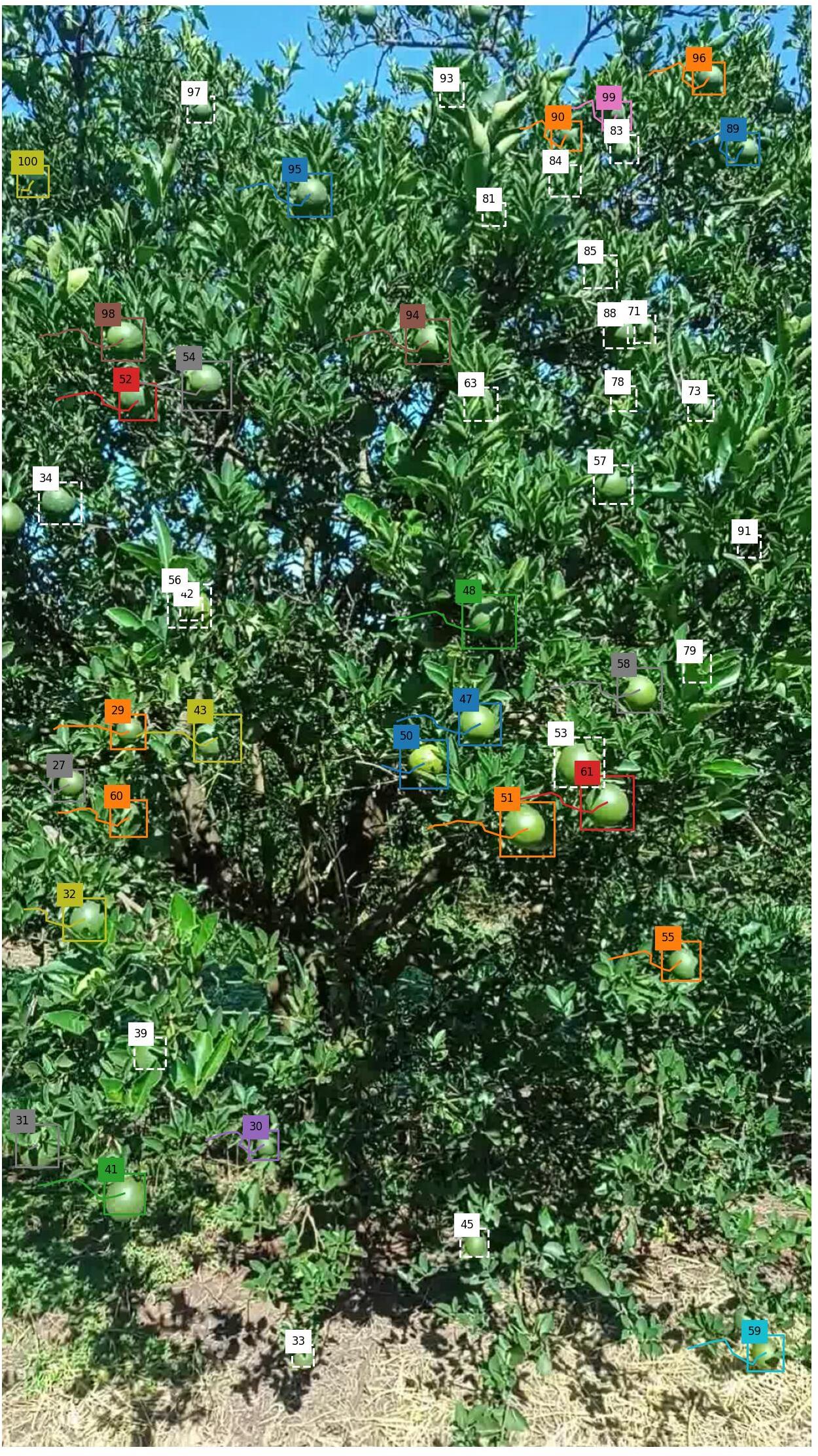}
         \caption{V07, $f_{280}$}
         \label{fig:1253A:frame280}
     \end{subfigure}  
     \begin{subfigure}[b]{0.18\textwidth}
         \centering
         \includegraphics[width=\textwidth]{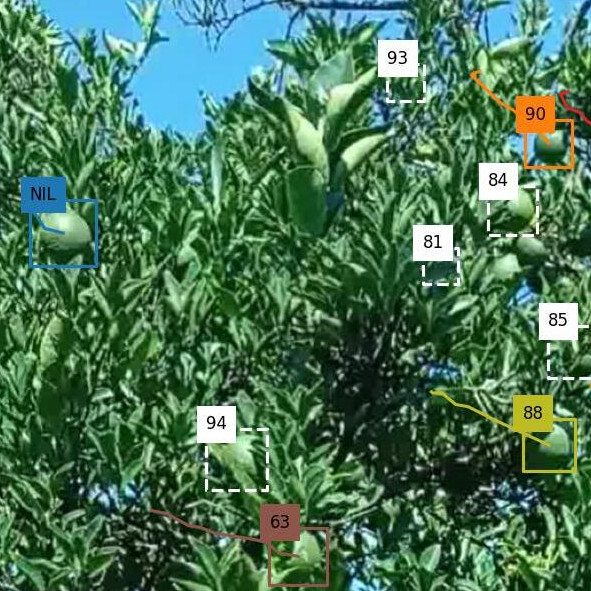}
         \caption{V07, $f_{260}$ (detail)}
         \label{fig:1253A:frame260:detail}
     \end{subfigure}  
     \hfill
     \begin{subfigure}[b]{0.18\textwidth}
         \centering
         \includegraphics[width=\textwidth]{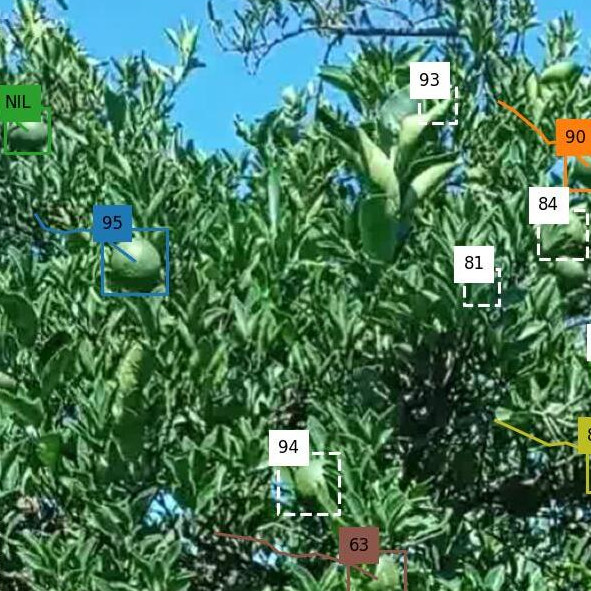}
         \caption{V07, $f_{265}$ (detail)}
         \label{fig:1253A:frame265:detail}
     \end{subfigure}  
     \hfill
     \begin{subfigure}[b]{0.18\textwidth}
         \centering
         \includegraphics[width=\textwidth]{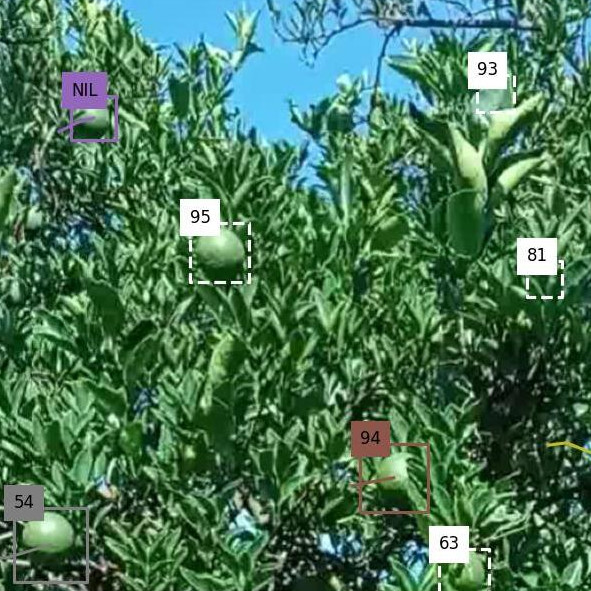}
         \caption{V07, $f_{270}$ (detail)}
         \label{fig:1253A:frame270:detail}
     \end{subfigure}  
     \hfill
     \begin{subfigure}[b]{0.18\textwidth}
         \centering
         \includegraphics[width=\textwidth]{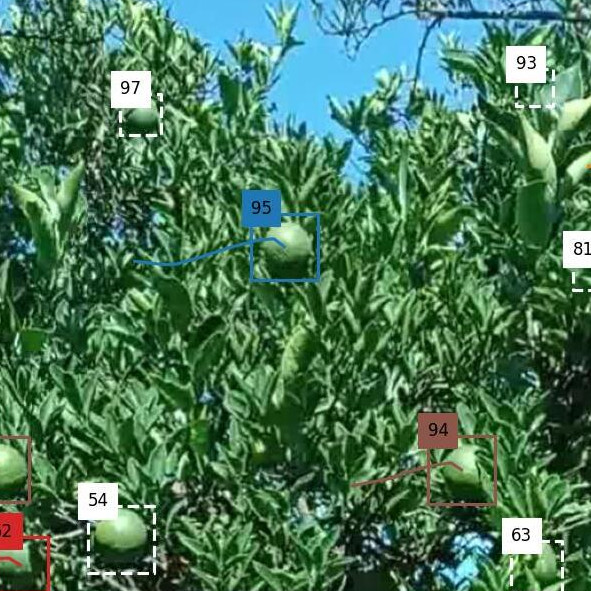}
         \caption{V07, $f_{275}$ (detail)}
         \label{fig:1253A:frame275:detail}
     \end{subfigure}  
     \hfill
     \begin{subfigure}[b]{0.18\textwidth}
         \centering
         \includegraphics[width=\textwidth]{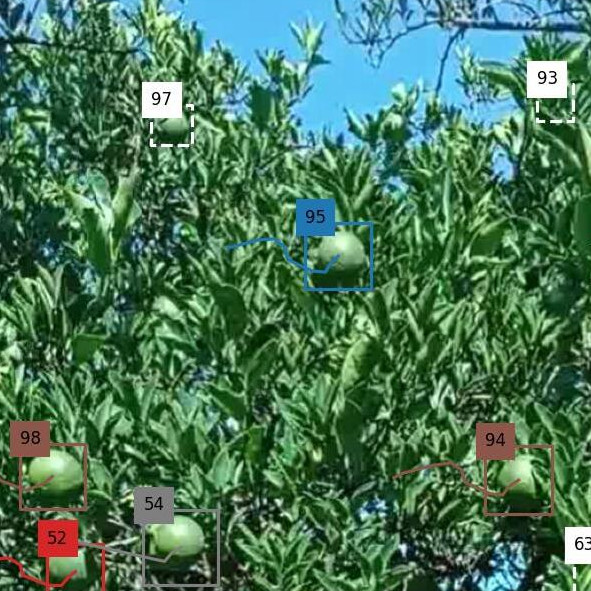}
         \caption{V07, $f_{280}$ (detail)}
         \label{fig:1253A:frame280:detail}
     \end{subfigure}  
     \begin{subfigure}[b]{0.18\textwidth}
         \centering
         \includegraphics[width=\textwidth]{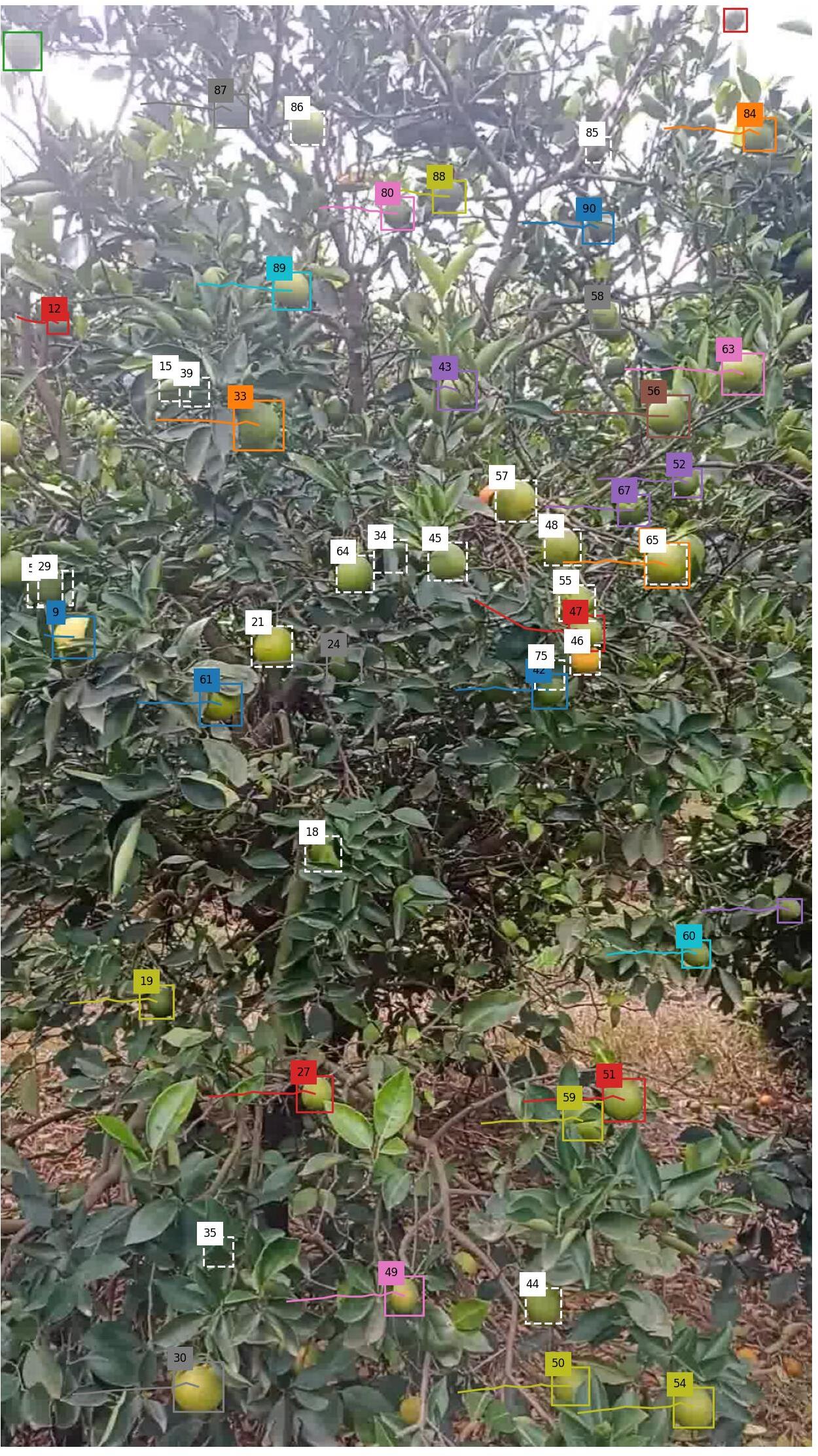}
         \caption{V12, $f_{140}$}
         \label{fig:1313B:frame140}
     \end{subfigure}  
     \hfill
     \begin{subfigure}[b]{0.18\textwidth}
         \centering
         \includegraphics[width=\textwidth]{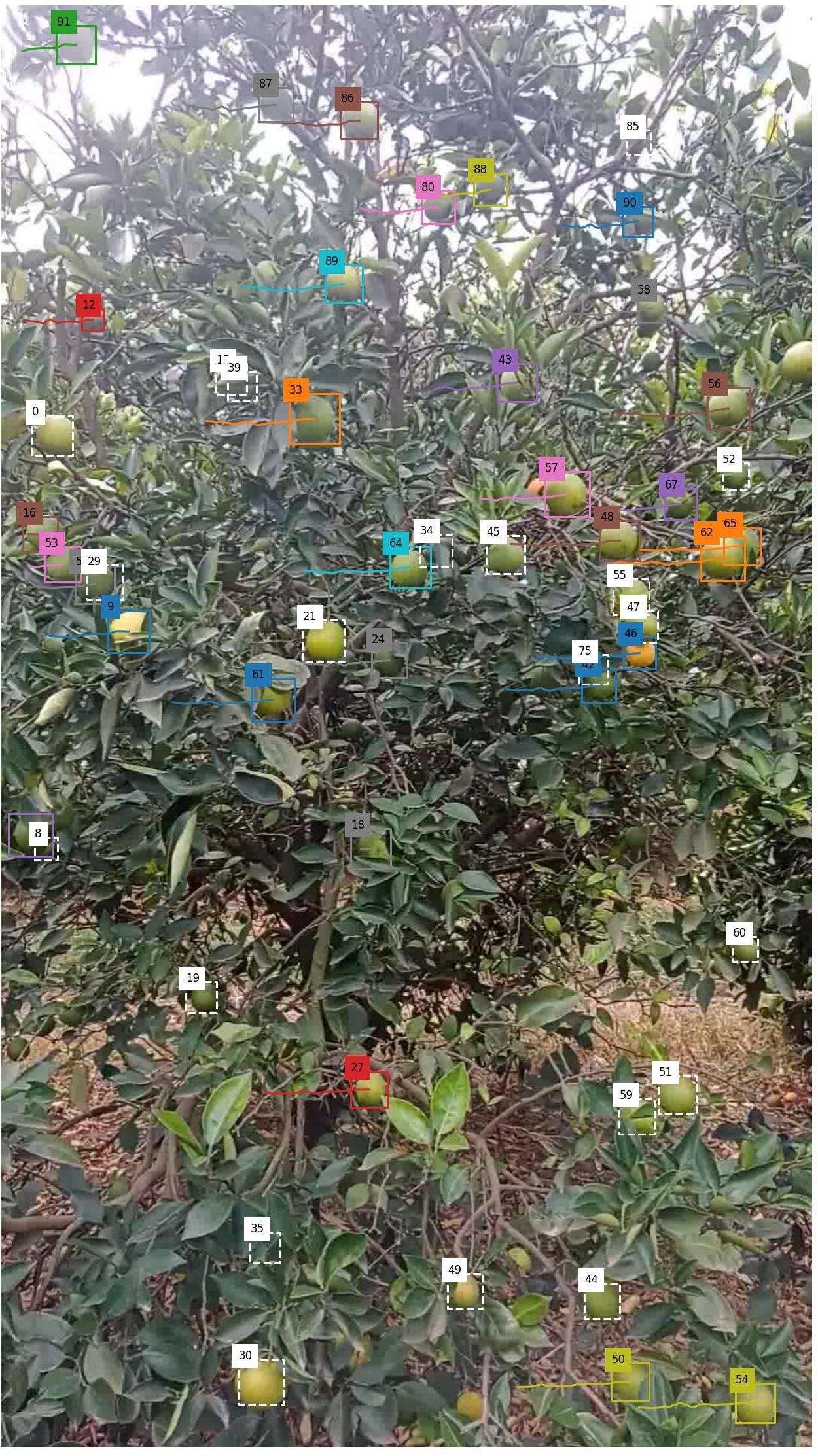}
         \caption{V12, $f_{145}$}
         \label{fig:1313B:frame145}
     \end{subfigure}  
     \hfill
     \begin{subfigure}[b]{0.18\textwidth}
         \centering
         \includegraphics[width=\textwidth]{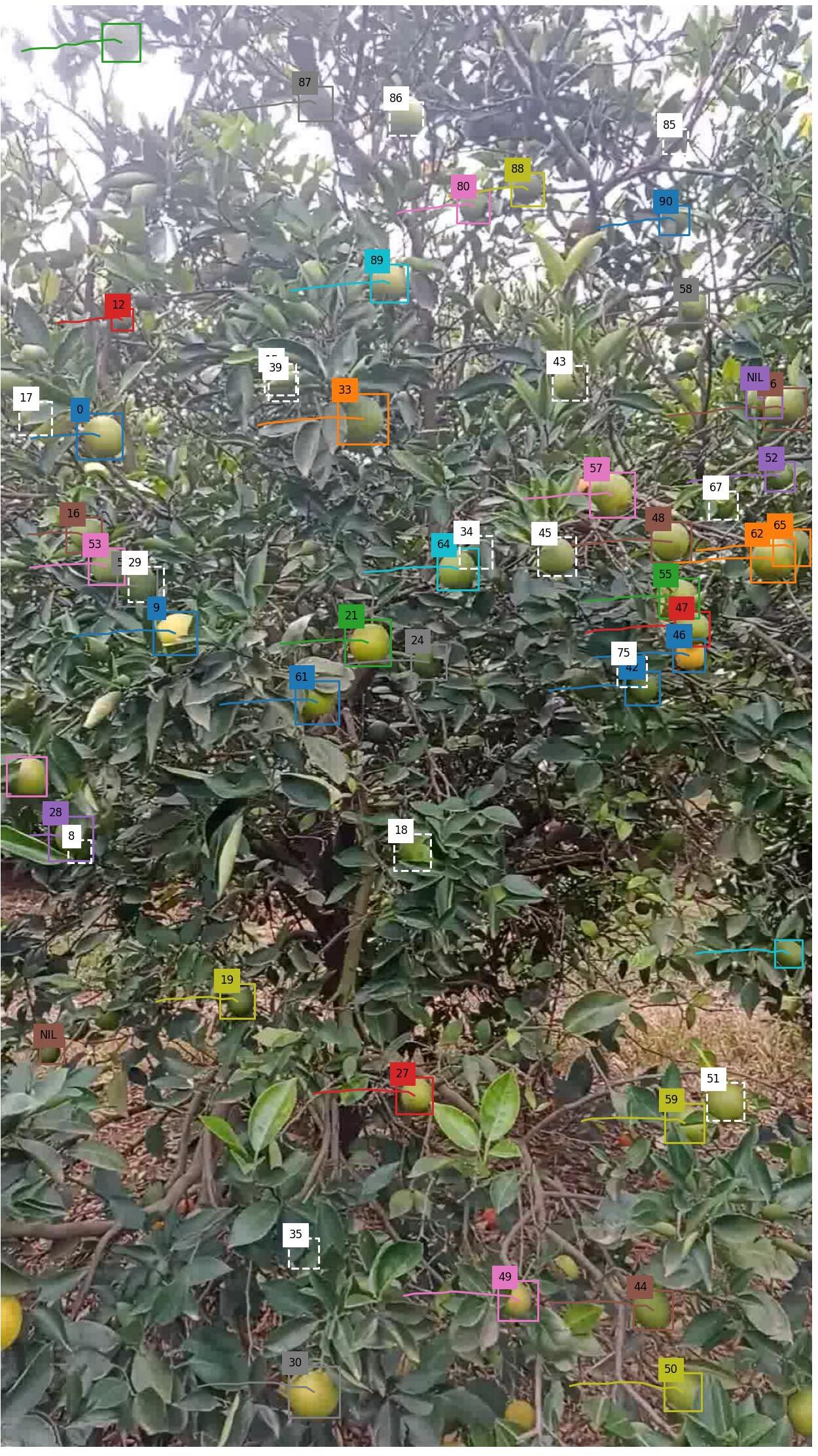}
         \caption{V12, $f_{150}$}
         \label{fig:1313B:frame150}
     \end{subfigure}  
     \hfill
     \begin{subfigure}[b]{0.18\textwidth}
         \centering
         \includegraphics[width=\textwidth]{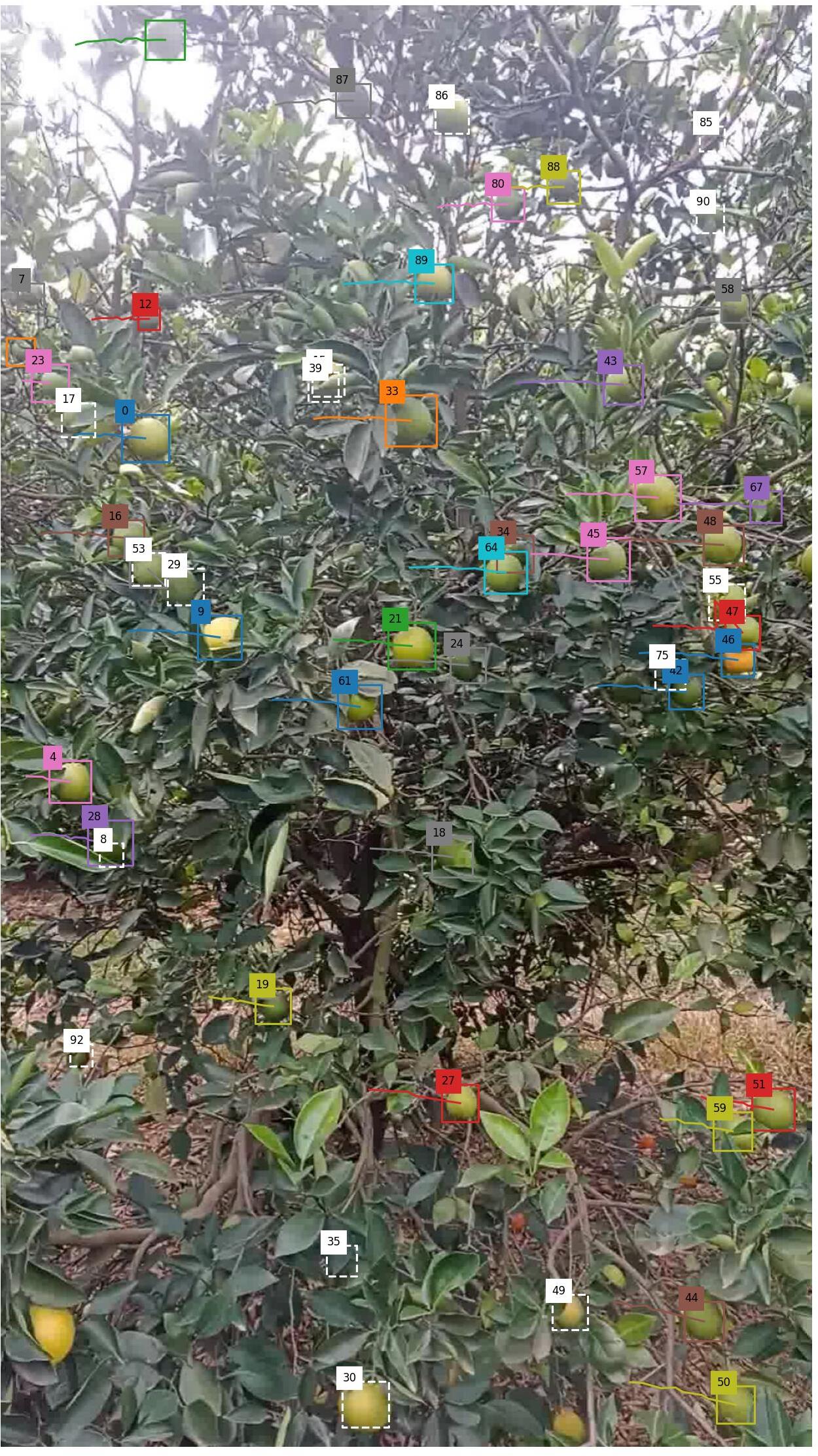}
         \caption{V12, $f_{155}$}
         \label{fig:1313B:frame155}
     \end{subfigure}  
     \hfill
     \begin{subfigure}[b]{0.18\textwidth}
         \centering
         \includegraphics[width=\textwidth]{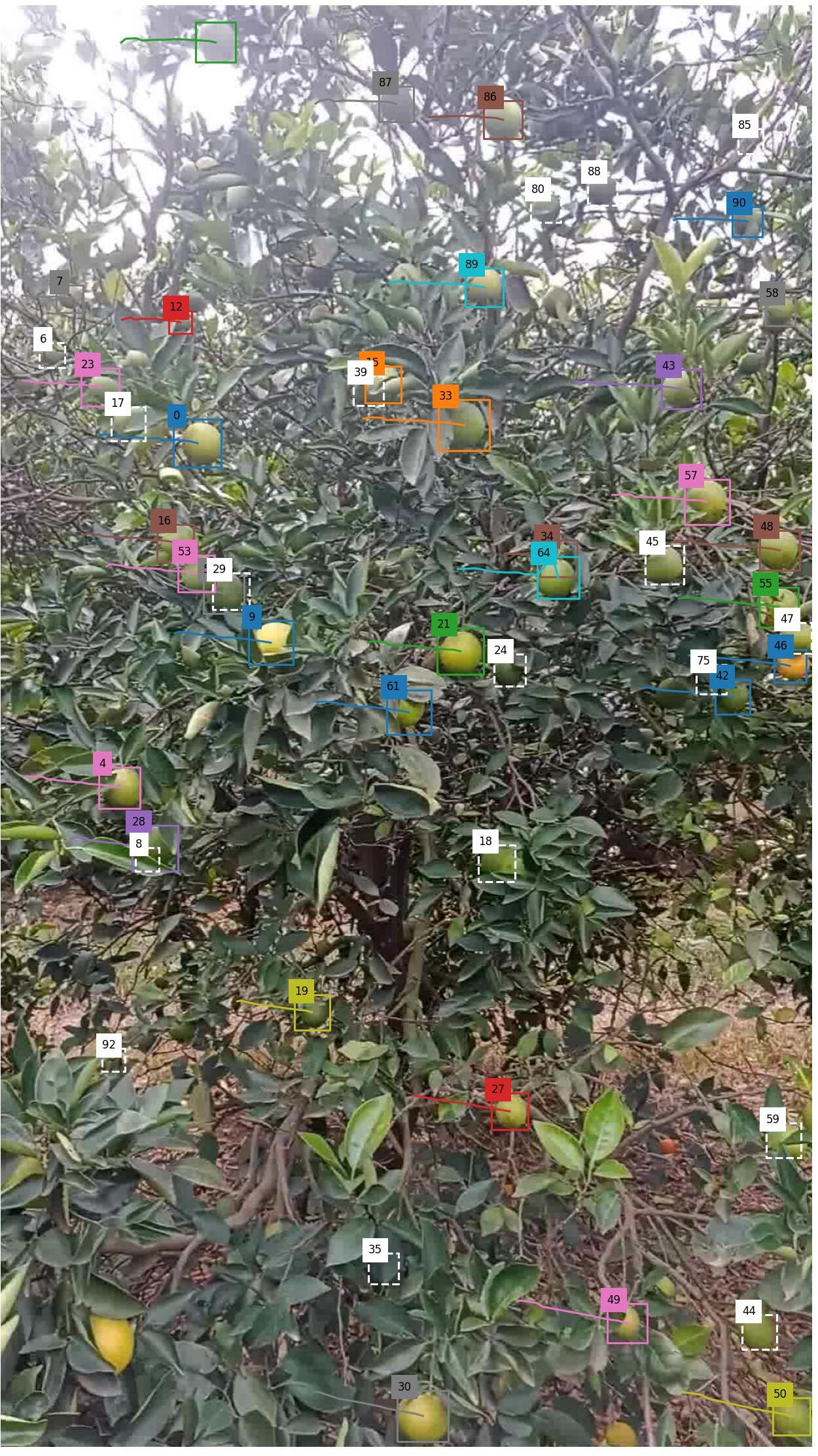}
         \caption{V12, $f_{160}$}
         \label{fig:1313B:frame160}
     \end{subfigure}       
     \caption{Tracking results for inputs presenting 80\% fruit detection ratio. Tracks whose oranges' 3-D models were estimated present numerics IDs. New tracks whose oranges were not yet estimated successfully presents the marker \textsc{Nil}. Tracks marked by dashed white lines are \textsc{Lost} tracks. (a-j) Frames from V07 (\emph{Valencia}). (k-o) Frames from V12 (\emph{Hamlin}). 
 Best seen in digital format.} 
     \label{fig:res-tracking-08}
\end{figure}

\begin{figure}
	\centering
        \includegraphics[width=\textwidth]{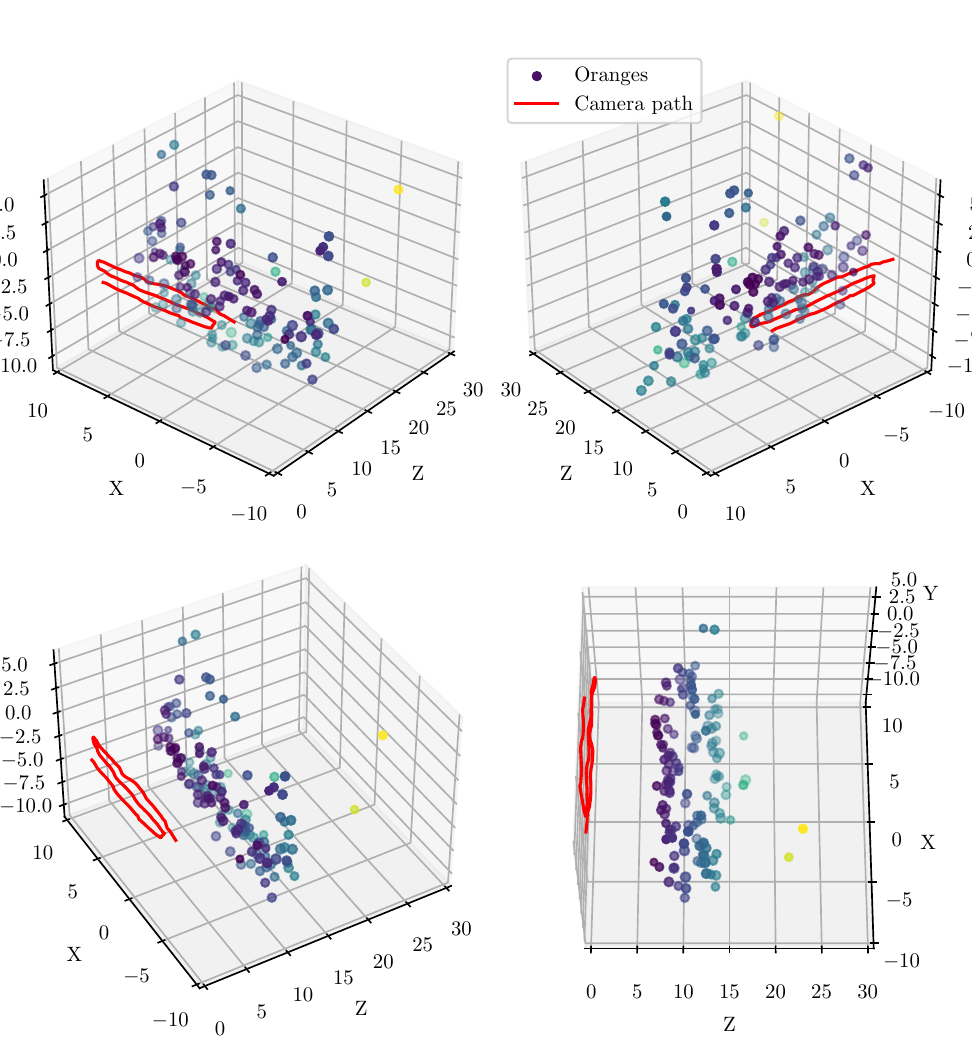}
	\caption{Estimations for oranges' 3-D localizations in V07. MOT was performed using an 80\% detection rate. Scale
 is arbitrary (SfM without metric scale calibration). The camera path is shown in red. The 145 found oranges are shown as circles,
 the color scale representing the distance in the $Z$ axis (depth). The four plots show different angles of the same structure.}
	\label{fig:oranges3d}
\end{figure}

Figure~\ref{fig:res-tracking-08} shows tracking examples for two videos, V07 and V12. The tracking was performed assuming an 80\% detection rate. When the orange position and ray were estimated by Algorithm~\ref{alg:orange_estimation}, the numeric ID for the track (fruit) is displayed. Tracks not yet presenting a successful estimation are marked with a \textsc{Nil} label, otherwise. \textsc{Lost} tracks are displayed as white, dashed boxes. Note as, at each frame, there is a significant number of \textsc{Lost}
tracks, but the system keeps projecting them on each frame at their expected locations. A relocalization example can be seen for
track 94 in V07: the track is lost at frames $f_{260}$ and $f_{265}$ because of occlusion, but the track became \textsc{Active}
at $f_{270}$. Examples of new tracks creation and orange estimation can be seen for fruits 95, 96 and 97, again in video V07. The
three-dimensional positions estimated for each orange in V07 are shown in Figure~\ref{fig:oranges3d}.

The importance of relocalization is highlighted in Table~\ref{tab:sensitivity:noreloc}, which presents sensitivity 
results for both 100\% and 80\% 
detection rates, but without incorporating relocalization. In the case of multiple orange tracking, effectively managing 
occlusions or re-entering fruits is crucial. Without proper handling of these scenarios, the generated counting becomes 
unstable and unusable, as observed in the relative error column. A comparison between the results in Table~\ref{tab:sensitivity} 
and Table~\ref{tab:sensitivity:noreloc} emphasizes the necessity of long-term tracking for accurate counting. Even a more 
advanced tracker that incorporates motion estimation would struggle to handle occlusions and re-entering objects without a 
dedicated long-term association component \citep{bewley2016sort}.

\begin{table}
\caption{Sensitivity analysis without relocalization. The table shows how tracking (and fruit counting) degrades with detection performance if the tracker does not incorporate relocalization.}
    \centering
    \footnotesize
    \begin{tabular}{llllllrrrr}        
Detection (\%) & Sequence & HOTA & DetA & AssA & MOTA & CbyT & CbyT-GT & Relative error (\%) \\ \hline

        \multirow{13}{*}{100\%} & V01 & 0.34960 & 0.44284 & 0.27629 & 0.44890 & 244 & 110 & 121.82\% & \multirow{12}{*}{57.40\%} \\ 
        ~ & V02 & 0.43208 & 0.54064 & 0.34543 & 0.55949 & 139 & 46 & 202.17\% & ~ \\ 
        ~ & V03 & 0.48072 & 0.50524 & 0.45741 & 0.51485 & 150 & 90 & 66.67\% & ~ \\ 
        ~ & V04 & 0.40903 & 0.42646 & 0.39232 & 0.43890 & 169 & 105 & 60.95\% & ~ \\ 
        ~ & V05 & 0.35899 & 0.55269 & 0.23319 & 0.52738 & 581 & 148 & 292.57\% & ~ \\ 
        ~ & V06 & 0.46115 & 0.45874 & 0.46365 & 0.47716 & 183 & 122 & 50.00\% & ~ \\ 
        ~ & V07 & 0.40752 & 0.40948 & 0.40565 & 0.42184 & 276 & 154 & 79.22\% & ~ \\ 
        ~ & V08 & 0.45201 & 0.44898 & 0.45514 & 0.46420 & 289 & 192 & 50.52\% & ~ \\ 
        ~ & V09 & 0.42806 & 0.31271 & 0.58611 & 0.31669 & 14 & 16 & 12.50\% & ~ \\ 
        ~ & V10 & 0.40208 & 0.42290 & 0.38235 & 0.43222 & 20 & 13 & 53.85\% & ~ \\ 
        ~ & V11 & 0.48725 & 0.44734 & 0.53124 & 0.46841 & 101 & 87 & 16.09\% & ~ \\ 
        ~ & V12 & 0.48120 & 0.50184 & 0.46146 & 0.51958 & 171 & 115 & 48.70\% & ~ \\ 
        ~ & \textbf{All} & 0.43426 & 0.46100 & 0.40914 & 0.47157 & 2337 & 1198 & \textbf{95.08\%} & ~ \\ \hline
        \multirow{13}{*}{80\%} & V01 & 0.17219 & 0.24220 & 0.12258 & 0.23293 & 272 & 110 & 147.27\% & \multirow{12}{*}{238.31\%} \\ 
        ~ & V02 & 0.16378 & 0.29666 & 0.09050 & 0.28548 & 176 & 46 & 282.61\% & ~ \\ 
        ~ & V03 & 0.13913 & 0.24712 & 0.07838 & 0.22648 & 328 & 90 & 264.44\% & ~ \\ 
        ~ & V04 & 0.13699 & 0.25693 & 0.07307 & 0.23968 & 436 & 105 & 315.24\% & ~ \\ 
        ~ & V05 & 0.17829 & 0.33577 & 0.09468 & 0.29965 & 575 & 148 & 288.51\% & ~ \\ 
        ~ & V06 & 0.15788 & 0.24045 & 0.10378 & 0.23092 & 368 & 122 & 201.64\% & ~ \\ 
        ~ & V07 & 0.13617 & 0.25015 & 0.07417 & 0.23541 & 609 & 154 & 295.45\% & ~ \\ 
        ~ & V08 & 0.13391 & 0.24435 & 0.07344 & 0.22780 & 766 & 192 & 298.96\% & ~ \\ 
        ~ & V09 & 0.12673 & 0.15943 & 0.10075 & 0.14665 & 37 & 16 & 131.25\% & ~ \\ 
        ~ & V10 & 0.14627 & 0.22291 & 0.09606 & 0.21137 & 36 & 13 & 176.92\% & ~ \\ 
        ~ & V11 & 0.16579 & 0.23478 & 0.11728 & 0.22643 & 214 & 87 & 145.98\% & ~ \\ 
        ~ & V12 & 0.15895 & 0.25511 & 0.09913 & 0.24286 & 359 & 115 & 212.17\% & ~ \\ 
        ~ & \textbf{All} & 0.15013 & 0.25560 & 0.08826 & 0.23942 & 4176 & 1198 & \textbf{248.58\%} \\ \hline
\end{tabular}

    \label{tab:sensitivity:noreloc}
\end{table}

\subsection{Detection, tracking and counting}

Table~\ref{tab:detection:results} shows the detection results for seven different orange models. The YOLOv5, YOLOv6, YOLOv7, and YOLOv8 
networks can use different backbone sizes. We have tested the \emph{small} and \emph{large} versions of these networks, the latter producing the best
results, so only the \emph{large} (l) models are shown. Two different backbone sizes are also tested for EfficientDet: the B0 (baseline) 
and B3 backbones \citep{tan2020effdet}. The results in Table~\ref{tab:detection:results} consider a minimum score of 0.5 for the orange class and an IoU of
0.5 between predicted boxes and ground-truth. The values are \emph{not} an average over images in the test set: they were computed considering
the \emph{entire} set of 121,685 bounding boxes (oranges) in the \textsc{OranDet} test set, being an assessment of all fruit misses and
false positives.

As shown in Table~\ref{tab:detection:results}, YOLOv5l, YOLOv8l and YOLOv7 presented close results considering the $F_1$-score, followed
by YOLOv3. However, there is a slight difference in their behavior: the YOLOv3 and YOLOv8l networks present better precision, a lower number
of false positives, while the YOLOv5l and YOLOv7l models show better recall, missing fewer fruits. Note that the sensitivity test
presented in Section~\ref{sec:sensitivity} does not consider any level of false positives in the detections, neither any misalignment between
boxes in the ground-truth and in the predictions. The different false positive rates of the detection models and their localization accuracy
can affect the tracking.

\begin{table}[!htb]
\caption{Detection results considering a ground-truth composed by 121,685 bounding boxes in \textsc{OranDet} test set. The displayed results correspond to
a score threshold of 0.5 (orange class) and a minimum IoU of 0.5 between detection and ground-truth. TP, FP, and FN correspond to
true positives, false positives and false negatives, respectively.}
    \centering
    \footnotesize
    \begin{tabular}{lrrrrrrr}
                     & Detections & TP (\%) & FP (\%) & FN (\%) & Precision & Recall & $F_1$-score   \\ \hline
YOLOv5l              & 109,919     & \textbf{69.27}     & 21.06               & \textbf{30.73}     & 0.77      & \textbf{0.69}   & \textbf{0.73} \\
YOLOv8l              &  96,245     & 64.72              & 14.37               & 35.28              & 0.82      & 0.65   & 0.72 \\
YOLOv7l              & 101,914     & 66.14              & 17.61               & 33.86              & 0.79      & 0.66   & 0.72 \\
YOLOv3               &  80,092     & 57.68              & 8.14                & 42.32              & \textbf{0.88} & 0.58   & 0.70 \\
YOLOv6l              &  85,997     & 57.50              & 13.17               & 42.50              & 0.81      & 0.58   & 0.67 \\
EfficientDet~B3      &  79,129     & 55.42              & 9.61                & 44.58              & 0.85      & 0.55   & 0.67 \\
EfficientDet~B0      &  70,760     & 50.66              & \textbf{7.49}       & 49.34              & 0.87      & 0.51   & 0.64 \\
\end{tabular}

    \label{tab:detection:results}
\end{table}

Table~\ref{tab:tracking:results} shows tracking results for different orange detection models. For each frame, tilling with overlap is employed:
the frame is split into 24 tiles with $416 \times 416$ pixels, presenting overlaps of 82 pixels. The tiles are stacked in a single batch, and 
submitted to the fruit detection CNN. The detection results are merged (``untilling") and non-maximum suppression (NMS) applied (we have adopted 0.2
as IoU threshold for NMS). We have also filtered detections by score, testing three different score thresholds (0.5, 0.6 and 0.7).
Table~\ref{tab:tracking:results} displays the five best tracking results, considering the score threshold that produced the best result for 
such detection model. The results for each frame sequence in \textsc{MOrangeT} for the two best models (YOLOv5l and YOLOv3) are shown in 
Table~\ref{tab:tracking:results:pervid}. It is noteworthy that when the tracking results are integrated, i.e., when the 1,198 fruits/tracks are 
considered, the counting error rates are low. 

\begin{table}[!htb]
\caption{Tracking results, considering the ground-truth in \textsc{MOrangeT}. The relative error is computed for
the entire set of tracks in the 12 frame sequences, summing 1,198 tracks. The value in the last column (Median) is 
computed by evaluating the relative error for each video and getting the median (see Table~\ref{tab:tracking:results:pervid}).}
    \centering
    \footnotesize

\begin{tabular}{lrrrrrrr}
                                & HOTA    & DetA    & AssA    & MOTA    & CbyT & Relative error (\%)  & Median \\ \hline
YOLOv5l (score $> 0.7$)         & 0.56039 & 0.48968 & 0.64591 & 0.58540 & 1,183 & \textbf{1.25\%} & \textbf{6.95\%} \\ 
YOLOv3  (score $> 0.5$)         & 0.51003 & 0.43885 & 0.59608 & 0.54786 & 1,153 & 3.76\% &  8.45\% \\ 
YOLOv8l (score $> 0.6$)         & 0.54216 & 0.46512 & 0.63690 & 0.56415 & 1,128 & 5.84\% &  9.40\% \\ 
YOLOv7l (score $> 0.6$)         & 0.52306 & 0.45016 &	0.61086 & 0.53131 & 1,140 & 4.84\% & 10.24\% \\ 
EfficientDet B3 (score $> 0.5$) & 0.55564 & 0.47085 & 0.66561 & 0.56775 & 1,086 & 9.34\% & 11.82\% \\ \hline
\end{tabular}
\label{tab:tracking:results}
\end{table}

\begin{table}[]
\caption{Tracking results per frame sequence in \textsc{MOrangeT} for the two best detection models: YOLOv5l and YOLOv3.}
    \centering
    \footnotesize
\begin{tabular}{llrrrrrrrr}
        Model             & Sequence & HOTA    & DetA    & AssA    & MOTA     & CbyT & CbyT-GT & Relative error (\%) & Median \\ \hline
\multirow{13}{*}{YOLOv5l} & V01      & 0.37662 & 0.32271 & 0.47037 & 0.06078  & 111  & 110     & 0.91\%              &  \multirow{13}{*}{6.95\%}\\
                          & V02      & 0.34375 & 0.24189 & 0.48938 & 0.17777  & 34   & 46      & 26.09\%             \\
                          & V03      & 0.43321 & 0.31249 & 0.60213 & 0.34024  & 81   & 90      & 10.00\%             \\
                          & V04      & 0.71576 & 0.69602 & 0.73719 & 0.84919  & 109  & 105     & 3.81\%              \\
                          & V05      & 0.60224 & 0.60302 & 0.60454 & 0.72446  & 151  & 148     & 2.03\%              \\
                          & V06      & 0.51853 & 0.45734 & 0.59217 & 0.55306  & 146  & 122     & 19.67\%             \\
                          & V07      & 0.66271 & 0.63850 & 0.68908 & 0.78735  & 148  & 154     & 3.90\%              \\
                          & V08      & 0.55260 & 0.48412 & 0.63734 & 0.62637  & 185  & 192     & 3.65\%              \\
                          & V09      & 0.36263 & 0.22681 & 0.58042 & 0.28029  & 10   & 16      & 37.50\%             \\
                          & V10      & 0.23916 & 0.12979 & 0.44391 & -0.02332 & 15   & 13      & 15.38\%             \\
                          & V11      & 0.52879 & 0.45788 & 0.61570 & 0.58926  & 77   & 87      & 11.49\%             \\
                          & V12      & 0.53879 & 0.46801 & 0.62259 & 0.59311  & 116  & 115     & 0.87\%              \\
                      & \textbf{All} & 0.56039 & 0.48968 & 0.64591 & 0.58540  & 1183 & 1198    & \textbf{1.25\%}     \\ \hline
\multirow{13}{*}{YOLOv3}  & V01      & 0.33805 & 0.27482 & 0.42379 & 0.20005  & 95   & 110     & 13.64\%             &  \multirow{13}{*}{8.45\%} \\
                          & V02      & 0.36436 & 0.25440 & 0.52319 & 0.27699  & 33   & 46      & 28.26\%             \\
                          & V03      & 0.41799 & 0.32480 & 0.53953 & 0.34516  & 94   & 90      & 4.44\%              \\
                          & V04      & 0.59680 & 0.55379 & 0.64500 & 0.71497  & 111  & 105     & 5.71\%              \\
                          & V05      & 0.58442 & 0.56636 & 0.60592 & 0.73024  & 152  & 148     & 2.70\%              \\
                          & V06      & 0.52529 & 0.46520 & 0.59506 & 0.56571  & 139  & 122     & 13.93\%             \\
                          & V07      & 0.53014 & 0.48528 & 0.58142 & 0.64985  & 136  & 154     & 11.69\%             \\
                          & V08      & 0.50786 & 0.43300 & 0.60138 & 0.56000  & 175  & 192     & 8.85\%              \\
                          & V09      & 0.31171 & 0.16878 & 0.57590 & 0.18565  & 7    & 16      & 56.25\%             \\
                          & V10      & 0.30117 & 0.18895 & 0.48078 & 0.05758  & 14   & 13      & 7.69\%              \\
                          & V11      & 0.50065 & 0.41959 & 0.60027 & 0.52388  & 80   & 87      & 8.05\%              \\
                          & V12      & 0.54416 & 0.47418 & 0.62598 & 0.60044  & 117  & 115     & 1.74\%              \\
                      & \textbf{All} & 0.51003 & 0.43885 & 0.59609 & 0.54786  & 1153 & 1198    & \textbf{3.76\%}     \\ \hline
\end{tabular}
\label{tab:tracking:results:pervid}
\end{table}

\begin{figure}
     \centering
     \begin{subfigure}[b]{0.18\textwidth}
         \centering
         \includegraphics[width=\textwidth]{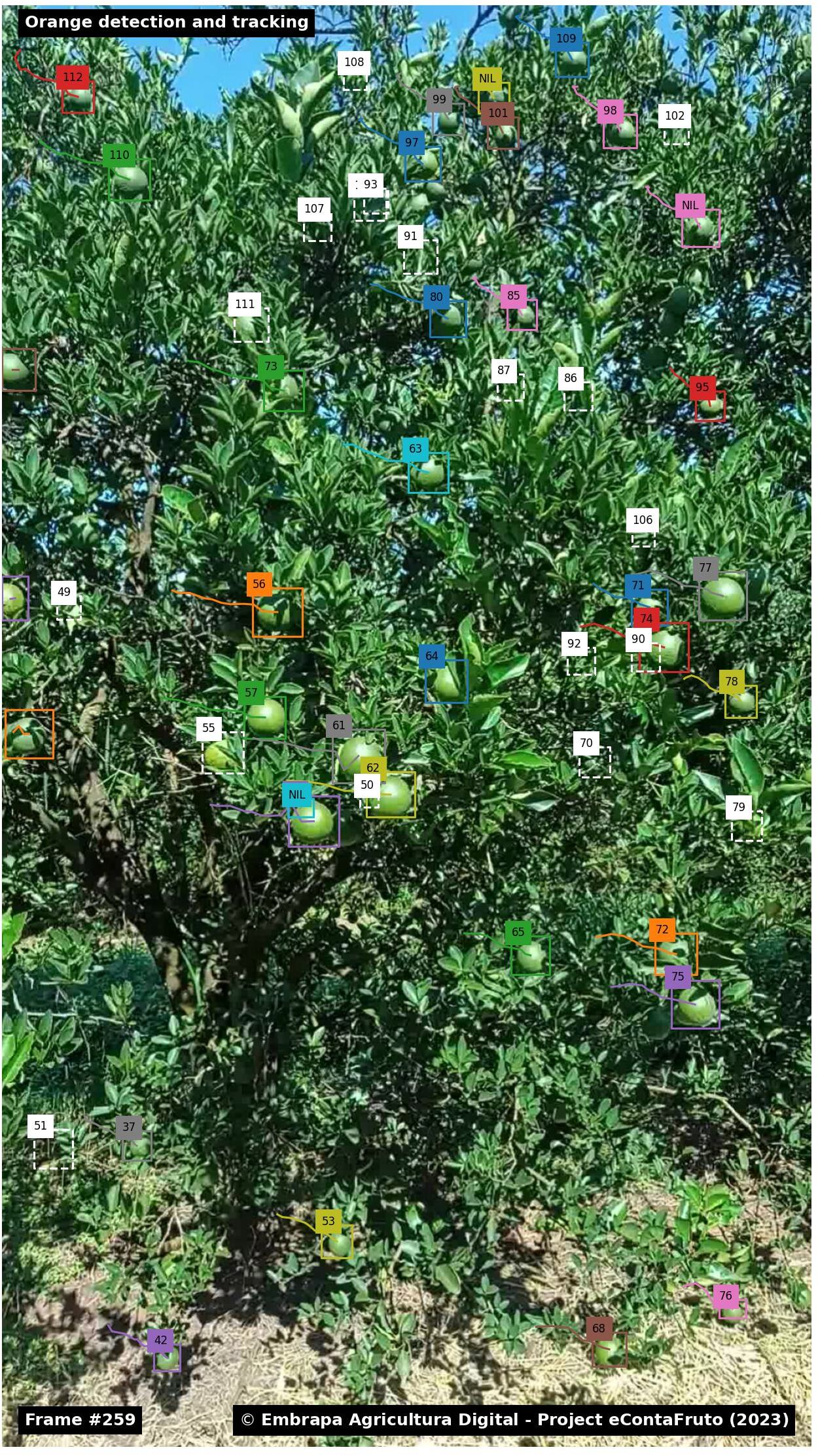}
         \caption{V07, $f_{260}$}
         \label{fig:yolov5:V07:frame260}
     \end{subfigure}  
     \hfill
     \begin{subfigure}[b]{0.18\textwidth}
         \centering
         \includegraphics[width=\textwidth]{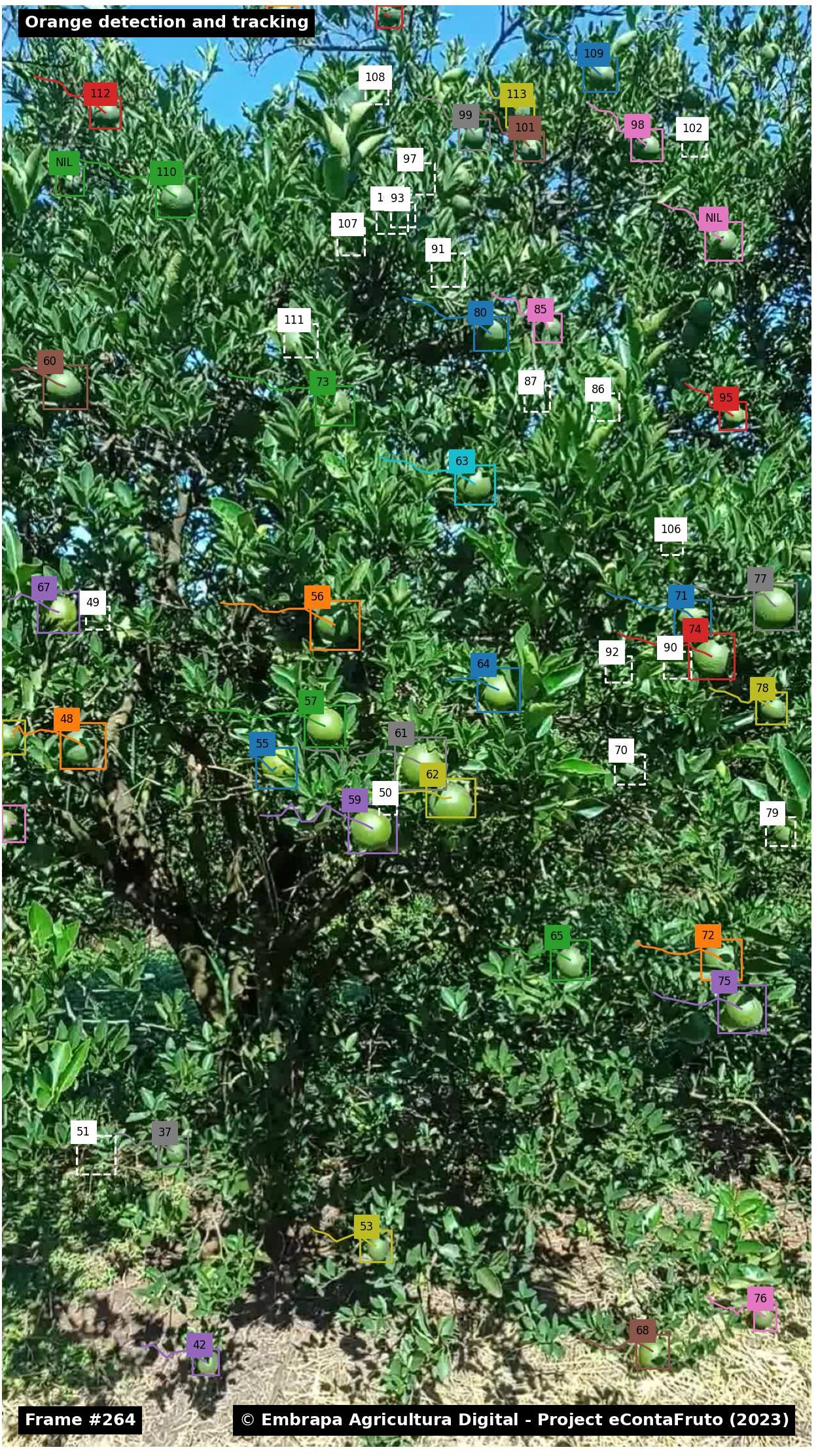}
         \caption{V07, $f_{265}$}
         \label{fig:yolov5:V07:frame265}
     \end{subfigure}  
     \hfill
     \begin{subfigure}[b]{0.18\textwidth}
         \centering
         \includegraphics[width=\textwidth]{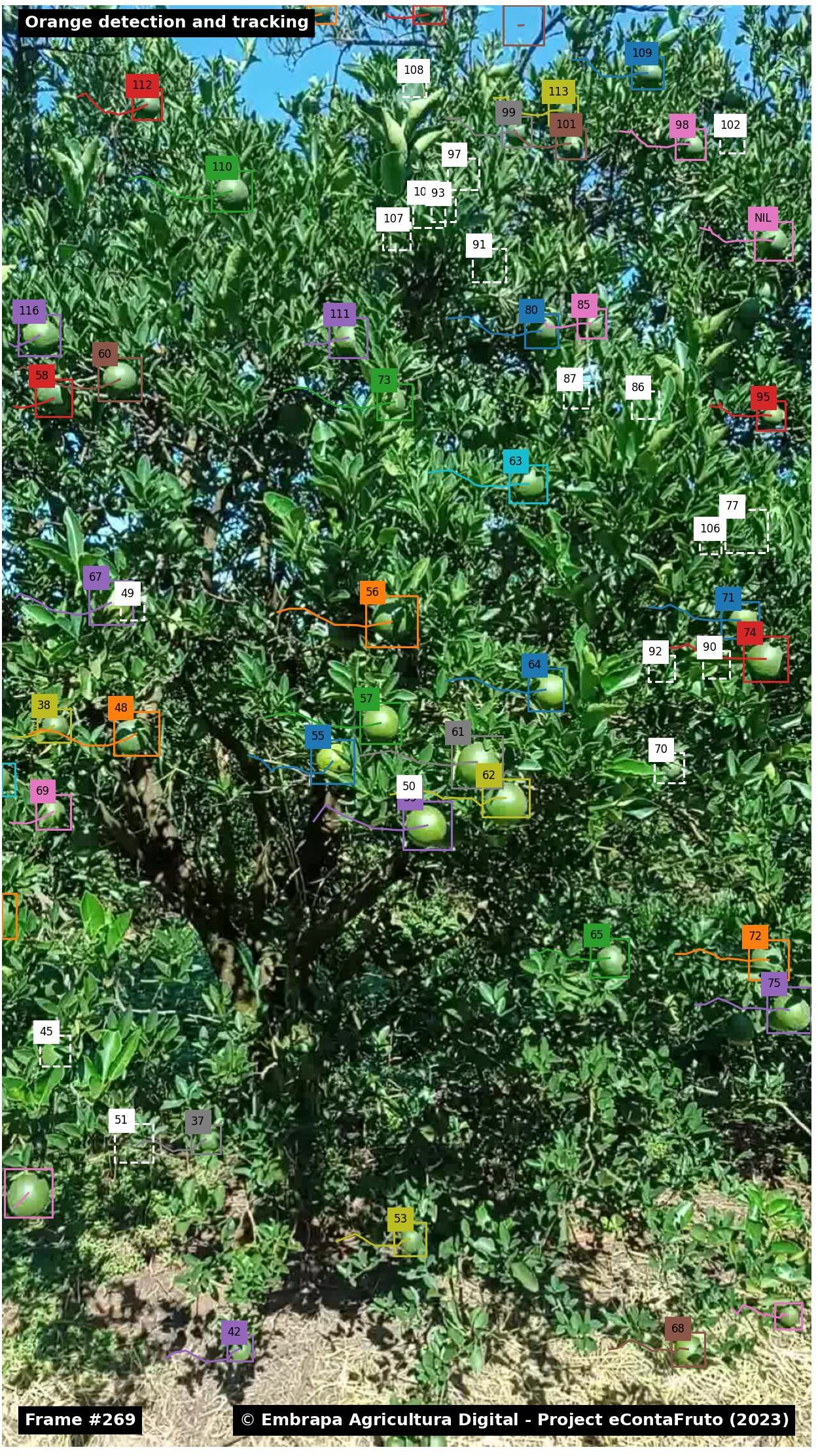}
         \caption{V07, $f_{270}$}
         \label{fig:yolov5:V07:frame270}
     \end{subfigure}  
     \hfill
     \begin{subfigure}[b]{0.18\textwidth}
         \centering
         \includegraphics[width=\textwidth]{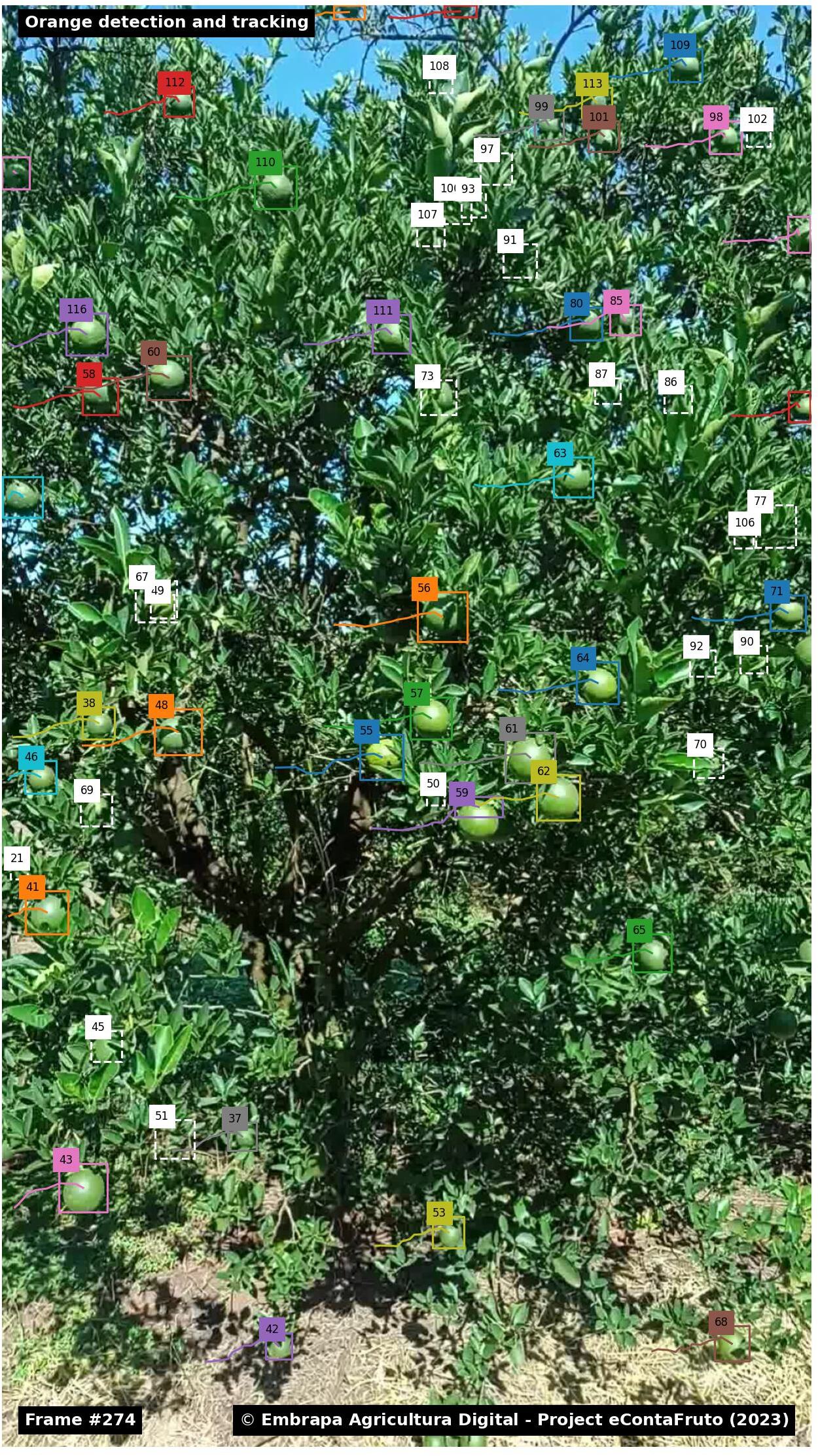}
         \caption{V07, $f_{275}$}
         \label{fig:yolov5:V07:frame275}
     \end{subfigure}  
     \hfill
     \begin{subfigure}[b]{0.18\textwidth}
         \centering
         \includegraphics[width=\textwidth]{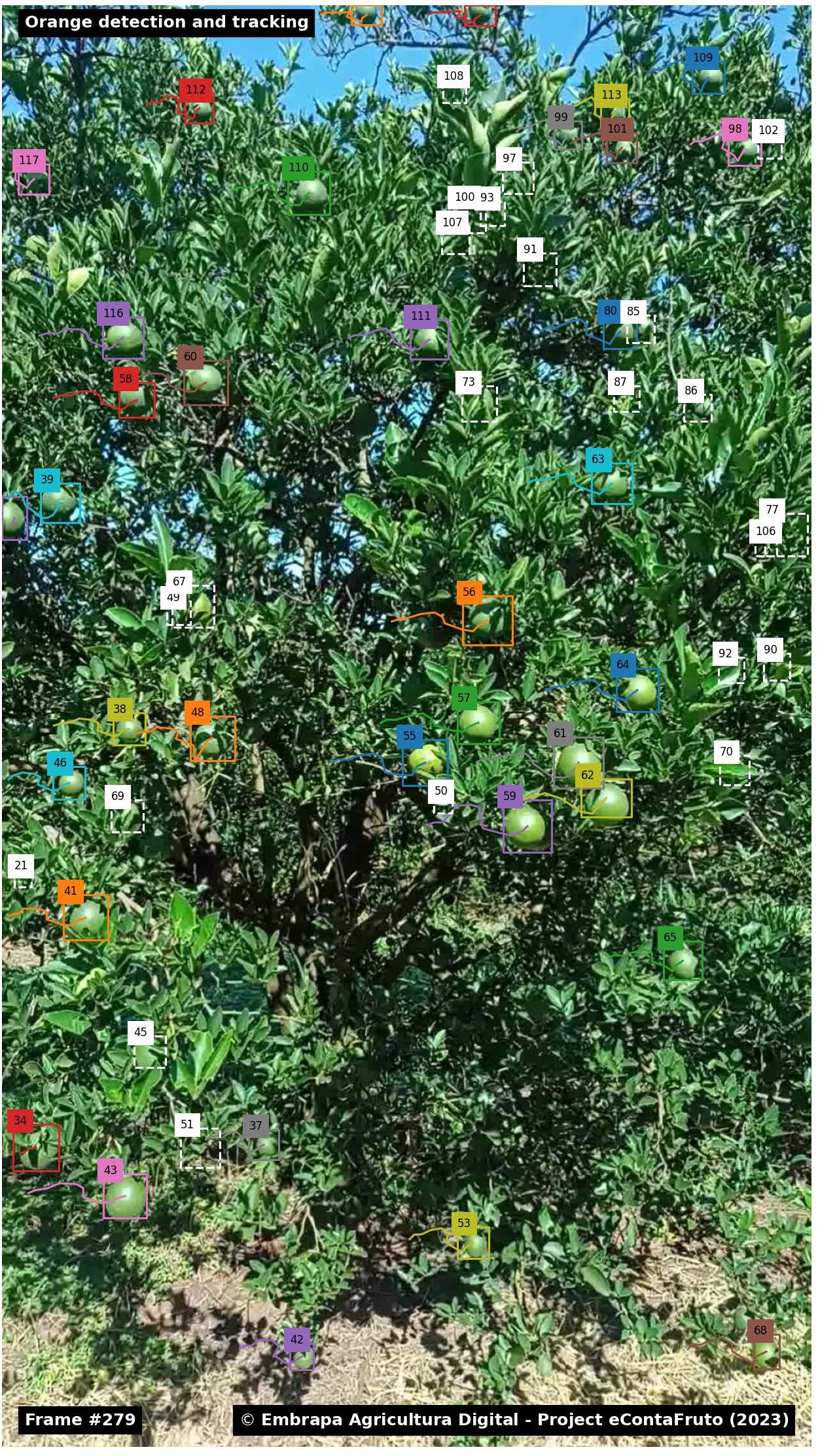}
         \caption{V07, $f_{280}$}
         \label{fig:yolov5:V07:frame280}
     \end{subfigure}  
     \begin{subfigure}[b]{0.18\textwidth}
         \centering
         \includegraphics[width=\textwidth]{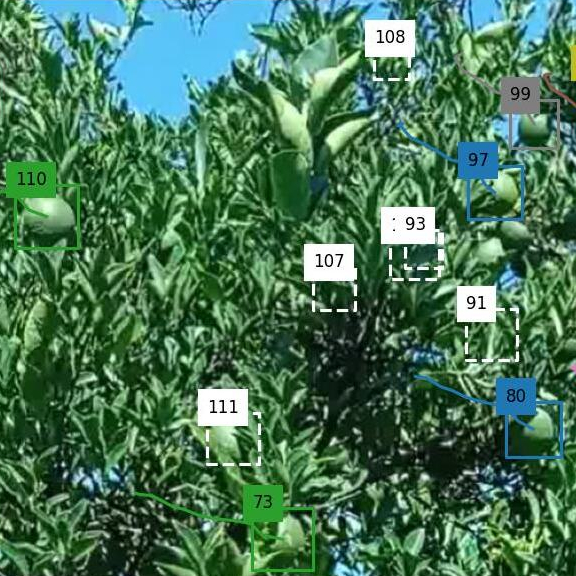}
         \caption{V07, $f_{260}$ (detail)}
         \label{fig:yolov5:V07:frame260:detail}
     \end{subfigure}  
     \hfill
     \begin{subfigure}[b]{0.18\textwidth}
         \centering
         \includegraphics[width=\textwidth]{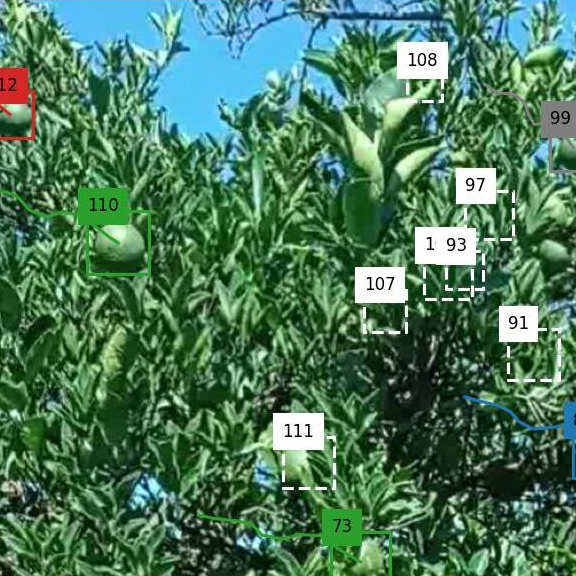}
         \caption{V07, $f_{265}$ (detail)}
         \label{fig:yolov5:V07:frame265:detail}
     \end{subfigure}  
     \hfill
     \begin{subfigure}[b]{0.18\textwidth}
         \centering
         \includegraphics[width=\textwidth]{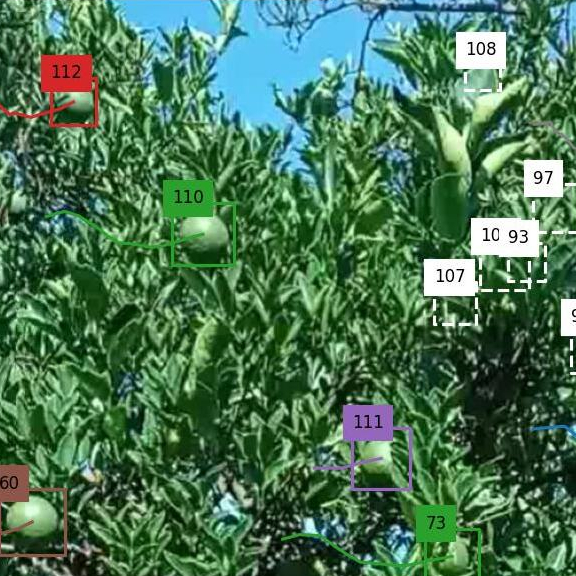}
         \caption{V07, $f_{270}$ (detail)}
         \label{fig:yolov5:V07:frame270:detail}
     \end{subfigure}  
     \hfill
     \begin{subfigure}[b]{0.18\textwidth}
         \centering
         \includegraphics[width=\textwidth]{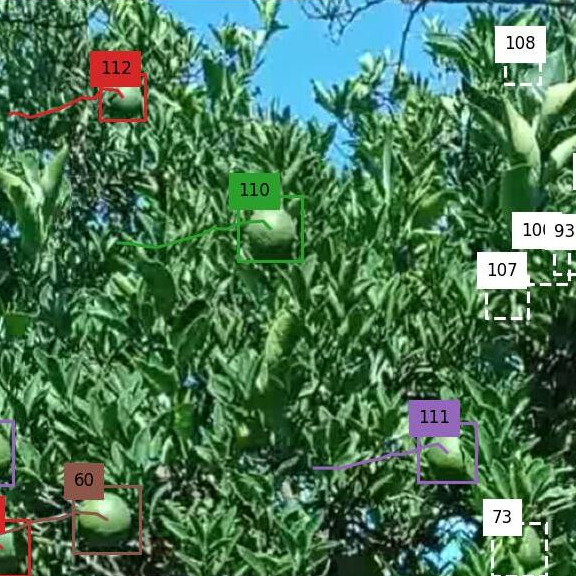}
         \caption{V07, $f_{275}$ (detail)}
         \label{fig:yolov5:V07:frame275:detail}
     \end{subfigure}  
     \hfill
     \begin{subfigure}[b]{0.18\textwidth}
         \centering
         \includegraphics[width=\textwidth]{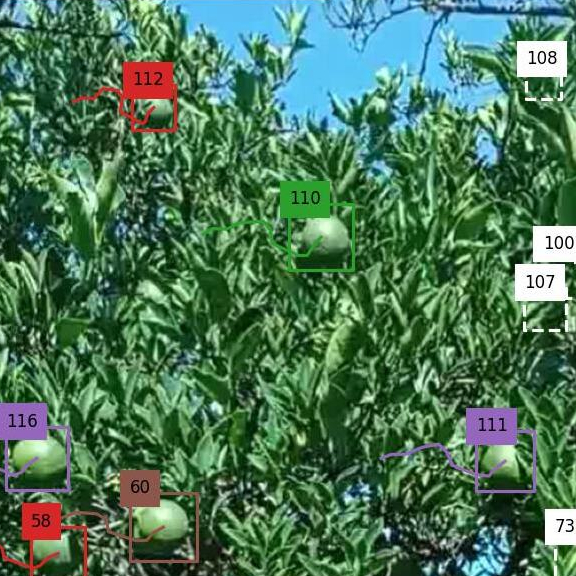}
         \caption{V07, $f_{280}$ (detail)}
         \label{fig:yolov5:V07:frame280:detail}
     \end{subfigure}  
     \begin{subfigure}[b]{0.18\textwidth}
         \centering
         \includegraphics[width=\textwidth]{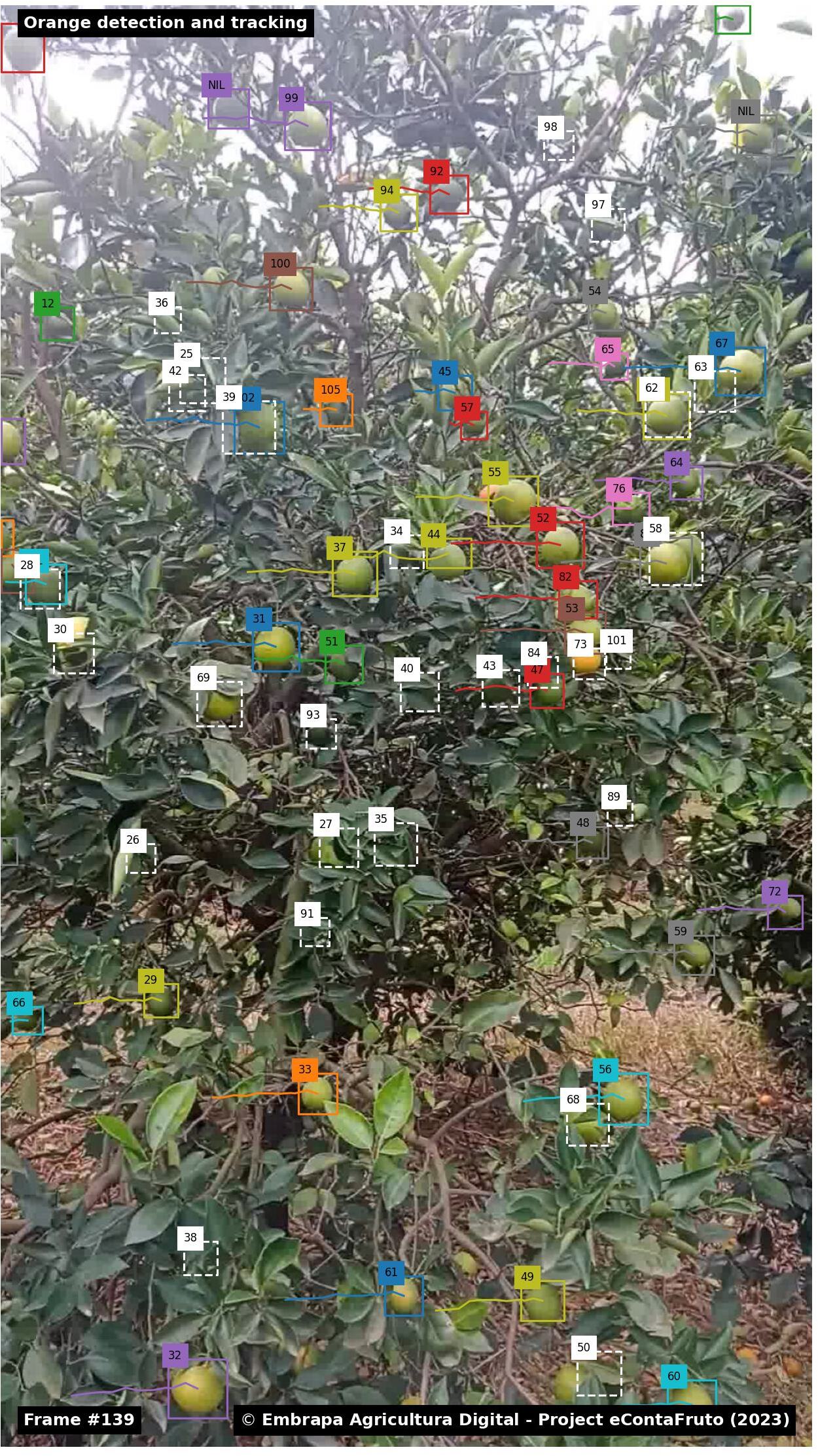}
         \caption{V12, $f_{140}$}
         \label{fig:yolov5:V12:frame140}
     \end{subfigure}  
     \hfill
     \begin{subfigure}[b]{0.18\textwidth}
         \centering
         \includegraphics[width=\textwidth]{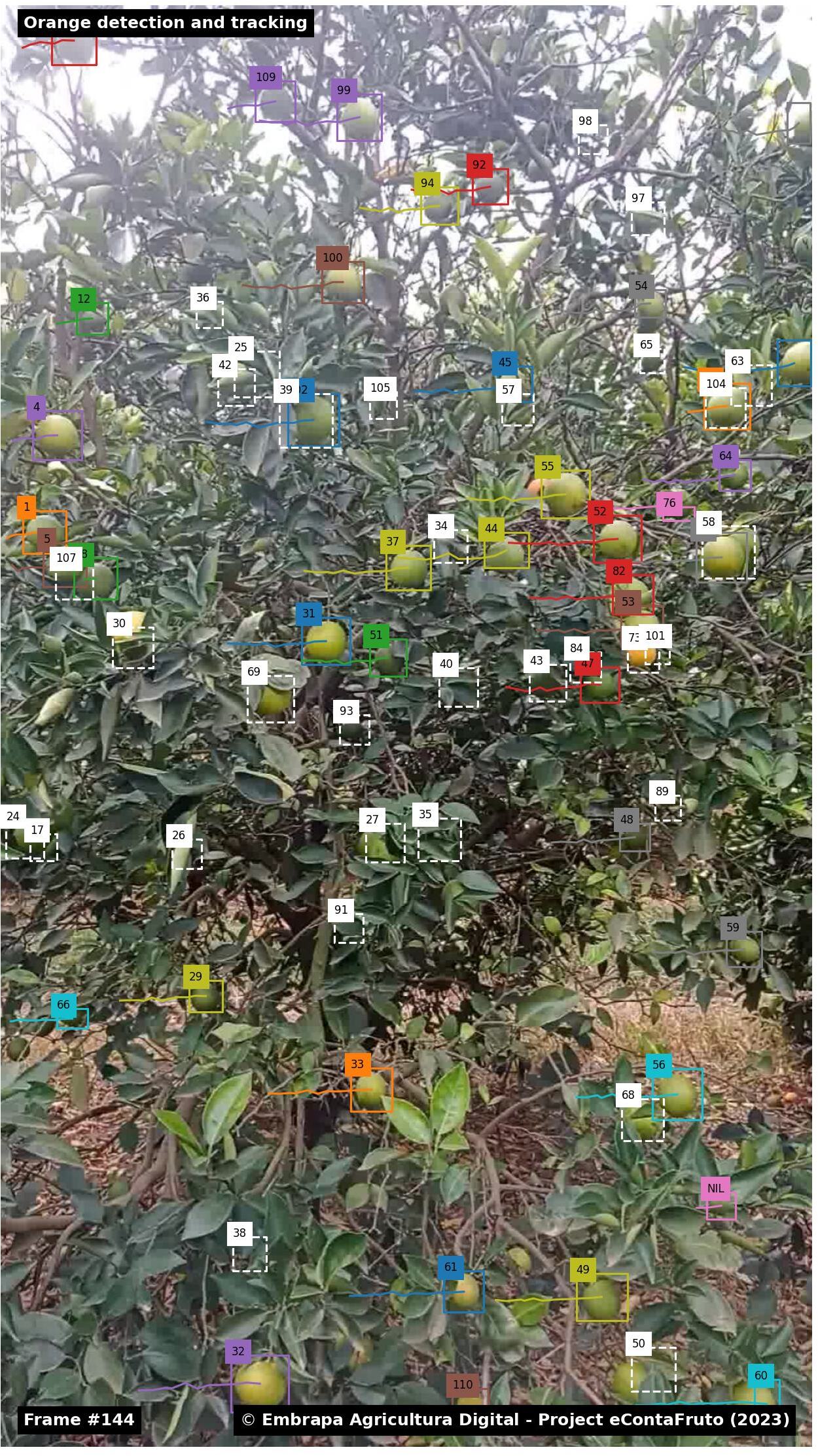}
         \caption{V12, $f_{145}$}
         \label{fig:yolov5:V12:frame145}
     \end{subfigure}  
     \hfill
     \begin{subfigure}[b]{0.18\textwidth}
         \centering
         \includegraphics[width=\textwidth]{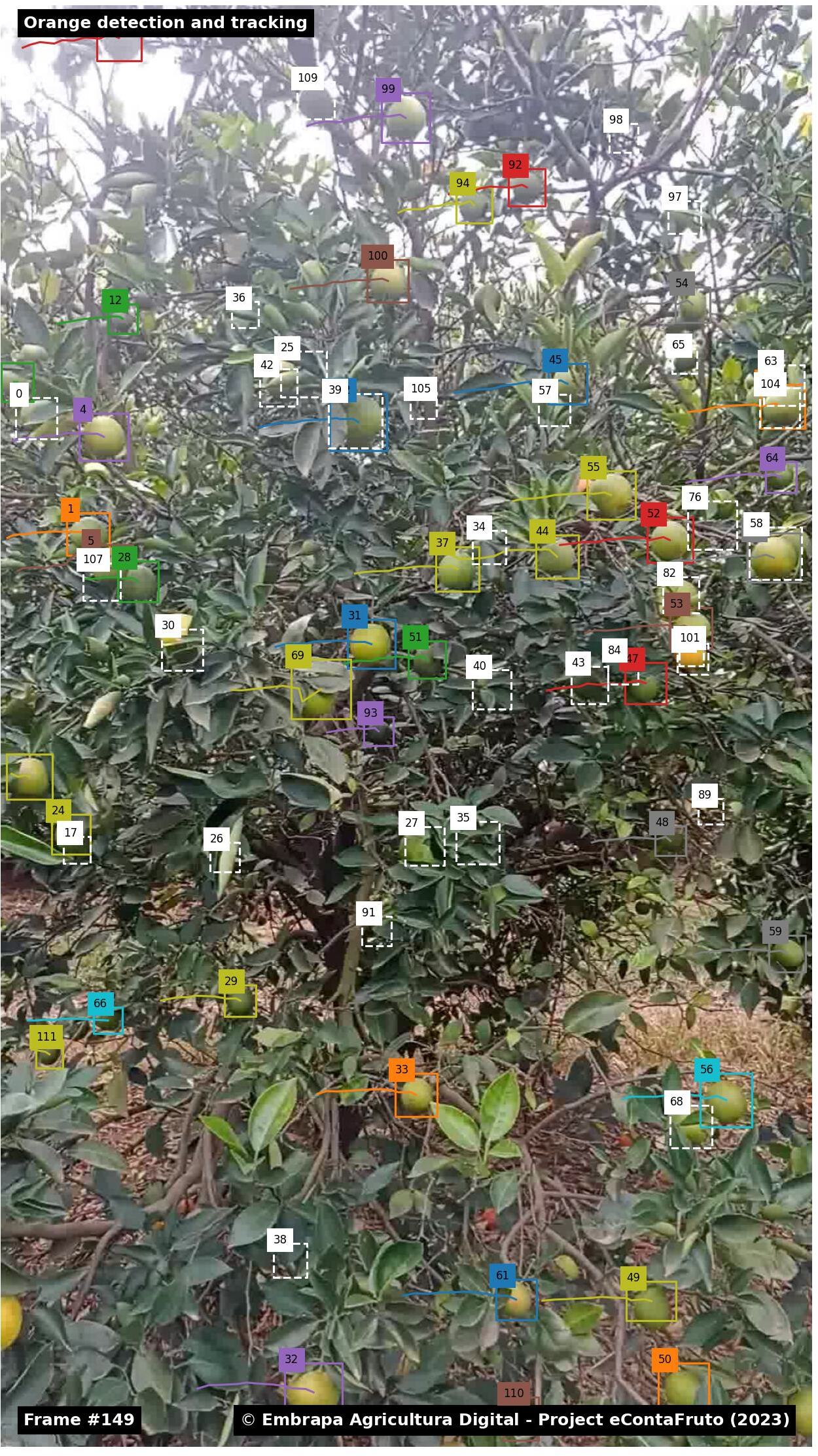}
         \caption{V12, $f_{150}$}
         \label{fig:yolov5:V12:frame150}
     \end{subfigure}  
     \hfill
     \begin{subfigure}[b]{0.18\textwidth}
         \centering
         \includegraphics[width=\textwidth]{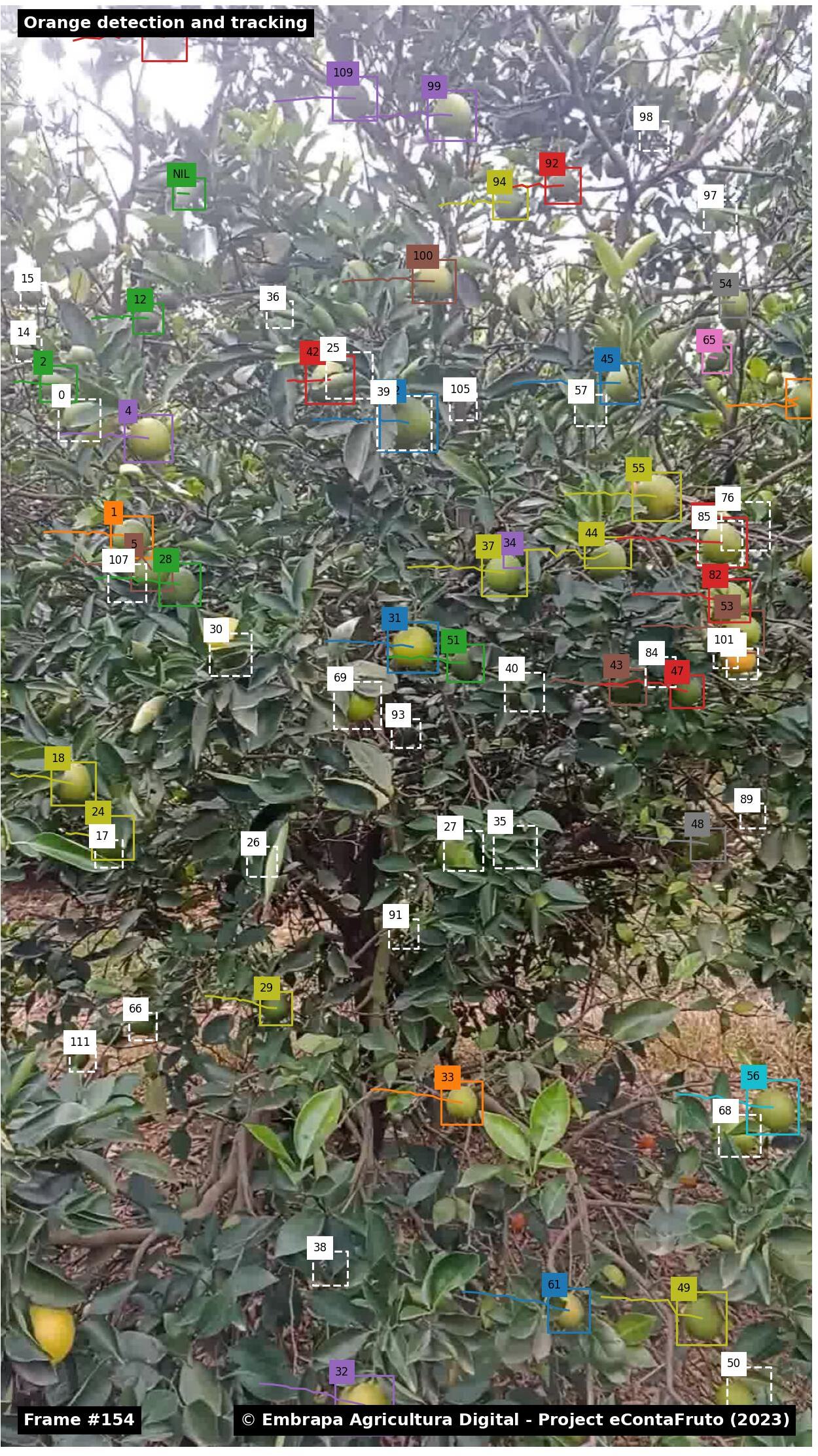}
         \caption{V12, $f_{155}$}
         \label{fig:yolov5:V12:frame155}
     \end{subfigure}  
     \hfill
     \begin{subfigure}[b]{0.18\textwidth}
         \centering
         \includegraphics[width=\textwidth]{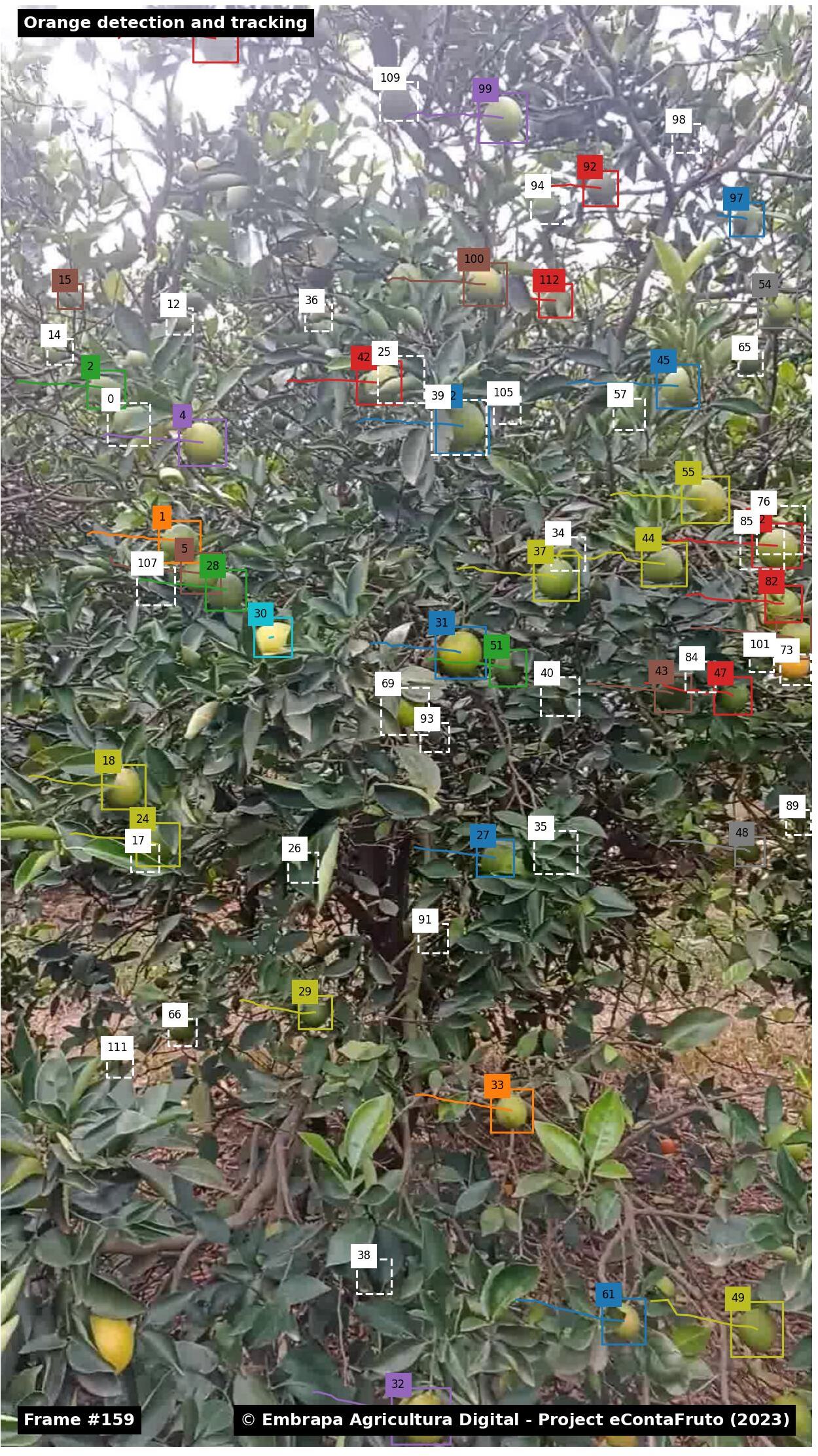}
         \caption{V12, $f_{160}$}
         \label{fig:yolov5:V12:frame160}
     \end{subfigure}       
     \caption{Tracking results for detection inputs from YOLOv5l. Tracks whose oranges' 3-D models were estimated present numerics IDs. New tracks whose oranges were not yet estimated successfully presents the marker \textsc{Nil}. Tracks marked by dashed white lines are \textsc{Lost} tracks. (a-j) Frames from V07 (\emph{Valencia}). (k-o) Frames from V12 (\emph{Hamlin}). 
 Best seen in digital format.} 
     \label{fig:res-tracking-yolov5l}
\end{figure}

Figure~\ref{fig:res-tracking-yolov5l} shows tracking examples for two videos, V07 and V12, the same previously seen in Figure~\ref{fig:res-tracking-08}
for comparison. The tracking was performed using detections from the YOLOv5l model. Again, when the orange position and ray were estimated by 
Algorithm 2, the numeric ID for the track (fruit) is displayed, and tracks not yet presenting a successful estimation are marked with a \textsc{Nil} label.
\textsc{Lost} tracks are displayed as white, dashed boxes. A relocalization example can be seen for track 111 in V07: the track is lost at frames $f_{260}$ 
and $f_{265}$, but the track became ACTIVE at $f_{270}$. A compilation of the results in a video demonstration is available on-line\footnote{\url{https://youtu.be/hOq42KMskLQ}}.

\begin{figure}
	\centering
	\includegraphics[width=\textwidth]{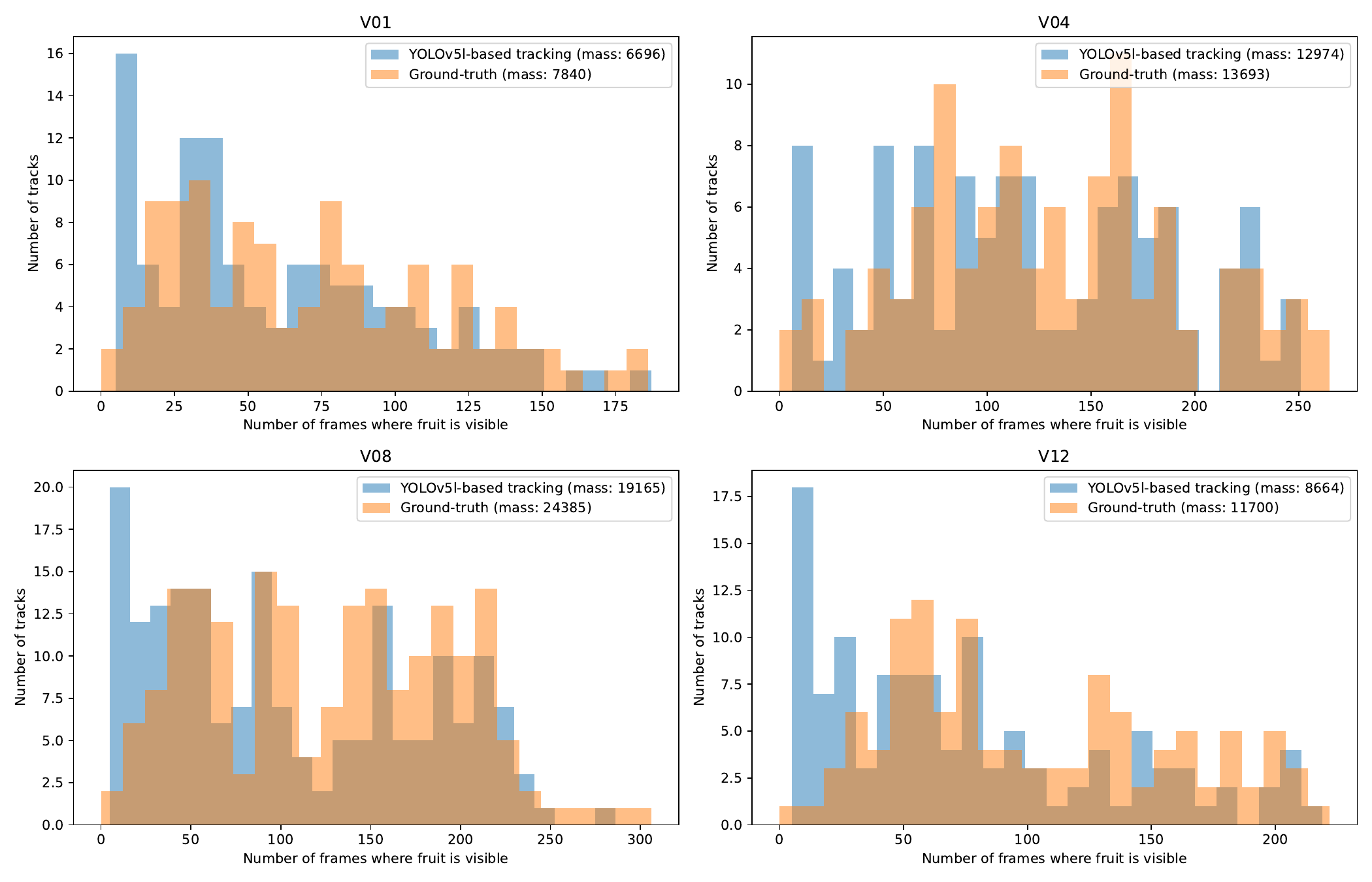}
	\caption{Histograms showing the accumulated number of oranges appearances, contrasting the ground-truth (orange histograms)
 and the pipeline tracking results (blue histograms) for four frame sequences in \textsc{MOrangeT}. More mass in the histograms
 means more appearances of fruits.}
	\label{fig:track_len_hist}
\end{figure}

\begin{figure}
	\centering
	\includegraphics[width=0.48\textwidth]{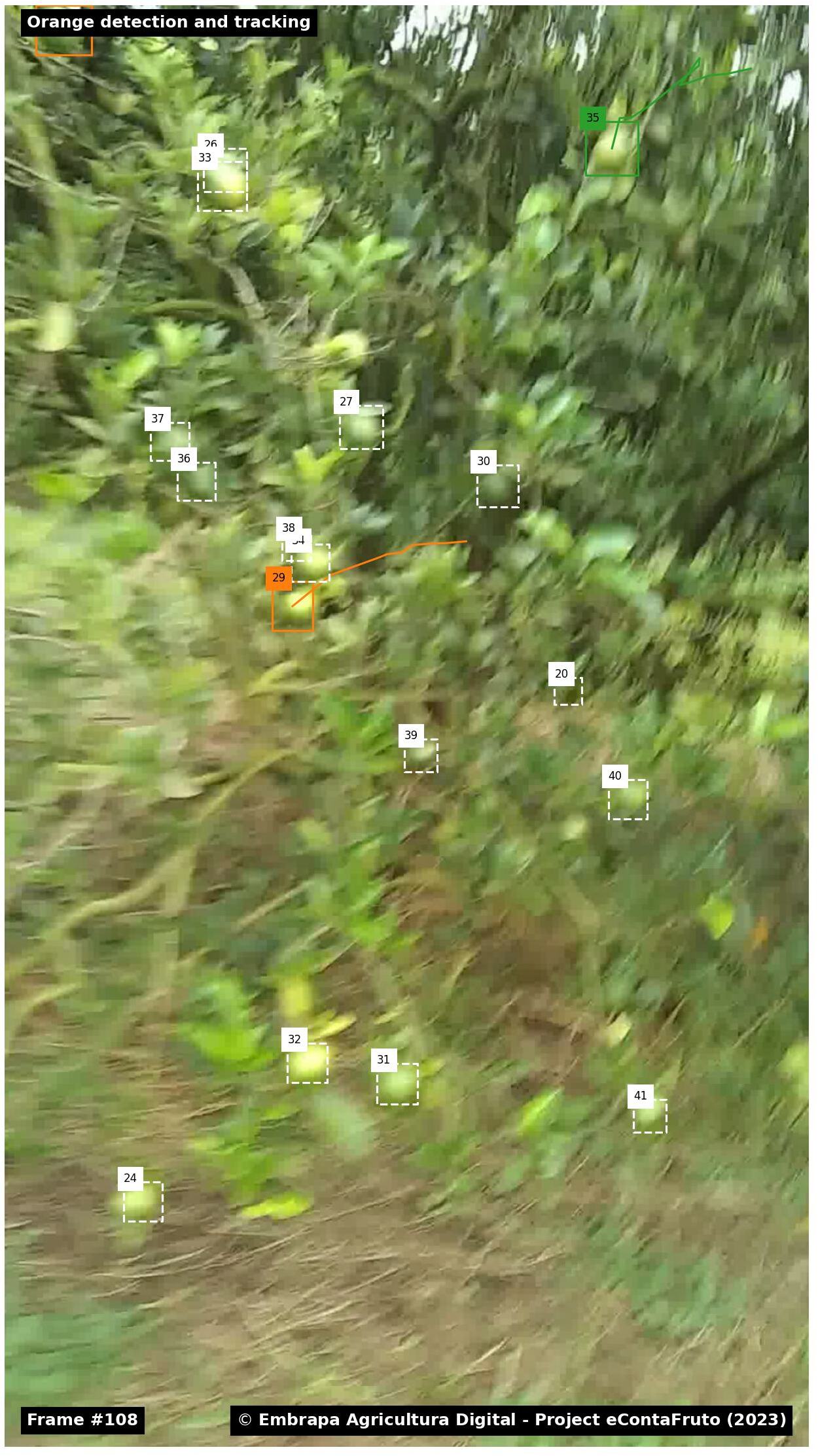}
    \includegraphics[width=0.48\textwidth]{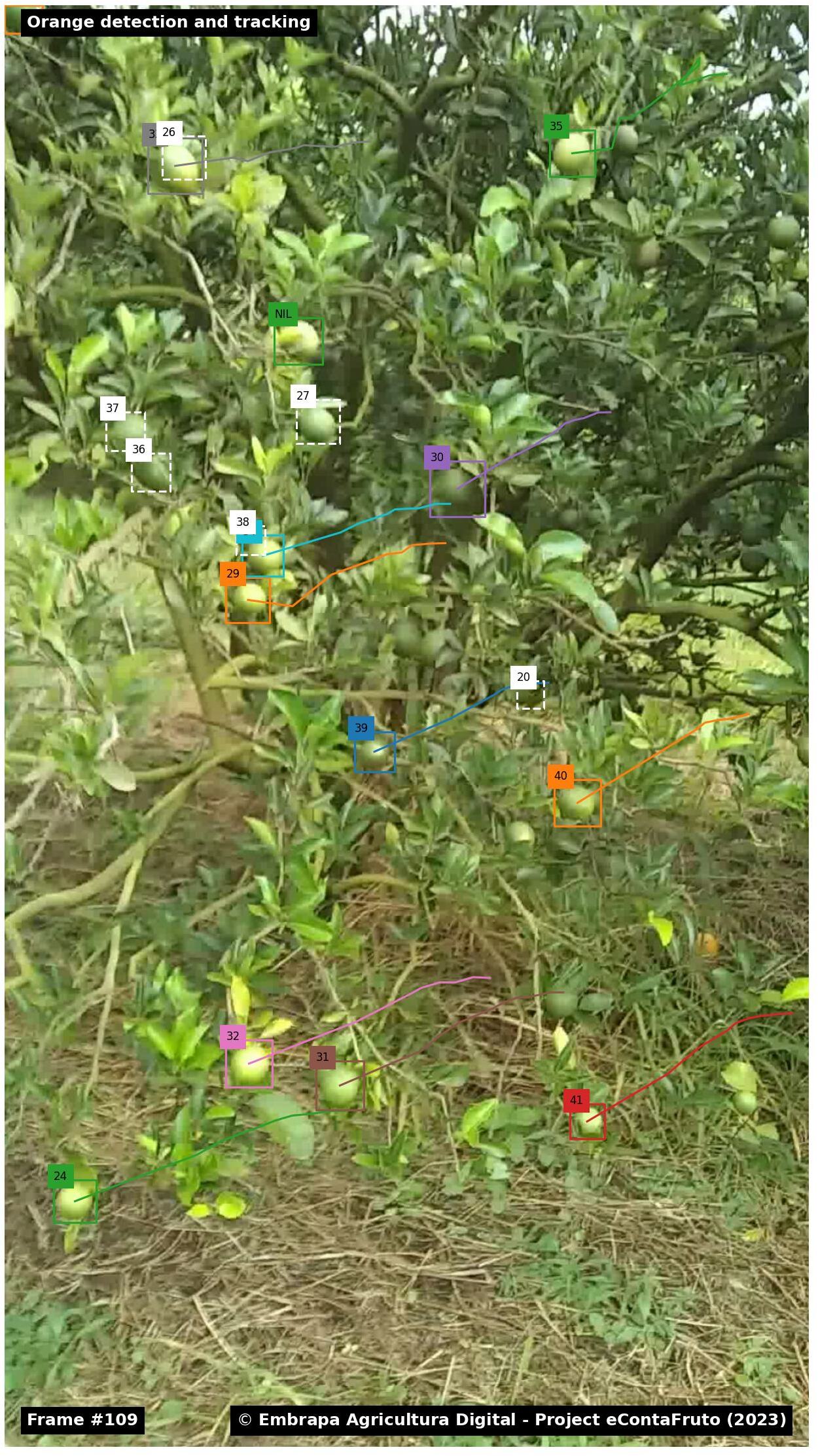}
	\caption{Fruits in \textsc{Lost} state during tracking (dashed white boxes) in frame~108 from V01. The misdetections are caused by the motion blur in
 the frame. Our tracking module follows the tracks, keeping them in \textsc{Lost} (non-visible) state, while ground-truth data 
 considers such tracks visible fruits. Part of the tracks are relocalized in the next frame~109. Best seen in digital format.}
	\label{fig:lost_fruits}
\end{figure}

Perfect tracking implies exact counting and a HOTA value equal to 1. As seen in Table~\ref{tab:sensitivity}, high HOTA (or MOTA) 
values are associated with low errors in counting. However, we can observe low errors in counting even for mediocre values of
HOTA (or MOTA), around 0.5, as seen in Table~\ref{tab:tracking:results}. Actually, multiple fruit tracking is a \emph{harder}
problem than fruit counting assessment. If the pipeline can track the fruit in \emph{part} of its trajectory and does not merge
tracks of different fruits, it can reach accurate counting values. But losing parts of the fruit' trajectories will severely
penalize the tracking quality measures. If the tracker is losing part of the trajectories, it is missing the orange appearances
in some frames. Figure~\ref{fig:track_len_hist} shows histograms for four frame sequences in \textsc{MOrangeT},
considering the number of appearances for each fruit in the ground-truth (orange colored histogram) and in the tracker results
(blue colored histogram). More ``mass'' in bins on the right side means the fruits are observed
non-occluded in more frames. Comparing the ground-truth against the tracking results (using YOLOv5l detection in 
this example), we can see that our pipeline missed parts of the tracks, even considering accurate counting: the relative counting 
errors are 0.91\%, 3.81\%, 3.65\% and 0.87\% for V01, V04, V08 and V12, respectively, but the missed appearances will degrade
HOTA and MOTA values. Our pipeline can keep fruits in the \textsc{Lost} state, passive of relocalization, so the oranges, 
even missed in some frames, are properly accounted. Figure~\ref{fig:lost_fruits} shows an example where motion blur jeopardized 
detection: the fruits are kept in the \textsc{Lost} state, but HOTA and MOTA values decrease because ground-truth 
consider them visible oranges (note as part fruits are properly relocalized in the next frame). 

\subsection{Yield regressor results}\label{sec:yiel:results}

The sample analyzed consisted of the 1,197 plants that resulted from the previous steps of the pipeline. However, some plants had to 
be excluded due to missing plant dimensions ($W$, $H$ or $D$ in Table~\ref{tab:dataregressor}) or lack of automatic fruit counting, 
resulting in a reduced set of 1,139 plants. The regressor was specifically designed to estimate the sum of fruits from the first to 
the third flowering, namely $F1 + F2 + F3$, as the fruits from the fourth flowering are too small to be detected in the previous 
stages of the pipeline. 

\begin{figure}
	\centering
	\includegraphics[width=0.35\textwidth]{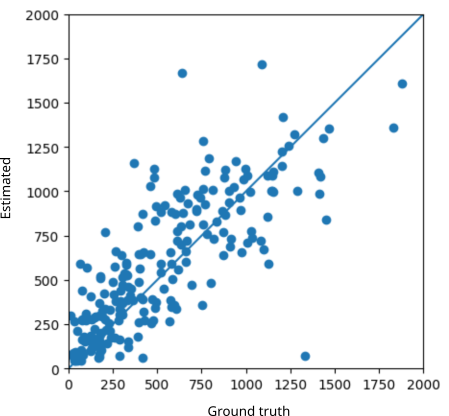}
	\caption{Processing results considering the entire dataset, achieving performance of $R^2 =0.61$.}
	\label{fig:alldata}
\end{figure}

Consider the counting from our tracking system fed by YOLOv3 orange detections. Out of the 1,139 fruits, 911 were used for training 
of the neural network regressor, and the remaining 228 for testing were reserved for testing. The results are presented in 
Figure~\ref{fig:alldata}, where we can observe a significant dispersion of points in the graph. This dispersion affected the 
overall performance, resulting in a $R^2$ value of 0.61. Notably, there is an isolated point in the lower part of the graph where 
the regressor estimated around 50 fruits, while the ground truth falls within the range of 1,250 to 1,500. Conversely, there are 
two isolated points at the top of the graph where the regressor overestimated the values. For these points, the regressor estimated 
values between 1,500 and 1,750, while the actual values are around 700 for the left point and around  1,200 for the right point.

During our analysis of the full dataset, which includes the fruit counting from videos captured on each side of the plant (CbyT-A + 
CbyT-B), the ground truth ($F1 + F2 + F3$), and the yield estimated by the regressor, we observed that a significant number of 
samples had a much lower number of counted visible fruits (CbyT-A + CbyT-B) compared to the sum $F1 + F2 + F3$. In fact, most of 
these samples lie below the 40\% detection rate mentioned in Table~\ref{tab:sensitivity}. 

This analysis led us to develop the hypothesis that implementing an acceptance threshold for the videos, based on the ratio between detected versus ground truth fruits, could potentially improve the results of the regressor. To test this hypothesis, we established thresholds for the acquisition process.  Initially, we considered only videos with a minimum identification rate of 20\% of the ground truth fruits. This new experiment reduced the dataset to 741 plants, with 148 plants reserved for testing purposes. 

\begin{figure}
     \centering
     \begin{subfigure}[b]{0.35\textwidth}
         \centering
         \includegraphics[width=\textwidth]{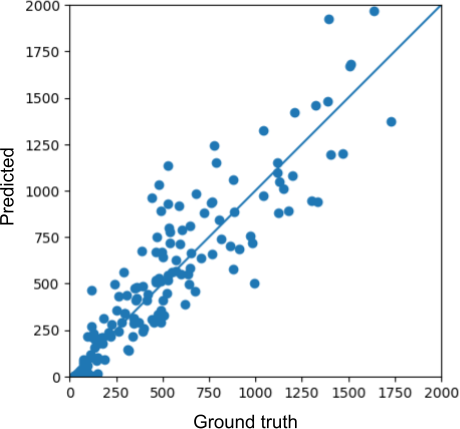}
         \caption{20\% threshold, 148 trees, $R^2 = 0.79$.}
         \label{fig:yield:YOLOv3:thresh20}
     \end{subfigure}
     \begin{subfigure}[b]{0.35\textwidth}
         \centering
         \includegraphics[width=\textwidth]{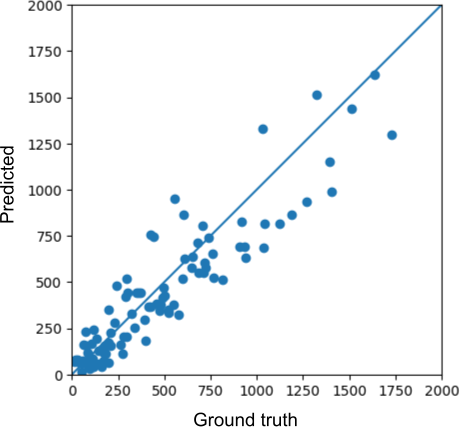}
         \caption{30\% threshold, 98 trees, $R^2 = 0.85$.}
         \label{fig:yield:YOLOv3:thresh30}
     \end{subfigure}  
     \caption{Yield regression, subject to different thresholds for fruit counting.  (a) Ensuring at least 20\% of the ground truth identified by counting (148 trees in the test set), $R^2 = 0.79$. (b) Ensuring at least 30\% of the ground truth identified by counting (98 trees in the test set), $R^2 = 0.85$. The counting was performed by tracking from YOLOv3.} 
     \label{fig:yield:YOLOv3:thresh}
\end{figure}

The results of this experiment showed a significant improvement, with the $R^2$ value increasing to 0.79, as illustrated in 
Figure~\ref{fig:yield:YOLOv3:thresh20}. We can also observe a reduced dispersion around the diagonal line, indicating better performance. 
This improved value of the $R^2$ suggests that our neural network regressor was able to explain a greater portion of the variance 
in yield estimation when considering this reduced dataset. 

Furthermore, when we raised the threshold to require the detection of at least 30\% of the ground truth (still below the 40\% in 
Table~\ref{tab:sensitivity}), the $R^2$ value further improved to 0.85, as shown in Figure~\ref{fig:yield:YOLOv3:thresh30}. This new experiment reduced 
the dataset to 558 trees, with 98 plants reserved for testing purposes, seen in Figure~\ref{fig:yield:YOLOv3:thresh30}. These results clearly 
indicate the performance of the regressor is closely tied to the quality of the machine learning identification and fruit counting results. It is 
important to acknowledge that it is not possible to identify 100\% of the fruits through an automatic counting process, as some fruits are located 
inside the canopies, not visible by camera in any pose. 

\begin{figure}
	\centering
	\includegraphics[width=0.35\textwidth]{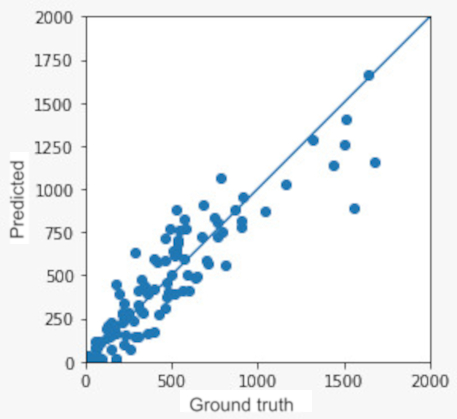}
	\caption{Yield regression for a test set of 111 trees, subject to a threshold of at least 30\% of the ground truth identified by counting. The counting was performed by tracking from YOLOv5l detections. The yield regression reached $R^2 = 0.82$.}
	\label{fig:yield:YOLOv5l:thresh30}
\end{figure}

We observe the same behavior when considering the counting produced using tracking from YOLOv5l detections. In this case, requiring detection 
of at least 30\% of the ground truth, we got a set of 668 trees. Of these, 557 were chosen for training the regressor, resulting in a $R^2$
value of 0.82 for the test set, which comprised the remaining 111 trees. A scatter plot of this result is seen in 
Figure~\ref{fig:yield:YOLOv5l:thresh30}.

%
%
\section{Discussion}
\label{sec:discussion}

The complex nature of the natural environment, coupled with the variability in fruit farming practices, continues to pose significant 
challenges for automating both monitoring and harvesting. As noted by \cite{he2022fruit}, achieving automated operations in complex 
agricultural fields requires a synergistic approach that combines engineering and standardization of agricultural management practices.
Such integration is crucial to enable future automated operations and management. It is essential that breeders and agronomists adapt
the plants for automated operation. For example, the results by \citet{gao2022novel} show the clear advantages of systems like 
\emph{fruiting wall} on automated counting and yield estimation. Our pipeline reached significant results in a diverse set of orchard 
management, using hand-held video streams recorded by smartphones. We expect even better results in more controlled settings, like 
fruiting walls and artificial lightning at night operations, and employing improved hardware like wide-angle lenses, and 
visual-inertial cameras. The image capture procedure employed in this work is coupled to the distributed nature of the 
Fundecitrus' Orange Crop Forecast, performed by 30 field agents covering a large territory. Video recordings by smartphones were 
a practical, cheap solution, after other designs involving UAVs and cameras arrays being considered, and prior tests performed on field. 

The Crop Forecast employs a well-established methodology \citep{fundecitrus2022}, relying on a large set of individual trees rather 
than the scanning of entire rows or plots. One challenge arising from this tree-based approach is the difficulty in visually defining 
the boundaries of neighboring trees, leading to branches from one tree entering the canopy of another. While the Crop Forecast team 
can manage this issue through manual fruit stripping, enabling accurate attribution of oranges to the correct tree (albeit a laborious
task for the forecast staff), a vision-based system would find error-free fruit-to-tree assignment impractical. Other approaches, 
such as those integrating row-level or plot-level data and utilizing vehicle-mounted cameras, as demonstrated by \citet{zhang2022citrus}, 
could be considered. In row-level counting, exact fruit-to-tree attribution would not pose an issue.

Besides the large number of works on fruit detection and impressive results reached by neural networks-based systems, fruit 
detection is still challenging. Considering large datasets containing a representative set of images, and presenting  diversity in 
light conditions, sensors, noise, crop varieties and phenological stage, detection performance can yet be significantly improved. However,
accurate detection is necessary but not sufficient for accurate fruit counting, mainly because of occlusion issues. Better detection
models depend on large annotated datasets for training and evaluation, but the scarcity of such datasets remains a key bottleneck 
in developing the next-generation of intelligent systems for precision agriculture, as pointed by \citet{lu2020survey}. We hope that
\textsc{MOrangeT} and \textsc{OranDet} be significant contributions in this regard.

Kalman filter-based trackers \citep{bewley2016sort} can be considered an alternative to handle short-term fruit lost, 
but it is not as efficient as relocalization for long occlusion periods and changes in the movement direction. 
Our relocalization module can be considered an alternative to the motion displacement estimation proposed by 
\citet{zhang2022citrus}, specially useful for non-linear movements and non-planar fruit spatial distributions. 
\citet{zhang2022citrus} argues that SORT-like tracking \citep{bewley2016sort, wojke2018deepsort} cannot handle dense fruit 
spatial distribution in space and heavy overlap. We have shown that relocalization based on 3-D is a viable alternative to complement 
inter-frame fruit tracking by data association algorithms. Full 3-D reconstruction is not needed, just ego-motion, which can be 
obtained by other methods like visual-inertial odometry, benefiting of new odometry-capable camera hardware that recently became 
commercially available. 

As seen in Table~\ref{tab:tracking:results:pervid}, some sequences present large relative errors. Consider sequences V09 and V10, that show 
two faces of the same small orange tree under a challenging setting: wind is shaking the branches, occlusion by leaves is severe, and 
oranges are in a more mature stage than the examples in the training set. Considering less than 20 oranges are visible, a few missing 
oranges in these examples represents a large relative error. In this challenge setting, both the short-term tracking component, 
inter-frame association, and the long-term tracking component, 3-D relocalization, are jeopardized. Noteworthy that when 
the tracking results are integrated, i.e., when the 1,198 fruits/tracks are considered, the counting error rates are low. This is 
specially important considering that a yield prediction system must integrate several trees for accuracy. Despite large errors regarding 
single trees, the results point to tracker stability when considering a larger number of oranges. 

Multiple fruit tracking is a harder problem than fruit counting. Seeking for fruits in images, accurately identifying them, 
and evaluating when they are visible/reachable and when not is a challenge. Our reported values for HOTA and MOTA,
and the results reported by \citet{jong2022apple} and \citet{villacres2023apple} indicate MOT in orchards research still needs 
significant improvements to achieve high tracking rates, above 0.9. However, accurate fruit counting could be achieved before
highly accurate MOT. It is noteworthy that fruit tracking on the field is a challenging task even for humans.

Our yield regression confronts a significant challenge: understanding the relationship between observable and unobservable fruits. 
\citet{koirala2021unseen} delved into a similar aspect concerning mango trees. Initially, our yield regression displayed a modest 
prediction capability, with $R^2 = 0.61$. However, ensuring that a minimum of 30\% of the fruits were accurately identified by the 
computer vision-based counting system elevated the $R^2$ values beyond 0.80. Yet, the validity of assuming that at least 30\% of 
the yield is visible in the canopy, amenable to vision-based counting, warrants scrutiny. Is this assumption realistically grounded? 
We anticipate that sweet orange trees predominantly bear fruit in the outer or upper regions of the canopy due to enhanced sunlight 
exposure.

Upon scrutinizing the \textsc{MOrangeT} dataset, compiled from high-quality video recordings, it became evident that in most sequences 
the oranges identified by annotators corresponded to 30 to 50\% of the true yield ($F1 + F2 + F3$) obtained through fruit stripping. 
An exception was noted in sequence V03, where only 12.5\% of the fruits were discerned by annotators. This serves as evidence that 
low-quality video input significantly contributed to errors, and accurate regression from visible fruits holds promise. Moreover, 
this underscores the need for advancements in image acquisition systems, potentially incorporating solutions like artificial 
lighting during night operations and employing high dynamic range image sensors.

%
%
\section{Conclusion}
\label{sec:conclusion}

This work proposes a complete pipeline for yield estimation in citrus trees, comprising imaging methodology for fruit detection, multiple orange tracking
and yield regression integrating fruit counting to other tree data as  size and age. Exploiting ego-motion data, i.e., camera pose information,
we were able to create (i) a practical process for ground-truth annotation, and (ii) a relocalization module for multiple fruit tracking, able to deal
with long-term occlusions and fruit entering and exiting the camera's field of view. Framed as a multiple object tracking (MOT) problem, fruit counting was 
developed and evaluated utilizing established MOT metrics such as MOTA and HOTA. A useful intermediary result is the 3-D fruit localization that could be employed in other tasks and analysis, as robotic harvesting. Orange trees under standard crop management practices will present
non-visible fruit, hidden in the deeper parts of the canopy. A yield regressor, that takes to account tree size and age besides crop variety, in addition to 
image-based fruit counting data, was developed and validated considering a large set of trees (> 1,000) from a real crop forecast effort in one of
the largest sweet orange production sites in the world. In this challenging scenario, our pipeline was able to reach an $R^2$ up to 0.85 to true yield
from fruit stripping. Even higher values for $R^2$ could be reached for better quality video input. Future prospects for this research involve adapting the pipeline for deployment in vehicles and exploring contemporary deep learning-based models tailored for object tracking in image sequences.

\section*{CRediT authorship contribution statement}
\label{sec:credit}

\textbf{Thiago T. Santos}: Methodology, Software, Validation, Formal Analysis, Investigation, Data Curation, Writing -- Original Draft, 
Visualization. \textbf{Kleber X. S. de Souza}: Methodology, Software, Validation, Formal Analysis, Investigation, Data Curation, Writing -- Original Draft, Visualization. \textbf{João Camargo~Neto}: Methodology, Software, Formal Analysis, Investigation, Data Curation, Writing -- Review. \textbf{Luciano V. Koenigkan}: Methodology, Software, Formal Analysis, Investigation, Data Curation, Writing -- Review. \textbf{Alécio S. Moreira}: Data Curation, Writing -- Review and Editing. \textbf{Sônia Ternes}: Project administration, Methodology, Investigation, Validation, Data Curation, Writing -- Review and Editing.

\section*{Acknowledgments}
\label{sec:acknow}

This work was supported by the Brazilian Agricultural Research Corporation (Embrapa) under grant 10.18.03.016.00.00. T.~T.~Santos is 
partially funded by FAPESP (grants 2017/19282-7 and 2022/09319-9). We thank the PES/Fundecitrus team for providing video,
counting ground-truth and other plant data.

\bibliographystyle{plainnat}
\bibliography{references}  

\appendix

\section{HOTA}\label{apx:hota}

\newcommand{\bip}{\mathbf{b}_i^{(p)}}
\newcommand{\Bi}{\mathcal{B}_i}
\newcommand{\hbiq}{\hat{\mathbf{b}}_i^{(q)}}
\newcommand{\hBi}{\hat{\mathcal{B}}_i}
\newcommand{\bjr}{\mathbf{b}_j^{(r)}}
\newcommand{\hbjs}{\hat{\mathbf{b}}_j^{(s)}}
\newcommand{\TPa}{\mathrm{TP}^{(\alpha)}}
\newcommand{\FNa}{\mathrm{FN}^{(\alpha)}}
\newcommand{\FPa}{\mathrm{FP}^{(\alpha)}}
\newcommand{\TPAa}{\mathrm{TPA}^{(\alpha)}}
\newcommand{\FNAa}{\mathrm{FNA}^{(\alpha)}}
\newcommand{\FPAa}{\mathrm{FPA}^{(\alpha)}}
\newcommand{\pid}{\mathrm{pid}}
\newcommand{\gid}{\mathrm{gid}}
\newcommand{\mipq}{\mathbf{m}_i^{(p,q)}}
\newcommand{\mjrs}{\mathbf{m}_j^{(r,s)}}
\newcommand{\HOTA}{\mathrm{HOTA}}

Consider an assignment matrix $\hat{\mathtt{A}}_i^{(\alpha)}[p,q]$ that matches bounding boxes $\bip \in \Bi$ to \emph{ground-truth boxes} $\hbiq \in \hBi$, 
the set of annotated bounding boxes for frame $f_i$, ensuring that $\mathrm{IoU}(\bip, \hbiq) \geq \alpha$. The assignments are one-to-one: a box $\bip$ can
be assigned to at most one box $\hbiq$ in the ground-truth and vice versa. As seen in Section~\ref{sec:mot}, $\hat{\mathtt{A}}_i^{(\alpha)}[p,q] = 1$ in the
case of assignment and zero otherwise. Let $\mipq = \langle \mathbf{b}_i^{(p)}, \hat{\mathbf{b}}_i^{(q)} \rangle$ be a compact representation for an 
assignment. The set of \emph{true positives} $\TPa_i$ for frame $f_i$ is defined as:
\begin{equation}\label{eq:TPai}
\mathrm{TP}_{i}^{(\alpha)} = \{ \mipq = \langle \bip, \hbiq \rangle \mid \hat{\mathtt{A}}_i^{(\alpha)}[p,q] = 1 \}
\end{equation}
Integrating the true positives for all $M$ frames, we have the set $\TPa$:
\begin{equation}\label{eq:TPa}
\TPa = \bigcup_{i=1}^{M} \TPa_i
\end{equation}
Similarly, the \emph{false negatives} set $\FNa_i$ for $f_i$ and the set of all false negatives $\FNa$ are defined as:
\begin{equation}\label{eq:FNai}
\FNa_i = \{ \hbiq \mid \sum_p \hat{\mathtt{A}}_i^{(\alpha)}[p,q] = 0\}
\end{equation}
(bounding boxes in the ground-truth not matched to any detected box) and
\begin{equation}\label{eq:FNa}
\FNa = \bigcup_{i=1}^{M} \FNa_i
\end{equation}
Finally, the \emph{false positive} set $\FPa_i$ for $f_i$ and the set of all false positives $\FPa$ are defined as:
\begin{equation}\label{eq:FPai}
\FPa_i = \{ \bip \mid \sum_{q} \hat{\mathtt{A}}_i^{(\alpha)}[p,q] = 0\}
\end{equation}
(detected boxes not assigned to any box in the ground-truth) and
\begin{equation}\label{eq:FPa}
\FPa = \bigcup_{i=1}^{M} \FPa_i
\end{equation}

Assume $\gid(\cdot): \bigcup_{i}\hBi \rightarrow \mathbb{N}$ a function that maps boxes in the ground-truth to numeric identifiers, and $\pid(\cdot): 
\bigcup_{i}\mathcal{B}_i \rightarrow \mathbb{N}$ a function that does the same for detected boxes. To evaluate tracking in HOTA, \citet{luiten2021hota} 
proposed three novel concepts. For a matching $\mipq$, the 
\emph{true positive associations} set $\TPAa$ is composed of all true positives $\mjrs$ that present the same IDs 
for prediction and ground-truth as $\mipq$:
\begin{equation}\label{eq:TPA}
\TPAa(\mipq) = \{ \mjrs \in  \TPa \mid  \pid(\bip) = \pid(\bjr) \wedge 
\gid(\hbiq) = \gid(\hbjs) \}
\end{equation}
The \emph{false negative associations} set $\FNAa$ is composed of matches that present the same ground-truth identifier $\gid(\hbiq)$ that $\mipq$, but detections attributed to different IDs, plus the set of false negatives also identified by $\gid(\hbiq)$: 
\begin{equation}\label{eq:FNA}
\begin{split}
\FNAa(\mipq) & = \{ \mjrs \in  \TPa \mid  \pid(\bip) \neq \pid(\bjr) \wedge 
\gid(\hbiq) = \gid(\hbjs) \\
& \cup \{ \hbjs \in \FNa  \mid \gid(\hbiq) = \gid(\hbjs) \}
\end{split}
\end{equation}
The last concept is the \emph{false positive associations} set $\FPAa$, composed of matches that present the 
same detected identifier $\pid(\bip)$, but whose ground-truth presents different identifiers, plus all 
false positives also identified by $\pid(\bip)$:
\begin{equation}\label{eq:FPA}
\begin{split}
\FPAa(\mipq) & = \{ \mjrs \in  \TPa \mid  \pid(\bip) = \pid(\bjr) \wedge 
\gid(\hbiq) \neq \gid(\hbjs) \\
& \cup \{ \bjr \in \FPa  \mid \pid(\bip) = \pid(\bjr) \}
\end{split}
\end{equation}

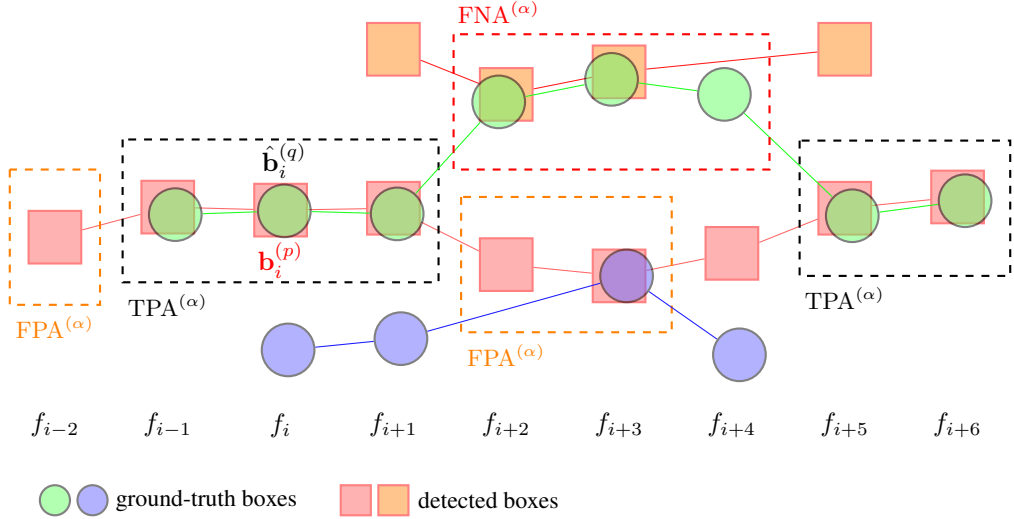
\begin{figure}
\centering

\tikzstyle{bbox1}=[shape=rectangle, draw=red, thick, fill=red!60, semitransparent,minimum size=7.0mm]
\tikzstyle{bbox2}=[shape=rectangle, draw=red, thick, fill=orange!90, semitransparent,minimum size=7.0mm]
\tikzstyle{gbox1}=[shape=circle, draw, thick, fill=green!60, semitransparent,minimum size=7.0mm]
\tikzstyle{gbox2}=[shape=circle, draw, thick, fill=blue!60, semitransparent,minimum size=7.0mm]
\tikzstyle{TPA}=[shape=rectangle, draw, thick, dashed]
\tikzstyle{FPA}=[shape=rectangle, draw=orange, thick, dashed]
\tikzstyle{FNA}=[shape=rectangle, draw=red, thick, dashed]

\begin{tikzpicture}
  \path node (b1) [bbox1] at (0,0) {};
  \path (b1)+(15mm, +4mm) node (b2) [bbox1] {};
  \path (b2)+(15mm, -0.4mm) node (b3) [bbox1] {};
  \path (b3)+(15mm, +0.25mm) node (b4) [bbox1] {};
  \path (b3.south)+(0, -3mm) node [text centered, red] {$\mathbf{b}_i^{(p)}$};
  \path (b4)+(15mm, -7.5mm) node (b5) [bbox1] {};
  \path (b5)+(15mm, -1.5mm) node (b6) [bbox1] {};
  \path (b6)+(15mm, +3mm) node (b7) [bbox1] {};
  \path (b7)+(15mm, +6mm) node (b8) [bbox1] {};
  \path (b8)+(15mm, +1.5mm) node (b9) [bbox1] {};
  \path [draw, red!60] (b1) -- (b2) -- (b3) -- (b4) -- (b5) -- (b6) -- (b7) -- (b8) -- (b9);

  \path (b2)+(-6mm, +9mm) node (x1) {};
  \path (b2)+(36mm, -10mm) node (x2) {};
  \draw[TPA] (x1) rectangle (x2);
  \path (b2)+(0mm, -13mm) node [align=left] {\footnotesize $\TPAa$};
  
  \path (b8)+(-6mm, +9mm) node (y1) {};
  \path (b8)+(22mm, -9mm) node (y2) {};
  \draw[TPA] (y1) rectangle (y2);
  \path (b8)+(0mm, -12mm) node {\footnotesize $\TPAa$};

  \path (b5)+(-6mm, +9mm) node (z1) {};
  \path (b5)+(22mm, -9mm) node (z2) {};
  \draw[FPA] (z1) rectangle (z2);
  \path (b5)+(0mm, -12mm) node [orange] {\footnotesize $\FPAa$};

  \path (b1)+(-6mm, +9mm) node (w1) {};
  \path (b1)+(6mm, -9mm) node (w2) {};
  \draw[FPA] (w1) rectangle (w2);
  \path (b1)+(0mm, -12mm) node [orange] {\footnotesize $\FPAa$};

  \path node (bp1) [bbox2] at (45mm, 25mm) {};
  \path (bp1)+(15mm, -6mm) node (bp2) [bbox2] {};
  \path (bp2)+(15mm, +3mm) node (bp3) [bbox2] {};
  \path (bp3)+(30mm, +3mm) node (bp4) [bbox2] {};
  \path [draw, red!90] (bp1) -- (bp2) -- (bp3) -- (bp4);  

  \path node (g1) [gbox1] at (16mm, +3mm) {};
  \path (g1)+(14.5mm, +0.5mm) node (g2) [gbox1] {};
  \path (g2)+(15mm, -0.4mm) node (g3) [gbox1] {};
  \path (g2.north)+(0, +3mm) node [text centered] {$\hat{\mathbf{b}}_i^{(q)}$};
  \path (bp2)+(-1mm, -1mm) node (g4) [gbox1] {};
  \path (g4)+(15mm, +3mm) node (g5) [gbox1] {};
  \path (g5)+(15mm, -2mm) node (g6) [gbox1] {};
  \path (b8)+(1mm, -1mm) node (g7) [gbox1] {};
  \path (g7)+(15mm, +2mm) node (g8) [gbox1] {};
  \path [draw, green!90] (g1) -- (g2) -- (g3) -- (g4) -- (g5) -- (g6) -- (g7) -- (g8);  

  \path node (f1) [gbox2] at (31mm, -15mm) {};
  \path (f1)+(15mm, +1.5mm) node (f2) [gbox2] {};
  \path (b6)+(1mm, 0) node (f3) [gbox2] {};
  \path (f3)+(15mm, -10.5mm) node (f4) [gbox2] {};
  \path [draw, blue!90] (f1) -- (f2) -- (f3) -- (f4);  

  \path (g4)+(-6mm, +9mm) node (v1) {};
  \path (g4)+(36mm, -9mm) node (v2) {};
  \draw[FNA] (v1) rectangle (v2);
  \path (g4)+(0mm, 12mm) node [red] {\footnotesize $\FNAa$};

  \path node (gleg1) [gbox1, minimum size=4.0mm] at (0mm, -35mm) {};
  \path node (gleg2) [gbox2, minimum size=4.0mm] at (5mm, -35mm) {};
  \path (gleg2.east)+(13mm, 0mm) node [align=left] {\footnotesize ground-truth boxes};
  \path node (bleg1) [bbox1, minimum size=4.0mm] at (40mm, -35mm) {};
  \path node (bleg2) [bbox2, minimum size=4.0mm] at (45mm, -35mm) {};
  \path (bleg2.east)+(10.5mm, 0mm) node [align=left] {\footnotesize detected boxes};

  \path node (f1)  at (0,-25mm) {$f_{i-2}$};
  \path (f1)+(15mm, 0mm) node (f2) {$f_{i-1}$};
  \path (f2)+(15mm, 0mm) node (f3) {$f_{i}$};
  \path (f3)+(15mm, 0mm) node (f4) {$f_{i+1}$};
  \path (f4)+(15mm, 0mm) node (f5) {$f_{i+2}$};
  \path (f5)+(15mm, 0mm) node (f6) {$f_{i+3}$};
  \path (f6)+(15mm, 0mm) node (f7) {$f_{i+4}$};
  \path (f7)+(15mm, 0mm) node (f8) {$f_{i+5}$};
  \path (f8)+(15mm, 0mm) node (f9) {$f_{i+6}$};
\end{tikzpicture}

\caption{The three association sets in HOTA: true positive association ($\TPAa$), composed by pairs (matches) of detected and 
ground truth boxes (marked in dashed black lines); false negative association ($\FNAa$) composed by false negatives and matches 
with a different detection ID (red dashed lines), and false 
positive association ($\FPAa$), composed by false positives and matches with different ground-truth ID (orange dashed lines). 
Ground-truth boxes are displayed as circles for visualization purposes. The colors represent the different IDs for 
ground-truth and detected boxes. Based on Figure~2 in \citet{luiten2021hota}.}
\label{fig:hota}
\end{figure}

Figure~\ref{fig:hota} illustrates the three kinds of association sets. Finally, $\HOTA_\alpha$ can be defined as
\begin{equation}\label{eq:HOTAa}
    \mathrm{HOTA}_\alpha = \sqrt{\frac{\sum_{\mipq \in \TPa} \mathcal{A}(\mipq)}{|\TPa| + |\FNa| + |\FPa|}}
\end{equation}
where $\mathcal{A}(\mipq)$ measures the alignment between predicted and ground-truth tracks: 
\begin{equation}
    \mathcal{A}(\mipq) = \frac{|\TPAa(\mipq)|}{|\TPAa(\mipq)| + |\FNAa(\mipq)| + |\FPAa(\mipq)|}
\end{equation}

Luiten et al. call this a \emph{double Jaccard} formulation: the Jaccard metric is employed on the evaluation of
detection, with the matches $\mipq \in \TPa$ being weighted by the association score $\mathcal{A}(\mipq)$, which also
is a Jaccard metric. The proponents also argue that the metric is the geometric mean of a detection score and an association score, 
considering that
\begin{equation}\label{eq:DetAa}
  \mathrm{DetA}_\alpha = \frac{|\TPa|}{|\TPa| + |\FNa| + |\FPa|}
\end{equation}
\begin{equation}\label{eq:AssAa}
  \mathrm{AssA}_\alpha = \frac{1}{|\TPa|} \sum_{\mipq \in \TPa} \mathcal{A}(\mipq)
\end{equation}
\begin{equation}\label{eq:geomean}
\begin{split}
\HOTA_\alpha & = \sqrt{\frac{\sum_{\mipq \in \TPa} \mathcal{A}(\mipq)}{|\TPa| + |\FNa| + |\FPa|}}\\
             & = \sqrt{\mathrm{DetA}_\alpha \cdot \mathrm{AssA}_\alpha}
\end{split}
\end{equation}

$\HOTA_\alpha$ evaluates detection and association, but not the localization accuracy for the matches, i.e., the spatial fit 
between boxes' areas. Localization is evaluated by the final HOTA score, that integrates $\HOTA_\alpha$ for different 
values of $\alpha$:
\begin{equation}\label{eq:HOTA}
    \HOTA = \int_0^1 \HOTA_\alpha d\alpha \approx \frac{1}{19} \sum_{\alpha \in [0.05, 0.1, \dots, 0.9, 0.95]} \HOTA_\alpha
\end{equation}

Equations~\ref{eq:TPA} to \ref{eq:HOTA} correspond to \citet{luiten2021hota} formulations, rewritten to employ the notations
adopted in the present work. It is important to note that, like in MOTA \citep{bernardin2008evaluating}, the employed assignments 
$\hat{\mathtt{A}}_i^{(\alpha)}[p,q]$ are the ones that maximize the final HOTA score (see \citet{luiten2021hota} 
for details about the matching optimization procedure). Here, we will employ HOTA for the evaluation of 
our multiple orange tracking, and $\mathrm{DetA}$ and $\mathrm{AssA}$ scores to decompose tracking assessment 
into its detection and association components (note that DetA and AssA metrics are computed from DetA$_\alpha$ and AssA$_\alpha$ 
integrating different $\alpha$ thresholds, as seen for HOTA in Equation~\ref{eq:HOTA}). We have employed the original code from HOTA 
authors\footnote{\url{https://github.com/JonathonLuiten/TrackEval.git}.}. 

\end{document}